\documentclass[a4paper,11pt]{article}
\usepackage[utf8]{inputenc}
\usepackage[margin = 0.5in]{geometry}
\usepackage{epsfig}
\usepackage{graphics}
\usepackage{graphicx}
\usepackage{tikz}
\usetikzlibrary{positioning}
\usepackage{pgfplots}
\usepackage{stmaryrd}
\usepackage[font=small,labelfont=bf]{caption}
\usetikzlibrary{calc,decorations.markings}
\usepackage{mathtools}
\usepackage[colorlinks]{hyperref}
\usepackage{color}
\usepackage{bm}
\usepackage{comment}

\usepackage{amsmath}
\usepackage{amsthm}
\numberwithin{equation}{section}
\usepackage{amssymb}
\usepackage{multirow}
\usepackage{tabularx,ragged2e,booktabs,caption}
\usepackage{amsthm}
\usepackage{mathptmx}
\usepackage{mathrsfs}
\theoremstyle{plain}
\newtheorem{theorem}{Theorem}[section]
\theoremstyle{plain}
\newtheorem{lemma}{Lemma}[section]
\newtheorem{definition}[theorem]{Definition}

\newtheorem{proposition}{Proposition}[section]

\newtheorem{corollary}{Corollary}[section]
\newtheorem{rem}{Remark}[section]
\numberwithin{equation}{section}
\usepackage[colorlinks]{hyperref}
\hypersetup{
    colorlinks=true,
     linkcolor=blue,
     citecolor=blue}

\usepackage{listings}
\usepackage{algorithm}
\usepackage[noend]{algpseudocode}

\usepackage[toc,page]{appendix}
\usepackage{authblk}

\makeatletter
\def\BState{\State\hskip-\ALG@thistlm}
\makeatother

\makeatletter % Definition de la norme operateur
\newcommand{\opnorm}{\@ifstar\@opnorms\@opnorm}
\newcommand{\@opnorms}[1]{%
  \left|\mkern-1.5mu\left|\mkern-1.5mu\left|
   #1
  \right|\mkern-1.5mu\right|\mkern-1.5mu\right|
}
\newcommand{\@opnorm}[2][]{%
  \mathopen{#1|\mkern-1.5mu#1|\mkern-1.5mu#1|}
  #2
  \mathclose{#1|\mkern-1.5mu#1|\mkern-1.5mu#1|}
}

\def\XXint#1#2#3{{\setbox0=\hbox{$#1{#2#3}{\int}$}
     \vcenter{\hbox{$#2#3$}}\kern-.5\wd0}}

\makeatother

\usepackage{pythonhighlight}

\usepackage{xcolor}
\usepackage{titlesec}
\titleformat{\part}[block]{\color{blue}\Huge\bfseries\filcenter}{}{0.5em}{}

\usepackage[square,sort,comma,numbers]{natbib}
\setcitestyle{plain, open={[},close={]}}

\usepackage{cancel}

\title{\textbf{Generalized infinite dimensional Alpha-Procrustes based geometries}}

\author{
Salvish Goomanee\footnote{\textit{Corresponding author.}} \,\thanks{CIRB, Collège de France, CNRS, INSERM, Université PSL, Paris, France. \textit{Current address}: Universit\'e C\^ote d'Azur, INRIA, I3S, Sophia Antipolis, France (salvish.goomanee@inria.fr).}, \,\,\, Andi Han\thanks{University of Sydney, Australia and RIKEN AIP, Japan. (andi.han@sydney.edu.au).}, \,\,\, Pratik Jawanpuria\thanks{Indian Institute of Technology Bombay, India (pratik.jawanpuria@iitb.ac.in ).},  \,\,\, Bamdev Mishra\thanks{Microsoft India (bamdevm@microsoft.com).}}

\begin{document}
\maketitle

\begin{abstract}
    This work extends the recently introduced Alpha-Procrustes family of Riemannian metrics for symmetric positive definite (SPD) matrices by incorporating generalized versions of the Bures-Wasserstein (GBW), Log-Euclidean, and Wasserstein distances. While the Alpha-Procrustes framework has unified many classical metrics in both finite and infinite dimensional settings, it previously lacked the structural components necessary to realize these generalized forms. We introduce a formalism based on unitized Hilbert-Schmidt operators and an extended Mahalanobis norm that allows the construction of robust, infinite dimensional generalizations of GBW and Log-Hilbert-Schmidt distances. Our approach also incorporates a learnable regularization parameter that enhances geometric stability in high dimensional comparisons. Preliminary experiments reproducing benchmarks from the literature demonstrate the improved performance of our generalized metrics, particularly in scenarios involving comparisons between datasets of varying dimension and scale. This work lays a theoretical and computational foundation for advancing robust geometric methods in machine learning, statistical inference, and functional data analysis.
\end{abstract}

\tableofcontents

\section{Introduction $\bf{\&}$ motivations}
Symmetric positive definite (SPD) matrices, in particular covariance matrices and operators are central to a wide range of problems in data science  \citep{brooks2019exploring, guillaumin2009you, mahadevan2018unified, pennec2006riemannian, tsuda2005matrix} with applications ranging from kernel methods \citep{kulis2009low}, generative modeling \citep{de2022riemannian}, brain imaging by virtue of diffusion tensors \citep{dryden2009non} to human-based detection \citep{tosato2012characterizing, tuzel2008pedestrian}. The geometry of SPD matrices has been extensively studied in the finite dimensional setting with popular metrics such as the affine-invariant \citep{Bhatia2007, pennec2006riemannian, thanwerdas2019affine}, Log-Euclidean \citep{arsigny2006log, arsigny2007geometric}, and Bures-Wasserstein (BW) distances \citep{BJL2019, HMJG2021a, malago2018wasserstein, oostrum2022bures}. While the aforementioned distances induce a Riemannian metric and geometry on the manifold of SPD matrices, there exist a number of approaches that exploit instead divergences which are not Riemannian metrics \citep{sra2012new, sra2016positive, sra2021metrics, kulis2009low}. In this work we, however, focus on the BW metric (and distance) and its generalizations in finite and infinite dimensions. \\

The BW metric has found widespread applications in a variety of fields, including optimal statistical transport \citep{Bhatia2007, de2022riemannian}, computer graphics \citep{rabin2011wasserstein, solomon2015convolutional}, neurosciences \citep{gramfort2015fast}, and multi-omic data alignment \citep{demetci2020gromov}, among others. These are real-world applications which leverage the finite dimensional versions of the BW metric. However, a number of real-world applications may involve data which lie on very high dimensional spaces (\textit{i.e.}, multiple samples with very large feature space)  requiring, for example, the construction of SPD operators defined on infinite-dimensional Hilbert spaces \citep{shnitzer2022}. Unfortunately, most existing geometric frameworks fail to adequately generalize coherently across finite and infinite dimensions, leading to inconsistencies in modeling, analysis, and computation of noise in high dimensions for example. \\

To address this, \textbf{we propose a unifying family of Riemannian distances based on generalized alpha-Procrustes distances}. This includes the Log-Hilbert-Schmidt \citep{Minh2014, MINH202225} and infinite dimensional version of the recently introduced generalised Bures-Wasserstein \citep{HMJG2021b} distances as special cases which consequently allows for a continuous interpolation between them. It is designed to extend smoothly from finite dimensional SPD matrices to infinite dimensional positive-definite Hilbert-Schmidt operators, offering a robust and flexible geometric foundation for both theoretical analysis and practical machine learning applications. Consequently, establishing a unifying framework of unification facilitates learning across spaces of varying dimension and complexity, while preserving key geometric and spectral structures.  \\

In order to achieve this, we first explore the class of parametrized metrics \citep{MINH202225} defined as the \textbf{alpha-Procrustes family of metrics} which rigorously establishes a unified framework for the Bures-Wasserstein (BW), Log-Euclidean (Log-EU), and Wasserstein geometries between Gaussian measures with respect to the Euclidean functional space. This lays the foundation for a generalized definition of alpha-Procrustes-based geometries on an infinite dimensional manifold. A short discussion is provided in Appendix \ref{theorems}. In finite dimensions the alpha-Procrustes distances are shown to be equivalent to a a family of Riemannian metrics on the manifold of symmetric positive definite matrices \citep{Bhatia2007}. The aforementioned results are then generalized to the case where the functional space is an infinite dimensional separable Hilbert space $\mathcal{H}$ where the SPD matrices are represented by positive definite unitized Hilbert-Schmidt (HS) operators by \cite{Minh2014, MINH202225}. In this setting one can write down the infinite dimensional counterpart of the BW metric without much difficulty. This follows from the fact the optimal transport (OT) and Procrustes formulation remain licit when dim $\mathcal{H} = \infty$, where $\mathcal{H}$ represent a Hilbert space. The Log-EU metric on the other hand metric requires a more in depth treatment as pointed out in the papers \citep{Minh2014, MINH202225}. This leads naturally to the introduction of \textit{Log-Hilbert-Schmidt geometry} and the definition of a non-trivial functional space. \textbf{The core contribution of our work is the rigorous generalization of this construction beyond its original setting}. \\

Our starting point is the realization of the recently introduced generalized Bures-Wasserstein (GBW) metric \citep{HMJG2021b} as a special case of the alpha-Procrustes family of metrics in finite dimensions. We then propose a generalization procedure for when dim $\mathcal{H} = \infty$ which is laid out in the Section \ref{infinitDim}. 
$\mathbb{S}^{n}$ denotes the set of symmetric matrices and
\begin{equation}
    \mathbb{S}_{++}^{n} = \{ \textbf{X} : \textbf{X} \in \mathbb{R}^{n \times n}, \, \textbf{X}^\top = \textbf{X}, \, \textbf{X} \succ 0 \},
\end{equation}
where $\mathbb{S}_{++}^{n} \subset \mathbb{S}^{n} $ denotes the set of symmetric positive definite (SPD) matrices. Let $\mathbb{O}(n)$ denote the set of $n \times n$ orthogonal matrices. For the purpose of clarity, we recall the $p$-Wasserstein distance as well as the Bures-Wasserstein distance. The optimal transport problem \citep{villani2009} between two measures $\mu$ and $\nu$ over metric spaces $\mathcal{X}$ and $\mathcal{Y}$ respectively is 
\begin{equation}
    C(\mu, \nu) = \inf_{\pi \in \Gamma(\mu, \nu)}\int _{\mathcal{X} \times \mathcal{Y}} c(x, y)d\gamma(x, y),
\end{equation}
where $\gamma$ represent the transport plan and $\Gamma(\mu, \nu)$ is the set of joints distributions with marginals $\mu$ and $\nu$. $c : \mathcal{X} \times \mathcal{Y} \rightarrow \mathbb{R}_{+} : (x, y) \mapsto c(x, y)$ is the cost for transporting one unit mass from $x$ to $y$. When the cost function $c$ is expressed in terms of a distance one can construct a valid definition of the distance between the measures $\mu$ and $\nu$. 
\begin{definition}
    \citep{villani2009}\label{wassersteinDistance}
    Let $(\mathcal{X}, d)$ and $(\mathcal{Y}, d)$ be Polish metric spaces, and let $p \in [1, \infty)$. $\mathcal{B}(\mathcal{X}, d) = \mathcal{B}(\mathcal{X})$ will be the $\sigma$-algebra of Borel sets on this space and $\mathcal{P}(\mathcal{X}, d) = \mathcal{P}(\mathcal{X})$ the set of probability measures defined on $\mathcal{B}(\mathcal{X})$. Then, for any two probability measures $\mu, \nu$ on $\mathcal{X} \times \mathcal{Y}$, the Wasserstein distance of order $p$ between $\mu$ and $\nu$ is defined by 
    \begin{equation}\label{W_p distance}
        \begin{split}
            \mathcal{W}_{p}(\mu, \nu) & := \bigg\{ \inf_{\pi \in \Gamma(\mu, \nu)} \int _{\mathcal{X} \times \mathcal{Y}} d(x, y)^{p} d\gamma(x, y) \bigg\}^{1/p}  \\ & = \inf \bigg\{ \bigg[ \mathbb{E}\,d(X, Y)^{p}\bigg]^{1/p}, \, \mathrm{law}(X)=\mu, \, \mathrm{law}(Y)=\nu \bigg\}
        \end{split}
    \end{equation}
\end{definition}

Let $\textbf{X}, \textbf{Y} \in \mathbb{S}_{++}^{n}$. Let $\mu = \mathcal{N}(m_{1}, \textbf{X})$ and $\nu = \mathcal{N}(m_{1}, \textbf{Y})$ be two Gaussian measures with mean $m_{2}$ and $m_{2}$ respectively on $\mathbb{R}^{n}$. Denote $\Gamma(\mu, \nu)$ the set of joints distributions on $\mathbb{R}^{n} \times \mathbb{R}^{n}$ with marginals $\mu$ and $\nu$. By virtue of Definition (\ref{wassersteinDistance}) for $p=2$ and \citep{dowson1982frechet, Gelbrich1990, givens1984class, olkin1982distance}, the $2$-Wasserstein distance $d_{\mathrm{W}}$ between $\mu$ and $\nu$ is given by
\begin{equation}\label{W_2Distance}
    \begin{split}
        \mathcal{W}_{2}(\mu, \nu) =  \bigg\{ \inf_{\pi \in \Gamma(\mu, \nu)} \int _{\mathbb{R}^{n} \times \mathbb{R}^{n}} d(x, y)^{2} & d\gamma(x, y) \bigg\}^{1/2} \\ & = \vert\vert m_{1} - m_{2} \vert\vert_{\text{F}}^{2} + \mathrm{tr}\big[ \textbf{X} +\textbf{Y} - 2(\textbf{X}^{1/2}\textbf{Y}\textbf{X}^{1/2})^{1/2}\big]^{1/2}, 
    \end{split}
\end{equation}
where $d(x, y) = \vert\vert x -y \vert\vert_{\mathrm{F}}$ and the norm $\vert\vert \cdot \vert\vert_{\mathrm{F}}$ corresponds to the Frobenius norm\footnote{The Frobenius (F) norm generalises to the Hilbert-Schmidt (HS) norm for the case of infinite dimensional Hilbert spaces; one simply replaces $\vert\vert \cdot \vert\vert_{\text{Frobenius}}$ by $\vert\vert \cdot \vert\vert_{\mathrm{HS}}$.}. Eventually, for $m_{1} = m_{2}= 0$ the Bures-Wasserstein (BW) distance on $\mathbb{S}_{++}^{n}$ ensues
\begin{equation}\label{bwDistance}
    d_{\mathrm{BW}}(\textbf{X}, \textbf{Y}) = \big[ \mathrm{tr}(\textbf{X}) + \mathrm{tr}(\textbf{Y}) - 2\mathrm{tr}(\textbf{X}^{1/2}\textbf{Y}\textbf{X}^{1/2})^{1/2}\big]^{1/2}
\end{equation}
which corresponds to the $2$-Wasserstein distance between zero-centered non-degenerate Gaussian measures.  As established in \citep{BJL2019}, the BW distance can be constructed, also, from the Procrustes optimization problem, \textit{i.e.}, $ \min_{\textbf{O} \in \mathbb{O}(n)}\vert\vert \textbf{X}^{1/2} - \textbf{Y}^{1/2}\textbf{O} \vert\vert_{\mathrm{F}}$. The minimum is reached when $\textbf{O}$ is the unitary polar factor of $\textbf{Y}^{1/2}\textbf{X}^{1/2}$. The space of the $\mathbb{S}_{++}^{n}$ matrices can viewed as a quotient space on the general linear group $\mathrm{GL}(n)$ with the action of the orthogonal group $\mathbb{O}(n)$. The quotient map $\pi: \mathrm{GL}(n) \rightarrow \mathrm{GL}(n) / \mathbb{O}(n)$ thus defines a Riemannian submersion such that the distance in (\ref{bwDistance}) is effectively a Riemannian distance on $\mathbb{S}_{++}^{n}$ induced from the submersion $\pi$ \citep{BJL2019}. Note that the above constructions carry over to the set of positive semidefinite matrices as well. 
\paragraph{Organization of paper.}
In Section \ref{finiteDim} of the paper we provide a brief discussion recalling the Mahalanobis norm and the realization of the generalized Bures-Wasserstein metric \citep{HMJG2021b}. We also provide a generalized formulation of the Log-Euclidean distance in Appendix \ref{finiteDimGeneralizedLogEU} realized from the generalized alpha-Procrustes geometry. Section \ref{infinitDim} introduces the necessary functional analytic details for establishing a generalized infinite dimensional setting. The core part of this section is the construction of the extended Mahalanobis norm on the infinite dimensional Hilbert space that describe the manifold of positive definite extended (unitized) Hilbert-Schmidt operators. This section builds on the work of \citep{Minh2021, MINH202225}. This allows us to introduce the \textbf{infinite dimensional generalised alpha-Procrustes} distance (Proposition \ref{generalisedAlphaProcrusteInfDim}). We then show how the infinite dimensional versions of the GBW, generalized Wasserstein distances and generalized Log-Hilbert-Schmidt emerges from the latter. Finally, we present some early applications of this framework which crown our effort of establishing a unifying class of alpha-Procrustes distances before concluding in Section \ref{discussions}.

\section{Alpha-Procrustes based geometries: the finite dimensional setting}\label{finiteDim}
We proceed to define the generalized Bures-Wasserstein distance (GBW) as introduced in \citep{HMJG2021b}. In the latter, the authors show how this generalization of the BW geometry, the Procrustes distance and the Wasserstein distance for Gaussian measures coincide. The GBW represents the multivariate extension of the BW distance. This representation is particularly useful when dealing with multiple datasets presenting large amount of interconnected data. In such cases the BW distance with respect to the $\vert\vert \cdot \vert\vert_{\text{Frobenius}}$-norm is not efficient for characterizing such distances among the data points. The GBW distance represents a more robust way of measuring the distances. We, therefore, explore the GBW distance and show that it can be also realized as a special case of the class of Alpha-Procrustes distances. This represents a trivial generalization of the results in \citep{MINH202225} where we now express the latter in terms of the Mahalanobis norm $\vert\vert \cdot \vert\vert_{\textbf{M}^{-1}}$.

\begin{definition}(\textbf{Generalized Bures-Wasserstein} \citep{HMJG2021b})\label{gbwDistance}
    Let ${\bf{X}}, {\bf{Y}} \in \mathbb{S}_{++}^{n}$. Let ${\bf{M}}\in \mathbb{S}_{++}^{n}$, where ${\bf{M}}$ is a SPD matrix. The generalized Bures-Wasserstein distance is expressed as 
    \begin{equation}\label{GBW}
        d_{\mathrm{GBW}}({\bf{X}}, {\bf{Y}}) = \big[ \mathrm{tr}({\bf{M}}^{-1}{\bf{X}}) + \mathrm{tr}({\bf{M}}^{-1}{\bf{Y}})
        - 2\mathrm{tr}({\bf{X}}^{1/2}{\bf{M}}^{-1}{\bf{Y}}{\bf{M}}^{-1}{\bf{X}}^{1/2})^{1/2}\big]^{1/2}.
    \end{equation}
    The Mahalanobis norm is defined as $\vert\vert {\bf{X}} \vert\vert_{{\bf{M}}^{-1}} := \sqrt{\mathrm{tr}({\bf{X}}^\top{\bf{M}}^{-1}{\bf{X}})}$. When $\bf{M} = \bf{I}$, the Mahalanobis norm reduces to the standard Frobenius norm and the GBW distance reduces to the classical Bures-Wasserstein \cite{HMJG2021b}.
\end{definition}
\noindent
For the sake of clarity, we recall Proposition 1 of \citep{HMJG2021b} which realizes the GBW as a solution of the orthogonal Procrustes problem. 
\begin{theorem}\label{propGBWProcrustes}\citep{HMJG2021b}
    The generalized Bures-Wasserstein distance $d_{\mathrm{GBW}}$ (\ref{GBW}) is the solution of the orthogonal Procrustes distance
    \begin{equation}\label{GBWProcrustes}
        d_{\mathrm{GBW}}({\bf{X}}, {\bf{Y}}) = \min_{{\bf{O}} \in \mathbb{O}(n)} \vert\vert {\bf{X}}^{1/2} - {\bf{Y}}^{1/2}{\bf{O}}\vert\vert_{{\bf{M}}^{-1}}.
    \end{equation}
\end{theorem}
\noindent
Proceeding along the lines of Theorem \ref{propGBWProcrustes}, one remarks that the GBW distance can be realized also as special case of the Alpha-Procrustes distance with respect to the Mahalanobis norm. The latter depends on a parameter $\alpha$ where different values of alpha leads to the different distances studied in this work as it will be shown. We point out that the results in \citep{MINH202225} are established for $\alpha \geq 1/2$. The exploration of distances arising from alternative values of $\alpha$ in this setting (\textit{i.e.}, different from $0$ and $1/2$) lies beyond the scope of this work and is deferred to future investigations. 
\begin{definition}\label{alphaProcrustes}
    Let $\alpha \in \mathbb{R}_{>0}$ be fixed. The generalised $\alpha$-Procrustes distance between two matrices ${\bf{X}}, {\bf{Y}} \in \mathbb{S}^{n}_{++}$ is 
    \begin{equation}\label{alphaProcrustesM}
        d^{\alpha}_{\mathrm{proM}}({\bf{X}}, {\bf{Y}}) = \min_{{\bf{O}}\in \mathbb{O}(n)}\bigg\vert\bigg\vert
        \frac{{\bf{X}}^{\alpha} - {\bf{Y}}^{\alpha}{\bf{O}} }{\alpha}\bigg\vert\bigg\vert_{{\bf{M}}^{-1}}, 
    \end{equation}
    where $\vert\vert \cdot \vert\vert_{{\bf{M}}^{-1}}$ is the Mahalanobis norm. The notation $d^{\alpha}_{\mathrm{proM}}$ means the Procrustes distance with respect to the Mahalanobis norm in finite dimensions \footnote{We follow the notations of \citep{MINH202225} and keep this for the rest of the work.}.   

    For $\alpha \in \mathbb{R}^{+}_{0}$, $d^{\alpha}_{\mathrm{proM}}$ is defined on the set of positive semidefinite matrices $\mathbb{S}^{n}_{++}$.
\end{definition}
\begin{proposition}\label{propAlphaGBW}
    Let ${\bf{X}}, {\bf{Y}} \in \mathbb{S}_{++}^{n}$. Let $\alpha = 1/2$, one has that
    \begin{equation}
        d^{1/2}_{\mathrm{proM}}({\bf{X}}, {\bf{Y}}) = 2d_{\mathrm{GBW}}({\bf{X}}, {\bf{Y}}),
    \end{equation}
    which is precisely the generalized Bures-Wasserstein distance (\ref{GBW}) \citep{HMJG2021b}.
\end{proposition}
\noindent
The details of the proof and the generalization of the Log-Euclidean distance can be found in Appendix \ref{finiteDimProofs}. 
\section{Alpha-Procrustes based geometries: infinite dimensional setting}\label{infinitDim}
In this section, we treat the infinite dimensional versions of the aforementioned results. Computing the generalized distances on an infinite dimensional real and separable Hilbert space $\mathcal{H}$ is not all that trivial as one can expect. In this setting we need to: 
\begin{enumerate}
    \item \textit{address with inverse of $\bf{M}$ on $\mathcal{H}$ which manifest itself in the generalized Bures-Wasserstein metric in finite dimensions. This will allow us to consequently extend the definition of $|| \cdot ||_{\mathbf{M}^{-1}}$ accordingly between SPD operators on $\mathcal{H}$;}  

    \item \textit{ensure that the definition of the logarithm of a SPD operator, fits within a \textbf{generalized extended} algebra of Hilbert-Schmidt operators on $\mathcal{H}$ in order to have the correct formulas for the generalized Log-Hilbert-Schmidt geometry.}
\end{enumerate}
It is therefore natural to \textbf{rigorously establish a general formalism of the Mahalanobis norm on $\mathcal{H}$}. This requires an adequate functional analytic framework which we introduce by building upon the works of \citep{Minh2014, MINH202225, Gelbrich1990}. We first recall the analysis in \citep{Minh2014} which institutes the theory of symmetric positive definite operators on infinite dimensional Riemann-Hilbert manifolds.
\subsection{Trace class and Hilbert-Schmidt operators}
Define the set of trace class and Hilbert-Schmidt operators as 
\begin{equation}\label{trace_class}
    \mathcal{S}_{1} := \bigg\{ {\bf{A}} : \vert\vert {\bf{A}} \vert\vert_{\mathrm{1}} = \sum_{i \in [ 1, \mathrm{dim}(\mathcal{H})]} s_{i}({\bf{A}}) < \infty \bigg \},
\end{equation}
where $s_{i}$ are the singular values of ${\bf{A}}$. $\mathcal{S}_{1}$ is complete in the norm $\vert\vert \cdot \vert\vert_{\mathrm{1}}$. The set $\mathcal{S}_{2}$ of Hilbert-Schmidt operators is defined as 
\begin{equation}\label{HS}
    \mathcal{S}_{2} := \bigg\{ {\bf{A}}: \vert\vert {\bf{A}} \vert\vert_{\mathrm{2}} = \bigg(\sum_{i \in [ 1, \mathrm{dim}(\mathcal{H}) ]} s_{i}({\bf{A}})^{2}\bigg)^{1/2} < \infty  \bigg\}, 
\end{equation}
where $ \vert\vert {\bf{A}} \vert\vert_{\mathrm{2}} = (\sum_{i} \lambda_{i}({\bf{A}}^{*}{\bf{A}}))^{1/2}< \infty$ (\textit{i.e}: ${\bf{A}}^{*}{\bf{A}}$ is trace class) with $i \in [ 1, \mathrm{dim}(\mathcal{H}) ]$ and $\mathrm{dim}(\mathcal{H}) = \infty$ in both equations above. The space $\mathcal{S}_{2}$ is also complete in the norm $\vert\vert \cdot \vert\vert_{\mathrm{2}}$. The completeness shows that both $\mathcal{S}_{1}$ and $\mathcal{S}_{2}$ are Banach algebras with the norms $\vert \vert \cdot \vert \vert_{\mathrm{1}}$ and $\vert \vert \cdot \vert \vert_{\mathrm{2}}$ respectively. 
\subsection{Positive definite extended (unitized) Hilbert-Schmidt operators}
For $\mathrm{dim}(\mathcal{H}) = \infty$ the operator ${\bf{A}}$ is not invertible. Further, the Hilbert-Schmidt norm of the identity operator is not controlled. Consequently, to define the Hilbert-Schmidt inner product adequately and achieve stability with respect to the Hilbert-Schmidt norm, the cone of positive invertible Hilbert-Schmidt operators extended by scalar operators on $\mathcal{H}$ is defined as \citep{Larotonda2007, Minh2014}
\begin{equation}\label{HS_ext}
    \mathcal{S}_{2}^{(\mathrm{EXT})} := \{{\bf{A}} + \delta{\bf{I}} : {\bf{A}^{*}} = {\bf{A}},\,\, {\bf{A}} \in \mathcal{S}_{2},\,\, \delta \in \mathbb{R}_{0}^{+} \}.
\end{equation}
The \textbf{extended inner product} denoted by $\langle\,, \, \rangle_{\text{eHS}}$ with respect to the above space is then
\begin{equation}
    \langle {\bf{A}} + \delta{\bf{I}},\,\, {\bf{B}} + \gamma{\bf{I}} \rangle_{\mathrm{eHS}} = \mathrm{tr}({\bf{A}^{*}} {\bf{B}}) + \delta\gamma = \langle {\bf{A}}, {\bf{B}} \rangle_{\mathrm{HS}} + \delta\gamma,
\end{equation}
with the scalar operators orthogonal to the Hilbert-Schmidt operators. Coupled with the lower bound condition of the Eigenvalues of ${\bf{A}}$ established in the previous paragraph, one effectively defines the \textbf{infinite dimensional generalization of $\mathbb{S}^{n}_{++}$} as the \textbf{manifold of positive definite extended (unitized) Hilbert-Schmidt operators}
\begin{equation}\label{HilbertManifold}
    \Sigma_{\mathcal{S}_{2}^{(\mathrm{EXT})}}(\mathcal{H}) := \mathbb{P}(\mathcal{H}) \cap \mathcal{S}_{2}^{(\mathrm{EXT})} =  \{{\bf{A}} + \delta{\bf{I}} > 0 : {\bf{A}^{*}} = {\bf{A}},\,\, {\bf{A}} \in \mathcal{S}_{2},\,\, \delta \in \mathbb{R}_{0}^{+} \}
\end{equation}
In a similar manner, the \textbf{extended trace class} operator is defined as
\begin{equation}\label{trace_class_ext}
    \mathcal{S}_{1}^{(\mathrm{EXT})} := \{{\bf{A}} + \delta {\bf{I}}: {\bf{A}} \in \mathcal{S}_{1}, \delta \in \mathbb{R}_{0}^{+}\}
\end{equation}
and the set of \textbf{positive definite (extended) unitized trace class operators} is defined as 
\begin{equation}\label{traceClassManifold}
    \Sigma_{\mathcal{S}_{1}^{(\mathrm{EXT})}}(\mathcal{H}) := \mathbb{P}(\mathcal{H}) \cap \mathcal{S}_{1}^{(\mathrm{EXT})} =  \{{\bf{A}} + \delta{\bf{I}} > 0 : {\bf{A}^{*}} = {\bf{A}},\,\, {\bf{A}} \in \mathcal{S}_{1},\,\, \delta \in \mathbb{R}_{0}^{+} \}.
\end{equation}
This representation, therefore, ensures that the theorems of linear bounded operators on $\mathcal{H}$ still applies so that the by virtue of the boundedness of the Eigenvalues of ${\bf{A}} + \delta{\bf{I}}$ the limit $k\rightarrow \infty$ of $\lambda_{k} + \delta$ equals to $\delta$ where $\lambda_{k}$ are the Eigenvalues of ${\bf{A}}$. \textbf{This entails that the logarithm of operators with respect to this extended algebra is bounded and well-defined}. We build on this strategy to construct the inverse of positive definite extended operators on $\mathcal{H}$ below. 
\subsection{Inverse of positive definite extended operators on infinite dimensional Hilbert spaces}
Let $\textbf{M} \in \mathcal{S}_{2}(\mathcal{H})$ be compact, self-adjoint, and strictly positive (\textit{i.e.}, $\langle \mathbf{M}x,x\rangle > 0$ for all nonzero $x\in\mathcal{H}$, equivalently $\mathbf{M}$ injective with $\sigma(\mathbf{M})\subset\mathbb{R}_{>0}$ accumulating only at $0$; note that on an infinite-dimensional $\mathcal{H}$ no compact operator can be uniformly bounded below, so $\mathbf{M}$ is not "positive definite" in the bounded-below sense used in finite dimensions). By virtue of standard results of spectral theory we write 
\begin{equation}
    \vert \vert {\bf{M}}^{-1} \vert \vert_{\text{op}} = \sup \{ \omega_{k}^{-1}: \omega_{k}^{-1} \,\,\,\mathrm{are\,\,\,the\,\,\,eigenvalues\,\,\,of\,\,\,} {\bf{M}}^{-1}\}.
\end{equation}
Since $\textbf{M}$ is compact, its eigenvalues satisfy $\omega_{k} \rightarrow 0$ (see Theorem \ref{spectralTheorem}) , so $|| \textbf{M}^{-1} ||_{\textit{op}} = \sup_{k}  \, \omega_{k}^{-1} = \infty$ as  $\omega_{k} \rightarrow 0$ when $k\rightarrow \infty$ on $\mathcal{H}$. The operator $\textbf{M}^{-1}$ is unbounded and does not extend to a bounded operator on all of $\mathcal{H}$. To resolve this, we fix a scalar parameter $\rho \in \mathbb{R}_{>0}$ and replace $\textbf{M}$ by $(\mathbf{M} + \rho\mathbf{I})$. By Lemma \ref{completenessOfMEigvecLemma}, the eigenvectors $\{\psi_{k}\}$ of $\textbf{M}$ form a complete orthonormal system in $\mathcal{H}$, so for every $x\in \mathcal{H}$:
\begin{equation}\label{spec_decomp_inf_dim_M}
   ({\bf{M}} + \rho{\bf{I}} )^{-1}x = \sum_{k = 1}^{\mathrm{dim(\mathcal{H})}}(\omega_{k}({\bf{M}}) + \rho)^{-1}\langle x,  \psi_{k} \rangle \psi_{k}, 
\end{equation}
where $\langle\,, \,\rangle$ corresponds to the scalar inner product on $\mathcal{H}$. Since $\omega_{k} + \rho \geq \rho > 0$ for all $k$, the inverse satisfies $||({\bf{M}} + \rho{\bf{I}})^{-1}||_{\text{op}} \leq \frac{1}{\rho} < \infty$, so $({\bf{M}} + \rho{\bf{I}})^{-1}$ is bounded on all of $\mathcal{H}$ and the series (\ref{spec_decomp_inf_dim_M}) converges in norm. Finally, since $\mathbf{M} \in \mathcal{S}_{2}(\mathcal{H})$ and $\rho \in \mathbb{R}_{>{0}}$, the operator $({\bf{M}} + \rho{\bf{I}})$ belongs to $ \Sigma_{\mathcal{S}_{2}^{(\mathrm{EXT})}}(\mathcal{H})$ by definition of Eq. (\ref{HS_ext}), and is positive definite since $\langle (\textbf{M}+ \rho\textbf{I})x, x\rangle \geq \rho||x||^{2} > 0$ for all nonzero $x\in \mathcal{H}$. \\ 

We denote the operator $({\bf{M}} + \rho{\bf{I}})$ by $\mathbf{M}_{\infty}$ for simplicity in this setting. In the next section we will see how this \textit{extends} the Mahalanobis norm $||\cdot||_{\mathbf{M}^{-1}}$ defined in (\ref{gbwDistance}). In Appendix \ref{extMahalanobisNormProperties} we establish that $||\cdot||_{\mathbf{M}^{-1}}$ is a well-defined norm on $\Sigma_{\mathcal{S}_{2}^{(\mathrm{EXT})}}(\mathcal{H})$ by virtue of Lemma \ref{extMahalanobisNormPropertiesProp}. We call this generalized norm the \textbf{extended Mahalanobis norm} and write it as follows $||\cdot||_{\mathbf{M}^{-1}_{\infty}}$. On $\mathcal{H}$ this generalized norm is \textit{extended} by the new parameter $\rho$. Clearly, this defines a very specific class of \textit{infinite-dimensional generalized alpha-Procrustes} distances parametrized by $\rho$. In what follows, we will \textbf{always} assume $\rho$ to be fixed and only discuss the limit to $0$ such that $\lim_{\rho \rightarrow 0}||\cdot||_{\mathbf{M}^{-1}_{\infty}} = ||\cdot||_{\mathbf{M}^{-1}}$ to show how infinite dimensional distances reduces to their respective finite-dimensional counterpart (\textit{i.e., $\rho = 0$}). In this setting $\rho$ appears to act as a \textbf{regularization parameter} and large values of $\rho$ flattens $({\bf{M}} + \rho{\bf{I}})^{-1}$ in all directions leading to a loss of geometric information. In Section (\ref{genLESRNumRes}) we propose a \textit{truncated generalized log Hilbert-Schmidt} distance and study how effective it is at capturing the difference between complex geometries in the limit $\rho \rightarrow \infty$. A rigorous analysis pertaining to the limit $\rho \rightarrow 0$ and its implications for the specific class of infinite-dimensional distances it generates is left for future work.
\subsection{The infinite dimensional setting}
We are now in a position to introduce the infinite-dimensional representation of the \textbf{generalized alpha-Procrustes distances between positive definite operators on $\Sigma_{\mathcal{S}_{2}^{(\mathrm{EXT})}}(\mathcal{H})$}. We build on the results of \cite{Minh2014, Minh2021, MINH202225} where the infinite dimensional version of the alpha-Procrustes distance is rigorously established on $\mathcal{H}$. We proceed to show how the \textbf{infinite dimensional versions of the GBW distance} (\ref{GBW}), denoted as \textbf{$\mathrm{GBW}_{\infty}$}, can be realised from this generalised class of distance along the lines of ~\cite[Theorem 9]{MINH202225} and ~\cite[Corollary 1]{MINH202225}. Additionally, the \textbf{generalized Log-HS} is introduced and discussed further in the Appendix \ref{infiniteDimProofs}. The \textit{extended} Mahalanobis norm on the manifold of generalized positive definite extended (unitized) Hilbert-Schmidt operators is defined as

\begin{definition}(\textbf{Extended Mahalanobis norm between HS operators})\label{inftyMahalonobisNorm}
    Let $\mathcal{H}$ be a separable infinite-dimensional Hilbert space. Let $\rho \in \mathbb{R}_{>0}$ be fixed. For ${\bf{X}} \in \mathcal{S}_{2}(\mathcal{H})$ and $({\bf{M}} + \rho{\bf{I}}) \in \Sigma_{\mathcal{S}_{2}^{(\mathrm{EXT})}}(\mathcal{H})$ the extended Mahalanobis norm is expressed as
    \begin{equation}\label{inftyMahalonobisNormEq}
        \vert\vert {\bf{X}}\vert\vert_{{\bf{M}}^{-1}_{\infty}} = \sqrt{\mathrm{tr}\{{\bf{X}}^{*} ({\bf{M}} + \rho{\bf{I}})^{-1} {\bf{X}}\}}.
    \end{equation}
        When $\mathrm{dim}(\mathcal{H}) < \infty$ and $\mathbf{M}$ is invertible, the norm $\vert\vert \cdot \vert\vert_{{\bf{M}}^{-1}_{\infty}}$ recovers the classical Mahalanobis norm $\vert\vert \cdot \vert\vert_{{\bf{M}}^{-1}}$ in the limit $\rho \rightarrow 0^{+}$ (in the infinite-dimensional setting the limit is singular since $\mathbf{M}^{-1}$ is unbounded), \textit{i.e.}
    \begin{equation}\label{limitExtMahalanobisNorm}
        \lim_{\rho \rightarrow 0^{+}}\vert\vert {\bf{X}} \vert\vert_{{\bf{M}}^{-1}_{\infty}} = \vert\vert \bf{X} \vert\vert_{{\bf{M}}^{-1}}. 
    \end{equation}
\end{definition}
Below we test this definition to compare how the extended Mahalanobis norm matches its finite dimensional counterpart in the limit $\rho \rightarrow 0$ when dimensions $d$ increases. We see the expected collapse of information when $\rho$ gets moderately large. 
\begin{figure}[H]
    \centering
    \includegraphics[width=0.9\linewidth]{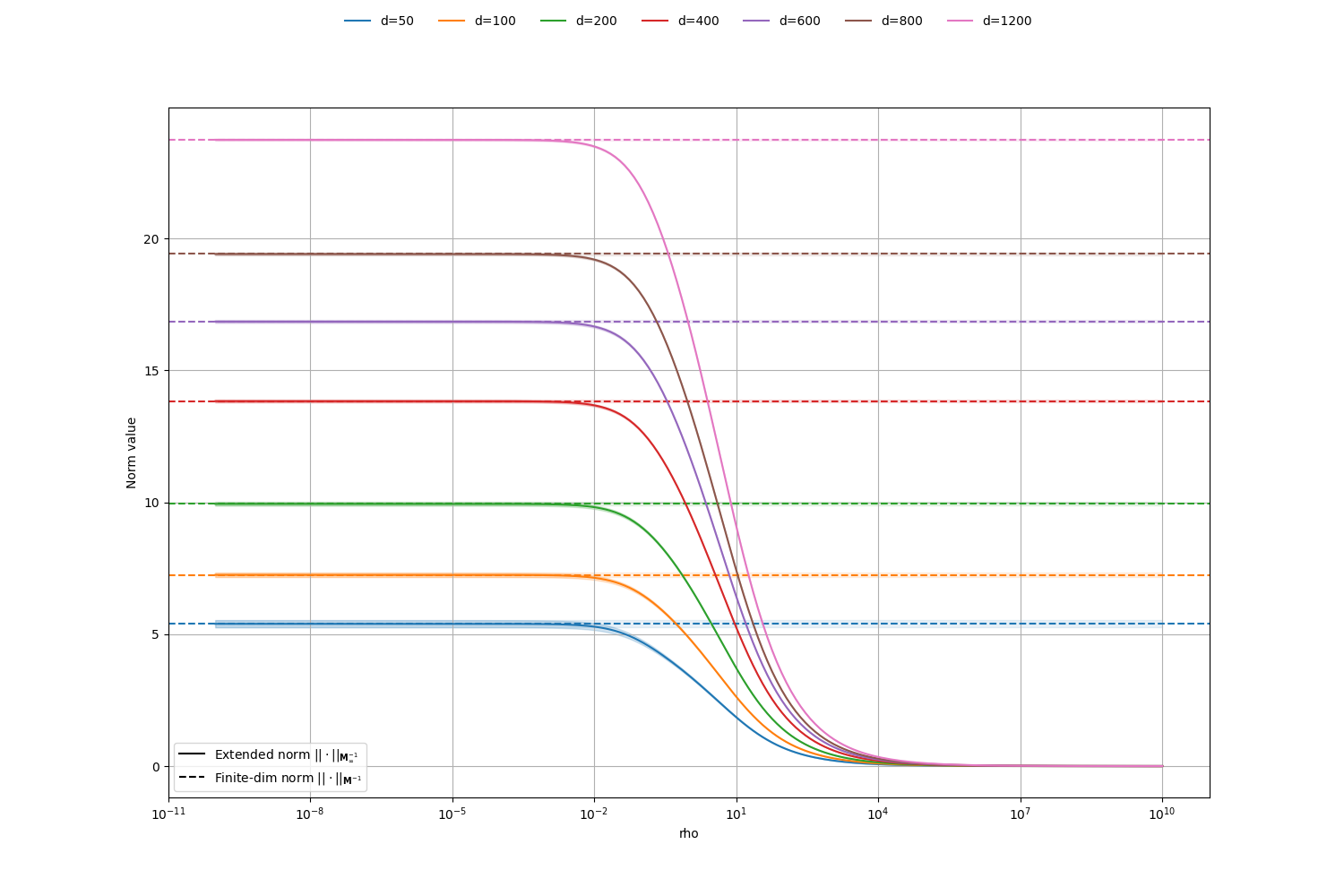}
    \caption{Dependence of the extended Mahalanobis norm on the regularization parameter $\rho$. Small values of $\rho$ preserve fine geometric information, while larger values progressively regularize the geometry by suppressing the contribution of low-energy spectral modes. }
    \label{fig:placeholder}
\end{figure}
\begin{rem}\label{remRhoValsStudy}
    The regularization parameter $\rho$ plays two complementary roles. From a functional-analytic perspective, it guarantees that $(\bf{M} + \rho\bf{I})^{-1}$ is a bounded operator on $\mathcal{H}$, ensuring that the extended Mahalanobis norm is well defined on $\Sigma_{\mathcal{S}_{2}^{(\mathrm{EXT})}}(\mathcal{H})$. From a geometric perspective, $\rho$ controls the trade-off between preserving fine spectral information and suppressing directions associated with small or noisy eigenvalues. As $\rho \rightarrow 0^{+}$, the geometry formally approaches its finite-dimensional counterpart whenever $\bf{M}^{-1}$ exists, whereas increasing $\rho$ progressively smooths the geometry by reducing the influence of poorly conditioned spectral modes. The numerical experiments in Section \ref{genLESRNumRes} illustrate this regularization effect, showing that intermediate values of $\rho$ can improve robustness in high-dimensional settings while excessively large values may attenuate discriminative geometric information.
\end{rem}
We extend Definition \ref{inftyMahalonobisNorm} to extended operators on $\Sigma_{\mathcal{S}_{2}^{(\mathrm{EXT})}}(\mathcal{H})$. 
\begin{definition}(\textbf{Extended Mahalanobis norm between extended HS operators})\label{inftyMahalonobisNormExtHSOps}
    Let $\mathcal{H}$ be a separable infinite-dimensional Hilbert space. Let $\delta \in \mathbb{R}_{>0}$ be fixed. Let $\rho \in \mathbb{R}_{>0}$ be fixed. For $({\bf{X}} + \delta{\bf{I}}) \in \Sigma_{\mathcal{S}_{2}^{(\mathrm{EXT})}}(\mathcal{H})$. The extended Mahalanobis norm is expressed as
    \begin{equation}\label{inftyMahalonobisNormEqExt}
        \vert\vert ({\bf{X}} + \delta{\bf{I}}) \vert\vert_{{\bf{M}}^{-1}_{\infty}} = \sqrt{\mathrm{tr}\{{\bf{X}}^{*} ({\bf{M}} + \rho{\bf{I}})^{-1} {\bf{X}}\} + \delta^{2}}.
    \end{equation}
    When $\mathrm{dim}(\mathcal{H}) < \infty$ and $\mathbf{M}$ is invertible, the norm $\vert\vert \cdot \vert\vert_{{\bf{M}}^{-1}_{\infty}}$ recovers the classical Mahalanobis norm $\vert\vert \cdot \vert\vert_{{\bf{M}}^{-1}}$ in the joint limit $\rho \rightarrow 0^{+}$, $\delta \rightarrow 0^{+}$ (in the infinite-dimensional setting the $\rho \to 0^{+}$ limit is singular), \textit{i.e.}
\begin{equation}\label{limitExtMahalanobisNormExt}
        \lim_{\substack{\rho \rightarrow 0^{+} \\ \delta \rightarrow 0^{+}}}\vert\vert ({\bf{X}} + \delta{\bf{I}}) \vert\vert_{{\bf{M}}^{-1}_{\infty}} = \vert\vert \bf{X} \vert\vert_{{\bf{M}}^{-1}}. 
    \end{equation}
\end{definition}

\begin{proposition}(\textbf{Infinite dimensional generalized Alpha-Procrustes})\label{generalisedAlphaProcrusteInfDim}
    Let $\alpha \in \mathbb{R}_{>0}$ be fixed and $\delta = \gamma > 0 \in \mathbb{R}$. We fix a $\rho \in \mathbb{R}_{>0}$. Let $({\bf{X}} + \delta{\bf{I}}), ({\bf{Y}} + \delta{\bf{I}}) \in \Sigma_{\mathcal{S}_{2}^{(\mathrm{EXT})}}(\mathcal{H})$. Then the infinite dimensional realization of the generalised alpha-Procrustes distance is then
    \begin{equation}
        d^{\alpha}_{\mathrm{ProM_{\infty}}}[({\bf{X}} + \delta{\bf{I}}), ({\bf{Y}} + \delta{\bf{I}})] = \min_{({\bf{I}} + {\bf{U}}) \in \mathbb{U}(\mathcal{H}) \cap \mathcal{S}_{2}^{(\mathrm{EXT})}(\mathcal{H})} \bigg\vert\bigg\vert \frac{({\bf{X}} + \delta{\bf{I}})^{\alpha} - ({\bf{Y}} + \delta{\bf{I}})^{\alpha}({\bf{I}} + {\bf{U}})}{\alpha} \bigg\vert\bigg\vert_{{\bf{M}}_{\infty}^{-1}}.
    \end{equation}
\end{proposition}
See Appendix \ref{propgeneralisedAlphaProcrusteInfDim} for the proof. We consider the case $\delta = \gamma > 0$ for simplicity for the rest of this work. \\ 
\begin{rem}[\textbf{Extended Gaussian convention}]\label{eHSGaussianConvention}
A genuine Borel Gaussian measure on the infinite-dimensional separable Hilbert space $\mathcal{H}$ requires a trace-class covariance, so for $\delta>0$ the shift $\delta{\bf{I}}$ is not itself a covariance of a Borel Gaussian on $\mathcal{H}$. Following \citep{Minh2021, MINH202225}, throughout this work the symbol $\mathcal{N}(0, {\bf{X}} + \delta{\bf{I}})$ denotes the canonical Gaussian element on the extended Hilbert--Schmidt manifold $\Sigma_{\mathcal{S}_{2}^{(\mathrm{EXT})}}(\mathcal{H})$, and all integral/expectation expressions are interpreted via the extended Hilbert--Schmidt inner product: in particular,
$$
    \mathbb{E}_{{\bf{x}}\sim\mathcal{N}(0,{\bf{X}}+\delta{\bf{I}})}\big[\|{\bf{x}}\|^{2}_{{\bf{M}}_{\infty}^{-1}}\big]
    := \mathrm{tr}\big[({\bf{M}}+\rho{\bf{I}})^{-1}{\bf{X}}\big] + \delta^{2},
$$
which is finite because ${\bf{X}}\in\mathcal{S}_{2}(\mathcal{H})$ and $({\bf{M}}+\rho{\bf{I}})^{-1}\in\mathcal{B}(\mathcal{H})$ with $\|({\bf{M}}+\rho{\bf{I}})^{-1}\|_{\mathrm{op}}\le 1/\rho$, and the operator-ideal property $\mathcal{S}_{2}\cdot\mathcal{B}(\mathcal{H})\cdot\mathcal{S}_{2}\subseteq\mathcal{S}_{1}$ makes the trace well-defined. When ${\bf{X}}+\delta{\bf{I}}$ is genuinely trace-class (e.g.\ $\delta=0$ and ${\bf{X}}\in\mathcal{S}_{1}(\mathcal{H})$), this convention reduces to the classical Borel Gaussian expectation.
\end{rem}
\begin{corollary}(\textbf{Infinite dimensional generalised Bures-Wasserstein distance})\label{corrInfAlphaGBW}
    Let $\alpha = 1/2$. Let $\delta = \gamma > 0$ and $\rho \in \mathbb{R}_{>0}$. Let $({\bf{X}} + \delta{\bf{I}}), ({\bf{Y}} + \delta{\bf{I}}) \in \Sigma_{\mathcal{S}_{2}^{(\mathrm{EXT})}}(\mathcal{H})$. Then the infinite dimensional generalised Bures-Wasserstein distance is realised as
    \begin{equation}        d^{1/2}_{\mathrm{ProM_{\infty}}}[({\bf{X}} + \delta{\bf{I}}), ({\bf{Y}} + \delta{\bf{I}})] = 2d_{\mathrm{GBW}_{\infty}}[({\bf{X}} + \delta{\bf{I}}), ({\bf{Y}} + \delta{\bf{I}})].
    \end{equation}
\end{corollary}
\begin{proof}
    Setting $\alpha= 1/2$ in Proposition \ref{generalisedAlphaProcrusteInfDim} and expanding
    \begin{equation*}
        \min_{({\bf{I}} + {\bf{U}}) \in \mathbb{U}(\mathcal{H}) \cap \mathcal{S}_{2}^{(\mathrm{EXT})}(\mathcal{H})}    
        \bigg\vert\bigg\vert \frac{({\bf{X}} + \delta{\bf{I}})^{\alpha} - ({\bf{Y}} + \delta{\bf{I}})^{\alpha}({\bf{I}} + {\bf{U}})}{\alpha} \bigg\vert\bigg\vert_{{\bf{M}}_{\infty}^{-1}}^{2}
    \end{equation*}
    leads to the $\mathrm{GBW}_{\infty}$ between the scaled infinite dimensional operators. \\
    
    Explicitly, in the eHS convention of Remark \ref{eHSGaussianConvention},
    \begin{equation}\label{gbwInftyExplicit}
        d^{2}_{\mathrm{GBW}_{\infty}}[({\bf{X}}+\delta{\bf{I}}),({\bf{Y}}+\delta{\bf{I}})]
        = \mathrm{tr}[({\bf{M}}+\rho{\bf{I}})^{-1}{\bf{X}}] + \mathrm{tr}[({\bf{M}}+\rho{\bf{I}})^{-1}{\bf{Y}}] + 2\delta^{2} - 2\,\mathrm{tr}\,F_{\infty}({\bf{X}}+\delta{\bf{I}},{\bf{Y}}+\delta{\bf{I}}),
    \end{equation}
    where $F_{\infty}$ is the eHS-extended fidelity functional introduced in Appendix \ref{proofinfDimW2}; reduction to $\delta=0$, ${\bf{M}}={\bf{I}}$, ${\bf{X}},{\bf{Y}}\in\mathcal{S}_{1}(\mathcal{H})$ recovers the classical infinite-dimensional Bures-Wasserstein formula of \citep{Minh2021}.
\end{proof}
We proceed to show the connection between $\mathrm{GBW}_{\infty}$ and the infinite dimensional generalized Wasserstein distance denoted here as $\widetilde{\mathcal{W}}_{2, \infty}$. This is motivated by the fact that this connection is rigorously established in ~\cite[Theorem 9]{Minh2021} between the infinite dimensional Bures-Wasserstein and the $2$-Wasserstein distance on $\mathcal{H}$, $\mathcal{W}_{2, \infty}$. This result rely on ~\cite[Theorem 3.5]{Gelbrich1990}. Furthermore, the authors in \citep{HMJG2021b} showed how the GBW is related to the generalized $2$-Wasserstein distance with the cost function $d(x, y) = \vert\vert x-y \vert\vert_{\mathbf{M}^{-1}}$ (see \ref{W_2Distance}). By employing the aforementioned results we generalize the argument of ~\cite[Proposition 3]{HMJG2021b} to the following (see Appendix \ref{proofinfDimW2} for the proof):
\begin{corollary}\label{infDimW2}(\textbf{Infinite dimensional generalized Wasserstein distance})
Let $\delta = \gamma > 0$ and $\rho \in \mathbb{R}_{>0}$. Let $\mathcal{B}(\mathcal{H})$ denote the Borel $\sigma$-algebra on $\mathcal{H}$ and $\mathcal{P}(\mathcal{H})$ the space of probability measures on $\mathcal{H}$. Define the zero mean non-degenerate Gaussian probability measures $\mu = \mathcal{N}(0, \bf{X} + \delta\bf{I})$ and $\nu = \mathcal{N}(0, \bf{Y} + \delta\bf{I})$ in the eHS sense of Remark \ref{eHSGaussianConvention}, with covariance operators as $({\bf{X}} + \delta{\bf{I}}), ({\bf{Y}} + \delta{\bf{I}}) \in \Sigma_{\mathcal{S}_{2}^{(\mathrm{EXT})}}(\mathcal{H})$. Then, for any $({\bf{M}} + \rho{\bf{I}})^{-1} \in \Sigma_{\mathcal{S}_{2}^{(\mathrm{EXT})}}(\mathcal{H})$, we have $ \widetilde{\mathcal{W}}^{2}_{2, \infty}(\mu, \nu) = d^{2}_{\mathrm{GBW}_{\infty}}[({\bf{X}} + \delta{\bf{I}}), ({\bf{Y}} + \delta{\bf{I}})]$.
\end{corollary}
The generalized version of the Log-Hilbert-Schmidt (Log-HS) distance is also be realised from Proposition (\ref{generalisedAlphaProcrusteInfDim}) for a fixed $\rho \in \mathbb{R}$ as follows (see Appendix \ref{proofCorrInfGenLogHS} for the proof):
\begin{corollary}(\textbf{Infinite dimensional generalized Log Hilbert-Schmidt distance})\label{corrInfGenLogHS}
Let $\alpha \in \mathbb{R}_{>0}$ be fixed, $\delta = \gamma > 0$ and $\rho \in \mathbb{R}_{>0}$.
Let $({\bf{X}} + \delta{\bf{I}}), ({\bf{Y}} + \delta{\bf{I}}) \in \Sigma_{\mathcal{S}_{2}^{(\mathrm{EXT})}}(\mathcal{H})$. The condition $\delta>0$ guarantees that $\log({\bf{X}}+\delta{\bf{I}})$ and $\log({\bf{Y}}+\delta{\bf{I}})$ are well-defined bounded operators since ${\bf{X}}+\delta{\bf{I}}$ is bounded below by $\delta\,\mathbf{I}$. The limit $\alpha \rightarrow 0^{+}$ is 
\begin{equation}\label{infDimGenLHSEq}
    \lim_{\alpha \rightarrow 0^{+}}d_{\mathrm{proM_{\infty}}}^{\alpha}[({\bf{X}} + \delta{\bf{I}}), ({\bf{Y}} + \delta{\bf{I}})] = \vert\vert \log ({\bf{X}} + \delta{\bf{I}}) - \log ({\bf{Y}} + \delta{\bf{I}}) \vert\vert_{{\bf{M}}^{-1}_{\infty}}.
\end{equation}
\end{corollary}

\begin{center}
    \begin{table}[H]
    \centering
    \caption{Distance values and parameters. }
    \begin{tabular}{||c c c c||} 
         \hline
         \textbf{Distance recovered} & \textbf{Parameter setting}  & & \\ [0.5ex] 
         \hline\hline
         & & &\\
         
         Bures-Wasserstein (BW) & $\alpha = \frac{1}{2}$ &  &\\ 
         & $\bf{M} = \bf{I}$ & &\\
         & $\delta = 0, \rho = 0$ & &\\
         & Norm: $||\cdot||_{\text{F}}$& &\\
         & & &\\
         
         Generalized BW & $\alpha = \frac{1}{2}$ & &\\
         & $\bf{M} \in \mathbb{S}^{n}_{++}$ arbitrary & &\\
         & $\delta =0 , \rho =0$  & &\\
         & Norm: $||\cdot||_{\bf{M}^{-1}}$ & &\\
         & & &\\
         
         Log-Euclidean & $\alpha \rightarrow 0^{+}$ & &\\
         & $\bf{M} = \bf{I}$ & &\\
         & $\delta = 0, \rho = 0$& &\\
         & & &\\
         
         Generalized Log-Euclidean & $\alpha \rightarrow 0^{+}$ & & \\
         & $\bf{M} \in \mathbb{S}^{n}_{++}$ arbitrary & &\\
         & $\delta =0 , \rho =0$& &\\
         & Norm: $||\cdot||_{\bf{M}^{-1}}$& &\\
         & & &\\
         
         $\text{GBW}_{\infty}$ & $\alpha = \frac{1}{2}$ & &\\ 
         & $\bf{M} \in \sum_{\mathcal{S}_{2}^{\text{(EXT)}}}(\mathcal{H})$ & &\\
         & $\delta > 0$ (unitization) &  &\\
         & $\rho \in \mathbb{R}_{>0}$ (regularization) & &\\
         & Norm: $||\cdot||_{\bf{M}_{\infty}^{-1}}$& &\\
         & & &\\
         
         Generalized 2-Wasserstein $\widetilde{W}_{2, \infty}$ & $\alpha=\frac{1}{2}$ (\textit{via} $\text{GBW}_{\infty}$) & &\\
         & $\bf{M} \in \sum_{\mathcal{S}_{2}^{\text{(EXT)}}}(\mathcal{H})$ & &\\
         & $\delta> 0$ (unitization) & &\\
         & $\rho \in \mathbb{R}_{>0}$ (regularization)& &\\
         & Norm: $||\cdot||_{\bf{M}_{\infty}^{-1}}$& &\\
         & Cost: $||\bf{x} - \bf{y}||_{\bf{M}_{\infty}^{-1}}$ & &\\
         & & &\\
         
         Generalized LES & $\alpha \rightarrow 0^{+}$ & & \\ 
         & $\bf{M} \in \sum_{\mathcal{S}_{2}^{\text{(EXT)}}}(\mathcal{H})$ & &\\
         & Top-$K$ eigenvalues of $\bf{M}$, $\{\omega_{i}\}_{i=1}^{K}$ & &\\
         & $ \delta > 0$ (log-stability)& &\\
         & $ \rho >0$ (weighted reg. from $(\bf{M} + \rho \bf{I})^{-1}$& &\\
         & Norm: $||\cdot||_{\bf{M}_{\infty}^{-1}}$& &\\
         & $\bf{M}$ is random init. or learned & &\\
         & & &\\ [1ex] 
         \hline
    \end{tabular}
    \label{distValuesAndParams}
    \end{table}
\end{center}
\section{Applications of the infinite dimensional generalized alpha-Procrustes geometries}\label{applications}
\subsection{The infinite dimensional robust GBW metric}
%
%
%
%
% The Subspace Robust Wasserstein (SRW) distance is an extension of the Wasserstein distance that improves its robustness by focusing on important subspaces. Instead of comparing two distributions in the full-dimensional space, SRW identifies and aligns the most important low-dimensional subspaces (based on variance or energy) and computes the Wasserstein distance within these subspaces.
%
%
%
%
% The Projection Robust Wasserstein (PRW) distance is a variant of SRW that further enhances robustness by optimally selecting subspaces where the distributions are compared. Instead of projecting data onto arbitrary subspaces, PRW searches for the subspace that maximizes the Wasserstein distance between the two distributions.
%
%
%
%
% Infinite-Dimensional Applications
%   Functional Data Analysis: The infinite-dimensional extension of the GBW distance can be applied to functional data analysis, where data are modeled as functions rather than vectors. This is especially relevant in fields like weather prediction, economics, or biomechanics.
%   Operator-Valued Kernels: In operator-theoretic machine learning, kernels can be represented by positive semidefinite operators in infinite dimensions. The GBW distance provides a robust way to measure distances between such kernels or operators, leading to more effective learning algorithms.
%
%
%
%
Handling high-dimensional data where irrelevant features and/or noise are present can make traditional Wasserstein distances less reliable. Those issues were accordingly addressed in \citep{paty2019, huang2021} by virtue of Riemannian optimization techniques which consequently established a new class Wasserstein distances called robust Wasserstein distances. The latter was naturally extended to the robust generalized Bures-Wasserstein (RGBW) in \citep{HMJG2021b}. Thanks to the equivalence established in Corollary \ref{infDimW2}, we proceed to rigorously establish the infinite dimensional version of the RGBW metric which we denote here as $\mathrm{RGBW}_{\infty}$. 
\begin{proposition}(\textbf{Infinite dimensional robust generalized Bures-Wasserstein distance})\label{infDimRGBW}
Let $\delta = \gamma > 0$ and $\rho \in \mathbb{R}_{>0}$. 
Let $\mu, \nu \in \mathcal{P}(\mathcal{H}) $ be zero-centered Gaussians measures (in the eHS sense of Remark \ref{eHSGaussianConvention}) with covariance operators $({\bf{X}} + \delta{\bf{I}}), ({\bf{Y}} + \delta{\bf{I}}) \in \Sigma_{\mathcal{S}_{2}^{(\mathrm{EXT})}}(\mathcal{H})$. Let $\Pi \in \mathcal{B}(\mathcal{H})$ be a bounded operator (representing a generalized partial isometry from $\mathcal{H}$ onto a closed subspace; the symmetric, positive-definite operator $\Pi\Pi^{*}\in\Sigma_{\mathcal{S}_{2}^{(\mathrm{EXT})}}(\mathcal{H})$ plays the role of $(\mathbf{M}+\rho\mathbf{I})^{-1}$). There exists a maximizer $\Pi^{\star}\in\mathcal{B}(\mathcal{H})$ with $\Pi^{\star}(\Pi^{\star})^{*}\in\widetilde{\mathcal{C}}:=\{\Pi\Pi^{*} : \Pi \in \mathcal{B}(\mathcal{H}),\ \Pi\Pi^{*}\in\Sigma_{\mathcal{S}_{2}^{(\mathrm{EXT})}}(\mathcal{H})\}$ such that 
\begin{equation}\label{infDimPRW}
    \mathfrak{P}_{\widetilde{\mathcal{C}}}(\mu, \nu) = \sup_{\Pi \in \mathcal{B}(\mathcal{H}),\,\Pi\Pi^{*}\in\widetilde{\mathcal{C}}} \inf_{\gamma \sim \Gamma(\mu, \nu)}\int ||\Pi(x -y) ||^{2}\,d\gamma(x, y).
\end{equation}

If $\Pi^{\star}$ is an optimal solution of (\ref{infDimPRW}) then one has that $\mathfrak{P}_{\widetilde{\mathcal{C}}}(\mu, \nu) = d^{2}_{\mathrm{GBW}_{\infty}}[({\bf{X}} + \delta{\bf{I}}), ({\bf{Y}} + \delta{\bf{I}})]$ for $({\bf{M}} + \rho{\bf{I}})^{-1} = \Pi^{\star}(\Pi^{\star})^{*}$(a symmetric positive operator of the form $PP^{*}$ with $P=\Pi^{\star}$). It follows that
\begin{equation}\label{rgbwInftyGbwInfty}
    d^{2}_{\mathrm{RGBW}_{\infty}}[({\bf{X}} + \delta{\bf{I}}), ({\bf{Y}} + \delta{\bf{I}})] = \max_{({\bf{M}} + \rho{\bf{I}})^{-1} \in \Tilde{\mathcal{C}}} d^{2}_{\mathrm{GBW}_{\infty}}[({\bf{X}} + \delta{\bf{I}}), ({\bf{Y}} + \delta{\bf{I}})] 
\end{equation}
for $\widetilde{\mathcal{C}} \subseteq \Sigma_{\mathcal{S}_{2}^{(\mathrm{EXT})}}(\mathcal{H})$.
\end{proposition}
See Appendix \ref{proofinfDimRGBW} for the proof and Appendix \ref{proofLowerSemCont} along with further details related to
$d^{2}_{\mathrm{RGBW}}$ (\textit{i.e.}, the finite dimensional version from \citep{HMJG2021b}).

\subsection{Spectrum truncation and estimation with the generalized Log-HS metric}\label{genLESRNumRes}

The measurement of distances between datasets that do not share a common reference frame is introduced in the paper \citep{shnitzer2022}. The core strategy leverages \textbf{Log-Euclidean Signatures (LES)} to define intrinsic distances between datasets, allowing for robust comparisons even when they are misaligned or undergo deformations. In particular, symmetric positive definite (SPD) matrices from the dataset are constructed using diffusion maps. The eigenvalues of the associated operators encode geometric and structural properties of the data. 
We find that a generalized framework provides an improved regularization term of said eigenvalues and consequently of the LES thus enabling better convergence of the LES distance \citep{shnitzer2022}. Therefore this allows us to compare datasets in a way that is invariant to transformations and even more robust to deformations. By working in the generalized Log-Hilbert-Schmidt space, the method ensures that distances remain well-defined and meaningful, particularly for high-dimensional or structured data, making it useful for tasks like shape analysis, manifold learning, and intrinsic dataset comparisons.
\begin{proposition}\label{generalisedLogEUS}(\textbf{Truncated generalized Log-Euclidean Signature distance}) Let $\delta \in \mathbb{R}_{>0}$ and $\rho \in \mathbb{R}_{>0}$ be fixed. Let $({\bf{X}} + \delta{\bf{I}}), ({\bf{Y}} + \delta{\bf{I}}) \in \Sigma_{\mathcal{S}_{2}^{(\mathrm{EXT})}}(\mathcal{H})$. The generalized Log-HS distance (see Corollary \ref{corrInfGenLogHS}) is defined as: 
\begin{equation}
    d_{\mathrm{GlogHS}_{\infty}}[({\bf{X}} + \delta{\bf{I}}), ({\bf{Y}} + \delta{\bf{I}})] = \vert\vert \log ({\bf{X}} + \delta{\bf{I}}) - \log ({\bf{Y}} + \delta{\bf{I}}) \vert\vert_{{\bf{M}}^{-1}_{\infty}}, 
\end{equation}
where by $\mathrm{GlogHS}_{\infty}$ we mean the generalized log-HS distance with respect to the norm ${\bf{M}}_{\infty}^{-1}$. \\ 

By treating ${\bf{X}}$ and ${\bf{Y}}$ as diffusion operators constructed in the same way as operators in \citep[Eq.~(4)]{shnitzer2022}, we approximate the $K$ leading eigenvalues $\{\lambda_{i}^{({\bf{X}})}\}_{i=1}^{K}$ and $\{\lambda_{i}^{({\bf{Y}})}\}_{i=1}^{K}$, respectively for each operator. Assuming that $\mathbf{M}$ is spectrally aligned with $\mathbf{X}$ and $\mathbf{Y}$, jointly diagonalizing the Mahalanobis-weighted squared norm in Corollary \ref{corrInfGenLogHS} on the top-$K$ eigenspace yields the \textbf{generalized LES distance}:
\begin{equation}
    d^{2}_{\mathrm{GlogHS}_{\infty}}[({\bf{X}} + \delta{\bf{I}}), ({\bf{Y}} + \delta{\bf{I}})] := \sum_{i=1}^{K}\frac{1}{\omega_{i}^{(\mathbf{M})} + \rho}\big[\log\big(\lambda_{i}^{({\bf{X}})} + \delta\big) - \log\big(\lambda_{i}^{({\bf{Y}})} + \delta\big)\big]^{2},
\end{equation}
where $\omega_{i}$ are the eigenvalues of ${\bf{M}}$. 
\end{proposition}
As reported in \citep{HMJG2021b}, learning of $\bf{M}$ achieves better modeling
in various applications. In the previous sections of this work we propose how to generalise this argument to infinite dimensions and Proposition \ref{generalisedLogEUS} introduces a way to deal with noise in a geometry characterized by $\bf{M}^{-1}_{\infty}$ through the $\rho$-regularized prefactor above. By virtue of theorems on eigenvalues approximation, we perform experiments in order to benchmark our framework against the results in \citep{shnitzer2022} for various values of $\rho$ by first initializing $\bf{\widetilde{M}}$ randomly and secondly using a learning framework. Here, $\bf{\widetilde{M}}$ represents the truncated $\bf{M}^{-1}_{\infty}$ up to the top $K$ eigenvalues.
\subsubsection{Numerical results}
To evaluate the proposed generalized Log-Hilbert-Schmidt (GLES) distance, we reproduce the synthetic benchmark of \cite{shnitzer2022} using point clouds sampled from 2D and 3D tori. Four datasets are considered: $T_2$, $T_2^{Sc}$, $T_3$, and $T_3^{Sc}$, where the superscript $Sc$ denotes scaling of the minor radius by a factor $c$. As $c\rightarrow0$, the corresponding geometries become progressively more distinct, providing a controlled benchmark for evaluating whether a distance correctly captures intrinsic geometric relationships. \\

For all experiments we fix $\delta=\gamma=10^{-8}$. The original LES corresponds to the special case $M=I$ and $\rho\rightarrow0$, whereas the proposed generalized framework introduces the weighted Mahalanobis geometry together with the regularization parameter $\rho$. \\

\begin{figure}[H]
    \centering
    \includegraphics[width=0.8\linewidth]{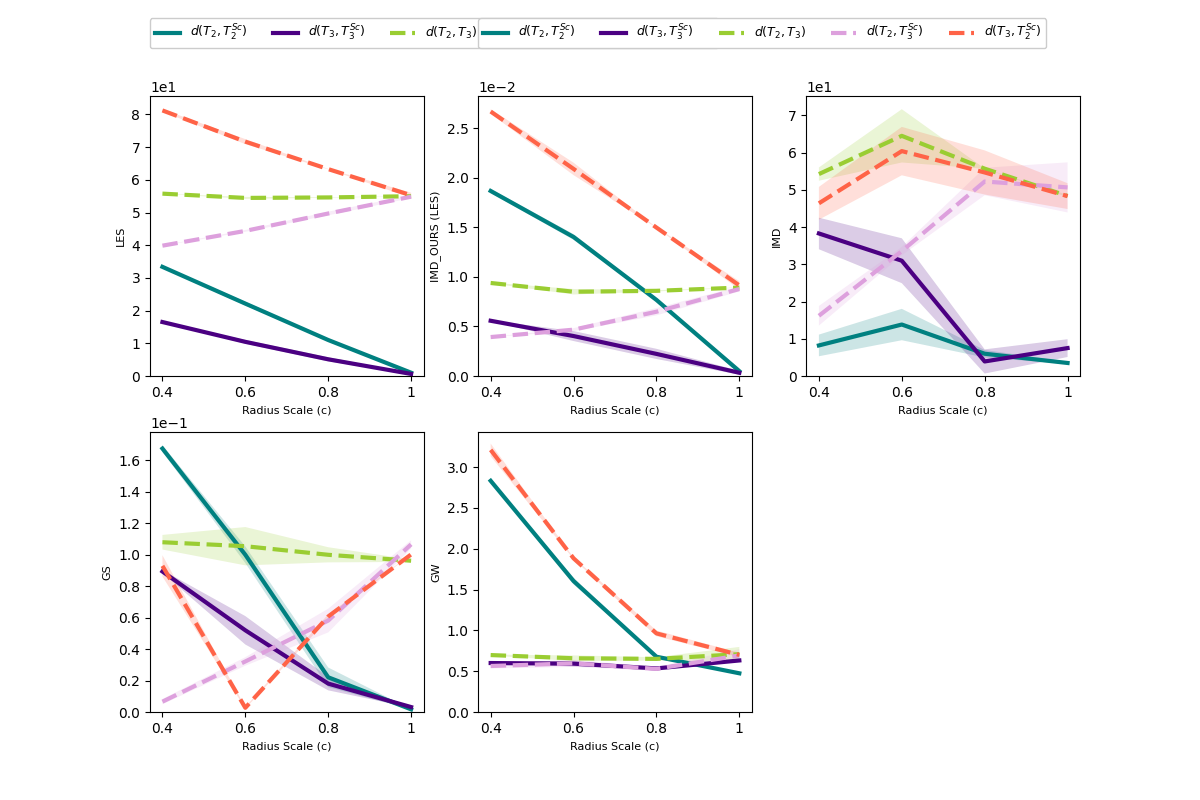}
    \caption{Comparison and evaluation of \textbf{LES distance} against IMD, GS and GW on 2D and 3D tori scaled by factor $c$ for $N=2000$ (number of eigenvalues estimated $= 200$). As $c \rightarrow 0$,  $d(T_{2}, T_{2}^{Sc}), \, d(T_{3}, T_{3}^{Sc})$ and $d(T_{3}, T_{2}^{Sc})$ increases indicating the discrepancy between geometries of different dimensions. On the other hand $d(T_{2}, T_{3}^{Sc})$ decreases since $\lim_{c\rightarrow0}T_{3}^{Sc} = T_{2}$ 
    \citep{shnitzer2022}. As reported in the aforementioned paper, \texttt{IMD\_OURS} with the LES descriptors is less sensitive than LES with respect to the scaling in $c$ as indicated by the intersection of the lines for $d(T_{2}, T_{3}^{Sc})$ and $d(T_{2}, T_{2}^{Sc})$ for $c$ close to $1$.} 
    \label{figLES}
\end{figure}

Figure \ref{figLES} reproduces the original LES benchmark. Consistent with the results reported by \cite{shnitzer2022}, LES correctly captures the expected ordering between geometries of different intrinsic dimensions while remaining computationally efficient.\\

\begin{figure}[H]
    \centering
    \includegraphics[width=0.8\linewidth]{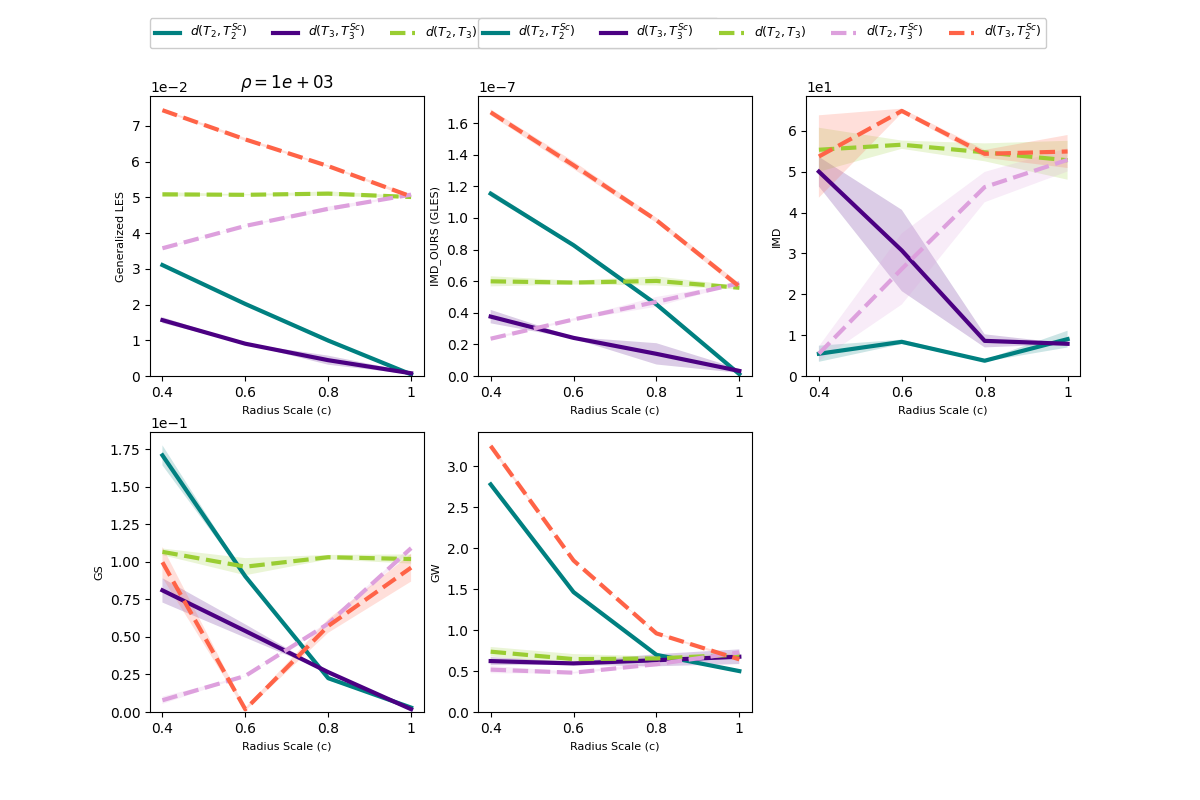}
    \caption{Comparison and evaluation of \textbf{generalized LES distance} against IMD, GS and GW on 2D and 3D tori scaled by factor $c$ for $N=2000$ and $\rho = 1.0 \times 10^{4}$ (number of eigenvalues estimated $= 200$). In the above we observe a similar trend as with LES in Figure \ref{figLES}. The distances with respect to GLES are overall \textbf{lower and smoother} as expected. At $N=2000$ we observe that the method seems less sensitive, the differences between tori are smaller, and there’s less contrast. However, the GLES performs much better for smaller values of $N$ and $\rho = 1.0 \times 10^{2}$ (see experiments in \ref{generalisedLESDetailsResults}).}
    \label{figGenLES}
\end{figure}

Figure \ref{figGenLES} presents the corresponding results obtained with the generalized LES (GLES). The overall geometric relationships are preserved, demonstrating that the generalized framework remains faithful to the original LES geometry. At the same time, the distances become smoother and exhibit reduced variability. This behavior is expected from the theoretical construction developed in Section~3: the operator $(\bf{M}+\rho \bf{I})^{-1}$ down-weights spectral directions associated with small or noisy eigenvalues, leading to a regularized geometry that is less sensitive to perturbations while preserving the dominant spectral structure.\\

We investigated several values of the regularization parameter $\rho$ (see Appendix \ref{generalisedLESDetailsResults}). Small values of $\rho$ recover behavior close to the original LES, whereas increasing $\rho$ progressively regularizes the geometry. Moderate values provide a good compromise between robustness and geometric discrimination, while excessively large values oversmooth the spectrum, reducing sensitivity to fine geometric differences. This behavior is consistent with the interpretation of $\rho$ as a spectral regularization parameter introduced in Definition \ref{inftyMahalonobisNorm} and discussed in Remark \ref{remRhoValsStudy}.\\

Although the generalized distance may slightly reduce discrimination for very large sample sizes, it exhibits improved robustness in regimes where the estimated spectra are more susceptible to sampling noise. Consequently, GLES provides a principled extension of LES that remains well-defined in the infinite-dimensional setting while introducing an explicit mechanism for controlling the stability of the induced geometry.\\ 

\begin{figure}[H]
    \centering
    \includegraphics[width=0.8\linewidth]{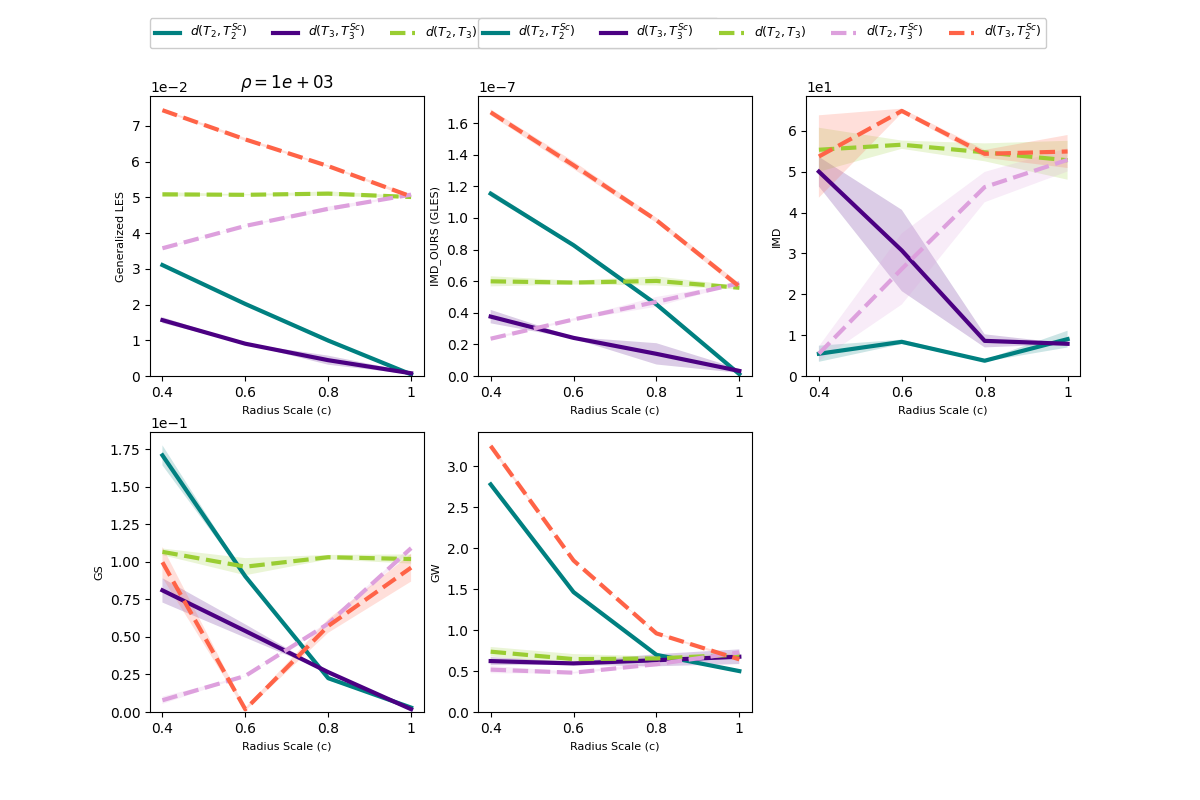}
    \caption{Comparison and evaluation of \textbf{generalized LES distance with $\mathbf{\widetilde{M}}$ optimized} against IMD, GS and GW on 2D and 3D tori scaled by factor $c$ for $N=2000$ and $\rho = 1.0 \times 10^{4}$ (number of eigenvalues estimated $= 200$). In the above we learn $\mathbf{\widetilde{M}}$ for cases where $K = 200, 500, 1000$. The distances with respect to an optimized GLES are \textbf{lower and smoother} (than LES) and also more sensitive to scaling with respect to different dimensions as seen is the separation of the lines for $d(T_{2}, T_{2}^{Sc})$ and $d(T_{3}, T_{3}^{Sc})$, which GLES (Figure \ref{figGenLES}) does not capture at large $N$.}
    \label{figGENLesOpt}
\end{figure}

Finally, Figure \ref{figGENLesOpt} considers the case where the truncated Mahalanobis operator $\widetilde{\bf{M}}$ is learned rather than randomly initialized. Learning $\widetilde{\bf{M}}$ allows the metric to adaptively re-weight the spectral directions that are most informative for the data. Compared with the randomly initialized case, the learned geometry produces better separation between geometries of different intrinsic dimensions while preserving the regularizing effect induced by $\rho$. This observation is consistent with the theoretical interpretation of the generalized Mahalanobis norm as an anisotropic metric that emphasizes informative directions while suppressing noisy ones. Consequently, the learned generalized LES provides a more faithful representation of the underlying geometry than either the original LES or the randomly weighted generalized metric.
Learning the Mahalanobis operator $\widetilde{\bf{M}}$ introduces an adaptive geometry that emphasizes informative spectral directions while suppressing less reliable ones. When the eigenspaces of $\widetilde{\bf{M}}$ and the diffusion operator are well aligned, the resulting geometry is both stable and interpretable. Conversely, poor spectral alignment may overweight directions carrying little geometric information, potentially reducing discriminative power. The optimization of $\widetilde{\bf{M}}$ therefore seeks a balance between robustness and sensitivity, allowing the generalized LES to adapt its geometry to the data. A theoretical characterization of this alignment, for example through perturbation analysis of spectral subspaces, remains an interesting direction for future work.
\section{Discussions}\label{discussions}
This work extends the recently introduced Alpha-Procrustes family of Riemannian metrics for symmetric positive definite (SPD) matrices to the infinite-dimensional setting by developing a rigorous functional analytic framework. Leveraging the structure of extended (unitized) Hilbert-Schmidt operators, we define infinite-dimensional analogues of important geometries such as the generalized Bures-Wasserstein (GBW) and Log-Hilbert-Schmidt (Log-HS) distances. \textbf{A central result is the introduction of the generalized $\alpha$-Procrustes distance on infinite-dimensional SPD operators, equipped with a Mahalanobis-type norm} defined via a compact, positive-definite covariance operator $\mathbf{M}$. This allows for the formulation of robust and geometry-aware distances that remain well-defined even when the underlying operators are not trace-class or when their spectrum is unbounded. We show that:

\begin{itemize}
    \item The infinite-dimensional GBW distance arises as a special case ($\alpha = 1/2$) of the generalized $\alpha$-Procrustes metric.

    \item The generalized Log-HS distance emerges naturally as the limit as $\alpha \rightarrow 0$.

    \item  A regularization parameter $\rho > 0$ provides numerical and geometric stability by ensuring invertibility and spectral control of operators. 

    \item The framework supports spectrum truncation, enabling computational tractability and robustness in high-dimensional or kernel-based settings.
\end{itemize}

Applications include robust comparisons of structured data such as diffusion operators, covariance kernels, or Gaussian processes—particularly when the data lie in different or infinite-dimensional ambient spaces. This is especially relevant for tasks in functional data analysis, manifold learning, and geometry-aware machine learning. From a theoretical standpoint, we push this framework by studying and establishing some initial convergence results in Appendix \ref{firstConvergenceStudies}. Finally, an interesting direction would be to use this framework for the construction and definition of novel embedding spaces for \textbf{graph neural networks} as briefly exposed in Appendix \ref{functionalGNNS}. Overall, this work provides both a theoretical foundation and practical tools for extending Procrustes-type geometries to infinite-dimensional settings, offering a unified and flexible approach to comparing positive operators beyond traditional finite-dimensional domains.

\bibliography{bibliography.bib}

@article{MINH202225,
title = {{Alpha Procrustes metrics between positive definite operators: A unifying formulation for the Bures-Wasserstein and Log-Euclidean/Log-Hilbert-Schmidt metrics}},
journal = {Linear Algebra and its Applications},
volume = {636},
pages = {25-68},
year = {2022},
issn = {0024-3795},
doi = {https://doi.org/10.1016/j.laa.2021.11.011},
url = {https://www.sciencedirect.com/science/article/pii/S0024379521004110},
author = {Ha Quang, M.},
}

@article{BJL2019,
title = {{On the Bures-Wasserstein distance between positive definite matrices}},
journal = {Expositiones Mathematicae},
volume = {37},
number = {2},
pages = {165-191},
year = {2019},
issn = {0723-0869},
doi = {https://doi.org/10.1016/j.exmath.2018.01.002},
url = {https://www.sciencedirect.com/science/article/pii/S0723086918300021},
author = {R. Bhatia and T. Jain and Y. Lim},
}

@inproceedings{HMJG2021a,
 author = {Han, A. and Mishra, B. and Jawanpuria, P. K. and Gao, J.},
 booktitle = {Advances in Neural Information Processing Systems},
 editor = {M. Ranzato and A. Beygelzimer and Y. Dauphin and P.S. Liang and J. Wortman Vaughan},
 pages = {8940--8953},
 publisher = {Curran Associates, Inc.},
 title = {{On Riemannian Optimization over Positive Definite Matrices with the Bures-Wasserstein Geometry}},
 url = {https://proceedings.neurips.cc/paper/2021/file/4b04b0dcd2ade339a3d7ce13252a29d4-Paper.pdf},
 volume = {34},
 year = {2021}
}

@article{HMJG2021b,
author = {Han, A. and Mishra, B. and Jawanpuria, P. K. and Gao, J.},
title = {{Generalized Bures-Wasserstein Geometry for Positive Definite Matrices}},
volume = {},
pages = {},
year = {2021},
language = {},
journal = {arXiv:2110.10464}
}

@book{villani2009,
  title={{Optimal transport: old and new}},
  author={Villani, C.},
  volume={338},
  year={2009},
  publisher={Springer}
}

@article{Gelbrich1990,
  title={{On a Formula for the $L^{2}$-Wasserstein metric between measures on Euclidean and Hilbert spaces}},
  author={M. Gelbrich},
  journal={Mathematische Nachrichten},
  year={1990},
  volume={147},
  pages={185-203}
}

@inproceedings{Minh2014,
 author = {Ha Quang, M. and Sam Biagio, M. and Murino, V.},
 booktitle = {Advances in Neural Information Processing Systems},
 editor = {Z. Ghahramani and M. Welling and C. Cortes and N. Lawrence and K.Q. Weinberger},
 pages = {},
 publisher = {Curran Associates, Inc.},
 title = {{Log-Hilbert-Schmidt metric between positive definite operators on Hilbert spaces}},
 url = {https://proceedings.neurips.cc/paper/2014/file/f7664060cc52bc6f3d620bcedc94a4b6-Paper.pdf},
 volume = {27},
 year = {2014}
}

@article{Larotonda2007,
  title={{Nonpositive curvature: A geometrical approach to Hilbert–Schmidt operators}},
  author={G. Larotonda},
  journal={Differential Geometry and its Applications},
  year={2007},
  volume={25},
  pages={679–700}
}

@book{bogachev1998,
  title={{Gaussian measures}},
  author={Bogachev, V. I.},
  volume={62},
  year={1998},
  publisher={American Mathematical Society Providence}
}

@book{Bhatia2007,
  title={{Positive definite matrices}},
  author={R. Bhatia},
  volume={},
  year={2007},
  publisher={Princeton University Press}
}

@article{Minh2021,
    author = {Ha Quang, M.},
    year = {2021},
    pages = {},
    title = {Estimation of Riemannian distances between covariance operators and Gaussian processes},
    journal = {arXiv:2108.11683v1}
}

@InProceedings{shnitzer2022,
  title = 	 {Log-{E}uclidean Signatures for Intrinsic Distances Between Unaligned Datasets},
  author =       {Shnitzer, T. and Yurochkin, M. and Greenewald, K. and Solomon, J. M.},
  booktitle = 	 {Proceedings of the 39th International Conference on Machine Learning},
  pages = 	 {20106--20124},
  year = 	 {2022},
  editor = 	 {Chaudhuri, Kamalika and Jegelka, Stefanie and Song, Le and Szepesvari, Csaba and Niu, Gang and Sabato, Sivan},
  volume = 	 {162},
  series = 	 {Proceedings of Machine Learning Research},
  month = 	 {17--23 Jul},
  publisher =    {PMLR},
  url = 	 {https://proceedings.mlr.press/v162/shnitzer22a.html}
}

@InProceedings{huang2021,
    title = 	 {Projection Robust Wasserstein Barycenters},
    author =       {Huang, M. and Ma, S. and Lai, L.},
    booktitle = 	 {Proceedings of the 38th International Conference on Machine Learning},
    pages = 	 {4456--4465},
    year = 	 {2021},
    editor = 	 {Meila, Marina and Zhang, Tong},
    volume = 	 {139},
    series = 	 {Proceedings of Machine Learning Research},
    month = 	 {18--24 Jul},
    publisher =    {PMLR},
    url = 	 {https://proceedings.mlr.press/v139/huang21f.html}
}

@InProceedings{paty2019,
    title = 	 {Subspace Robust {W}asserstein Distances},
    author =       {Paty, F-P. and Cuturi, M.},
    booktitle = 	 {Proceedings of the 36th International Conference on Machine Learning},
    pages = 	 {5072--5081},
    year = 	 {2019},
    editor = 	 {Chaudhuri, Kamalika and Salakhutdinov, Ruslan},
    volume = 	 {97},
    series = 	 {Proceedings of Machine Learning Research},
    month = 	 {09--15 Jun},
    publisher =    {PMLR},
    url = 	 {https://proceedings.mlr.press/v97/paty19a.html}
}

@misc{tropp2017fixed,
  title        = {Fixed‑Rank Approximation of a Positive‑Semidefinite Matrix from Streaming Data},
  author       = {Tropp, J. A. and Yurtsever, A. and Udell, M. and Cevher, V.},
  year         = {2017},
  eprint       = {1706.05736},
  archivePrefix= {arXiv},
  primaryClass = {cs.NA},
  note         = {arXiv preprint},
}

@misc{tropp2023randomized,
  title        = {Randomized Algorithms for Low-Rank Matrix Approximation: Design, Analysis, and Applications},
  author       = {J. A. Tropp and Robert J. W.},
  year         = {2023},
  eprint       = {2306.12418},
  archivePrefix= {arXiv},
  primaryClass = {math.NA},
  note         = {arXiv preprint arXiv:2306.12418}
}

@inproceedings{kolouri2021wasserstein,
  title     = {Wasserstein Embedding for Graph Learning},
  author    = {Kolouri, S. and Naderializadeh, N. and Rohde, Gustavo K. and Hoffmann, H.},
  booktitle = {International Conference on Learning Representations (ICLR)},
  year      = {2021},
  url       = {https://openreview.net/forum?id=AAes_3W-2z},
}

@PHDTHESIS{nolot2013,
url = "http://www.theses.fr/2013DIJOS016",
title = "Convexit\'es et probl\`emes de transport optimal sur l'espace de Wiener",
author = "Nolot, V.",
year = "2013",
note = "Th\`ese de doctorat dirig\'ee par Fang, Shizan Math\'ematiques Dijon 2013",
note = "2013DIJOS016",
url = "http://www.theses.fr/2013DIJOS016/document",
}

@article{de2022riemannian,
  title={Riemannian score-based generative modelling},
  author={De Bortoli, V. and Mathieu, E. and Hutchinson, M. and Thornton, J. and Teh, Y. W. and Doucet, A.},
  journal={Advances in neural information processing systems},
  volume={35},
  pages={2406--2422},
  year={2022}
}

@article{kulis2009low,
  title={Low-Rank Kernel Learning with Bregman Matrix Divergences.},
  author={Kulis, B. and Sustik, M. A and Dhillon, I. S.},
  journal={Journal of Machine Learning Research},
  volume={10},
  number={2},
  year={2009}
}

@inproceedings{brooks2019exploring,
  title={Exploring complex time-series representations for Riemannian machine learning of radar data},
  author={Brooks, D. A. and Schwander, O. and Barbaresco, F. and Schneider, J-Y. and Cord, M.},
  booktitle={ICASSP 2019-2019 IEEE International Conference on Acoustics, Speech and Signal Processing (ICASSP)},
  pages={3672--3676},
  year={2019},
  organization={IEEE}
}

@inproceedings{guillaumin2009you,
  title={Is that you? Metric learning approaches for face identification},
  author={Guillaumin, M. and Verbeek, J. and Schmid, C.},
  booktitle={2009 IEEE 12th international conference on computer vision},
  pages={498--505},
  year={2009},
  organization={IEEE}
}

@inproceedings{mahadevan2018unified,
  title={A unified framework for domain adaptation using metric learning on manifolds},
  author={Mahadevan, S. and Mishra, B. and Ghosh, S.},
  booktitle={Joint European Conference on Machine Learning and Knowledge Discovery in Databases},
  pages={843--860},
  year={2018},
  organization={Springer}
}

@article{pennec2006riemannian,
  title={A Riemannian framework for tensor computing},
  author={Pennec, X. and Fillard, P. and Ayache, N.},
  journal={International Journal of computer vision},
  volume={66},
  number={1},
  pages={41--66},
  year={2006},
  publisher={Springer}
}

@inproceedings{tsuda2005matrix,
  title={Matrix exponentiated gradient updates for on-line learning and Bregman projection},
  author={Tsuda, K. and R{\"a}tsch, G. and Warmuth, M. K.},
  booktitle={Journal of Machine Learning Research},
  pages={995--1018},
  year={2005}
}

@article{dryden2009non,
  title={Non-Euclidean statistics for covariance matrices, with applications to diffusion tensor imaging},
  author={Dryden, I. L. and Koloydenko, A. and Zhou, D.},
  journal={The Annals of Applied Statistics},
  pages={1102--1123},
  year={2009},
  publisher={JSTOR}
}

@article{tosato2012characterizing,
  title={Characterizing humans on riemannian manifolds},
  author={Tosato, D. and Spera, M. and Cristani, M. and Murino, V.},
  journal={IEEE Transactions on Pattern Analysis and Machine Intelligence},
  volume={35},
  number={8},
  pages={1972--1984},
  year={2012},
  publisher={IEEE}
}

@article{tuzel2008pedestrian,
  title={Pedestrian detection via classification on riemannian manifolds},
  author={Tuzel, O. and Porikli, F. and Meer, P.},
  journal={IEEE transactions on pattern analysis and machine intelligence},
  volume={30},
  number={10},
  pages={1713--1727},
  year={2008},
  publisher={IEEE}
}

@inproceedings{thanwerdas2019affine,
  title={Is affine-invariance well defined on SPD matrices? A principled continuum of metrics},
  author={Thanwerdas, Y. and Pennec, X.},
  booktitle={International Conference on Geometric Science of Information},
  pages={502--510},
  year={2019},
  organization={Springer}
}

@article{arsigny2006log,
  title={Log-Euclidean metrics for fast and simple calculus on diffusion tensors},
  author={Arsigny, V. and Fillard, P. and Pennec, X. and Ayache, N.},
  journal={Magnetic Resonance in Medicine: An Official Journal of the International Society for Magnetic Resonance in Medicine},
  volume={56},
  number={2},
  pages={411--421},
  year={2006},
  publisher={Wiley Online Library}
}

@article{arsigny2007geometric,
  title={Geometric means in a novel vector space structure on symmetric positive-definite matrices},
  author={Arsigny, V. and Fillard, P. and Pennec, X. and Ayache, N.},
  journal={SIAM journal on matrix analysis and applications},
  volume={29},
  number={1},
  pages={328--347},
  year={2007},
  publisher={SIAM}
}

@article{malago2018wasserstein,
  title={Wasserstein Riemannian geometry of Gaussian densities},
  author={Malag{\`o}, L. and Montrucchio, L. and Pistone, G.},
  journal={Information Geometry},
  volume={1},
  number={2},
  pages={137--179},
  year={2018},
  publisher={Springer}
}

@article{oostrum2022bures,
  title={Bures--Wasserstein geometry for positive-definite Hermitian matrices and their trace-one subset},
  author={Oostrum, J. V an},
  journal={Information geometry},
  volume={5},
  number={2},
  pages={405--425},
  year={2022},
  publisher={Springer}
}

@article{sra2012new,
  title={A new metric on the manifold of kernel matrices with application to matrix geometric means},
  author={Sra, S.},
  journal={Advances in neural information processing systems},
  volume={25},
  year={2012}
}

@article{sra2016positive,
  title={Positive definite matrices and the S-divergence},
  author={Sra, S.},
  journal={Proceedings of the American Mathematical Society},
  volume={144},
  number={7},
  pages={2787--2797},
  year={2016}
}

@article{sra2021metrics,
  title={Metrics induced by Jensen-Shannon and related divergences on positive definite matrices},
  author={Sra, S.},
  journal={Linear Algebra and its Applications},
  volume={616},
  pages={125--138},
  year={2021},
  publisher={Elsevier}
}

@inproceedings{rabin2011wasserstein,
  title={Wasserstein barycenter and its application to texture mixing},
  author={Rabin, J. and Peyr{\'e}, G. and Delon, J. and Bernot, M.},
  booktitle={International conference on scale space and variational methods in computer vision},
  pages={435--446},
  year={2011},
  organization={Springer}
}

@article{solomon2015convolutional,
  title={Convolutional wasserstein distances: Efficient optimal transportation on geometric domains},
  author={Solomon, J. and De Goes, F. and Peyr{\'e}, G. and Cuturi, M. and Butscher, A. and Nguyen, A. and Du, T. and Guibas, L.},
  journal={ACM Transactions on Graphics (ToG)},
  volume={34},
  number={4},
  pages={1--11},
  year={2015},
  publisher={ACM New York, NY, USA}
}

@inproceedings{gramfort2015fast,
  title={Fast optimal transport averaging of neuroimaging data},
  author={Gramfort, A. and Peyr{\'e}, G. and Cuturi, M.},
  booktitle={International Conference on Information Processing in Medical Imaging},
  pages={261--272},
  year={2015},
  organization={Springer}
}

@article{demetci2020gromov,
  title={Gromov-Wasserstein optimal transport to align single-cell multi-omics data},
  author={Demetci, P. and Santorella, R. and Sandstede, B. and Noble, W. S. and Singh, R.},
  journal={BioRxiv},
  pages={2020--04},
  year={2020},
  publisher={Cold Spring Harbor Laboratory}
}

@article{dowson1982frechet,
  title={The Fr{\'e}chet distance between multivariate normal distributions},
  author={Dowson, D.C. and Landau, B.},
  journal={Journal of multivariate analysis},
  volume={12},
  number={3},
  pages={450--455},
  year={1982},
  publisher={Elsevier}
}

@article{givens1984class,
  title={A class of Wasserstein metrics for probability distributions.},
  author={Givens, Clark R and Shortt, Rae Michael},
  journal={Michigan Mathematical Journal},
  volume={31},
  number={2},
  pages={231--240},
  year={1984},
  publisher={University of Michigan, Department of Mathematics}
}

@article{olkin1982distance,
  title={The distance between two random vectors with given dispersion matrices},
  author={Olkin, Ingram and Pukelsheim, Friedrich},
  journal={Linear Algebra and its Applications},
  volume={48},
  pages={257--263},
  year={1982},
  publisher={Elsevier}
}

@book{brezis2011functional,
  title={Functional analysis, Sobolev spaces and partial differential equations},
  author={Brezis, Haim},
  volume={2},
  number={3},
  year={2011},
  publisher={Springer}
}
\newpage
\appendix
\section{Theorems of use to the analysis}\label{theorems}
\begin{theorem}(\textbf{Spectral theorem})\label{spectralTheorem}
    Let $\bf{A}$ be a compact self-adjoint operator on $\mathcal{H}$. There exist an orthonormal system $\{\phi_{k}\}$ with $k \in [ 1, \mathrm{dim}(\mathcal{H})]$ of eigenvectors of $\bf{A}$ and corresponding eigenvalues ${\lambda_{k}}$ such that for all $x \in \mathcal{H}$,
    
    \begin{equation}\label{spec_thm_def}
        {\bf{A}}x = \sum_{k}\lambda_{k}\langle x, \phi_{k}\rangle\phi_{k}.
    \end{equation}

    If $\{\lambda_{k}\}$ is an infinite sequence, then it converges to zero.
\end{theorem}
We provide a brief explanation of the proof within the context of the generalization of the family of Alpha-Procrustes distances. Denote $L(\mathcal{H})$ the set of bounded linear operators on $\mathcal{H}$. Denote $\mathbb{S}^{n}_{++} \subset \mathbb{S}^{n}_{+} \subset L(\mathcal{H})$ the sets of strictly positive and positive operators on $\mathcal{H}$ respectively. Denote $\mathbb{U}(\mathcal{H})$ the set of unitary operators on $\mathcal{H}$.
Let ${\bf{A}} \in \mathbb{S}^{n}$ be a compact operator, then by virtue of the Riesz-Schauder theorem the spectrum $\sigma({\bf{A}})$ of ${\bf{A}}$ is a discrete set having no limit point except $\lambda = 0$, where $\lambda$ is an Eigenvalue of ${\bf{A}}$. Further, any non-zero Eigenvalues $\lambda \in \sigma({\bf{A}})$ is of finite multiplicity. This follows immediately from the fact that ${\bf{A}}$ is compact. If the compact operator ${\bf{A}}$ is self-adjoint then by the Hilbert-Schmidt theorem there is a complete orthonormal basis $\{\phi_{k}\}$, for $\mathcal{H}$ so that ${\bf{A}}\phi_{k} = \lambda_{k}\phi_{k}$ and $\lim_{k \rightarrow \infty}\lambda_{k}({\bf{A}}) = 0$ with $k \in [1, \mathrm{dim}(\mathcal{H})]$ where $\mathrm{dim}(\mathcal{H}) = \infty$. Since the Riesz-Schauder theorem states that each non-zero Eigenvalue has finite multiplicity and the only possible limit point of the $\lambda_{k}$ is zero, the limit $k \rightarrow \infty$ of $\lambda_{k}$ going to zero is licit. We can see how this becomes an issue when considering the logarithm of ${\bf{A}}$. As pointed out in \citep{Minh2014}, the latter is bounded if and only if $\mathrm{dim}(\mathcal{H}) < \infty$ and when $\mathrm{dim}(\mathcal{H}) = \infty$, one has that $\lim_{k\rightarrow \infty}\log(\lambda_{k}) = - \infty$. It is clear that the strict positivity of the operators is not sufficient for $\log({\bf{A}})$ to be well-defined. The strategy of \citep{Minh2014} is to consider the stronger condition of bounding the Eigenvalues of an operator ${\bf{A}} \in L(\mathcal{H})$ from below. ${\bf{A}}$ is said to be positive definite if there exists a constant $C$ such that $\langle {\bf{A}}x, x \rangle \geq C \cdot \vert\vert x \vert\vert^{2}$, where $\vert\vert \cdot \vert\vert$ is the operator norm and $x \in \mathcal{H}$. The set of self-adjoint, positive definite operators on $\mathcal{H}$ is then defined as follows \citep{Minh2014, MINH202225}
\begin{equation}\label{space_of_sa_pd_op}
    \mathbb{P}(\mathcal{H}) = \{ {\bf{A}}: {\bf{A}} \in L(\mathcal{H}), \,{\bf{A}}^{*} = {\bf{A}},\,\, \exists M > 0 \,\,\, \mathrm{s.t} \,\,\, \langle x, {\bf{A}}x \rangle \geq M \vert\vert x \vert\vert^{2}\,\,\, \forall x \in \mathcal{H}\}.
\end{equation}
\subsection{Completeness of eigenvectors of $\textbf{M}$ on $\mathcal{S}_{2}(\mathcal{H})$}\label{completenessOfMEigvec}
\begin{lemma}\label{completenessOfMEigvecLemma}
    Let $\textbf{M} \in \mathcal{S}_{2}(\mathcal{H})$ be compact, self-adjoint, and positive definite. Then its eigenvectors $\{\psi_{k}\}_{k\geq 1}$ form a complete orthonormal system in $\mathcal{H}$. 
\end{lemma}
\begin{proof}
    Orthonormality follows from self-adjointness and Gram-Schmidt within each finite-multiplicity eigenspace. Compactness of $\mathbf{M}$ follows from $\textbf{M}\in \mathcal{S}_{2}(\mathcal{H})$ via the finite-rank truncations $\textbf{M}_{n} = \sum_{k=1}^{n}\omega_{k}\langle\, \cdot\,, \psi_{k}\rangle\psi_{k}$, since $|| \textbf{M} - \textbf{M}_{n}||_{\text{op}}\leq || \textbf{M} - \textbf{M}_{n}||_{\mathcal{S_{2}(\mathcal{H})}} = (\sum_{k=n+1}^{\infty}\omega_{k}^{2})^{1/2} \rightarrow 0$. Existence of each successive eigenvector follows from attainment of $||\textbf{M}_{n}||_{\text{op}}$ and compactness; the eigenvalues satisfy $|\omega_{k}|\rightarrow 0$ since otherwise $\{\textbf{M}\psi_{n}\} = \{\omega_{n}\psi_{n}\}$ would have no convergent subsequence, contradicting compactness of $\mathbf{M}$ and orthonormality of $\{\psi_{n}\}$. For completeness, let $\mathcal{H}_{\infty} = \{\psi_{1}, \psi_{2}, \cdots\}^{\perp}$. Then $\langle\textbf{M}x, x\rangle \leq |\omega_{n+1}||x||^{2} \rightarrow 0$ for all $x \in \mathcal{H}_{\infty}$, so $\textbf{M}x = 0$ on $\mathcal{H}_{\infty}$. Positive definiteness of $\textbf{M}$ implies injectivity, hence $\mathcal{H}_{\infty} = \{0\}$. 
\end{proof}
\begin{rem}
    Positive definiteness is used only to conclude $\text{ker}(\textbf{M}) = 0$ and hence $\mathcal{H}_{\infty} = \{0\}$. For positive semi-definite $\textbf{M}$, completeness would fail as $\{\psi_{k}\}$ would span only $\text{ker}(\textbf{M})^{\perp}$, leaving $\text{ker}(\mathbf{M})$ unaccounted for. This is the precise reason positive definiteness is required in Definition \ref{inftyMahalonobisNorm}.
\end{rem}
% for every x ∈ H), x = Σₖ ⟨x, ψₖ⟩ ψₖ in norm, and the only vector orthogonal to all {ψₖ} is zero.
%
%
%
%
\subsection{Establishing $||\cdot||_{\textbf{M}_{\infty}^{-1}}$ as a norm on $\mathcal{S}_{2}^{(\mathrm{EXT})}(\mathcal{H})$}\label{extMahalanobisNormProperties}
\begin{lemma}\label{extMahalanobisNormPropertiesProp}
    Let $\delta, \rho\in \mathbb{R}_{>0}$. Let $(\bf{X} + \delta \mathbf{I}) \in \mathcal{S}_{2}^{(\mathrm{EXT})}(\mathcal{H})$. The functional $||\cdot||_{\bf{M}_{\infty}^{-1}}$ defined by $||(\bf{X} + \delta \mathbf{I})||^{2}_{\bf{M}_{\infty}^{-1}} = \text{tr}[\mathbf{X}^{*}(\mathbf{M} + \rho\mathbf{I})^{-1}\mathbf{X}] + \delta^{2}$ is a well-defined norm on $\mathcal{S}_{2}^{(\mathrm{EXT})}(\mathcal{H})$. Specifically:
    \begin{enumerate}
        \item (Finiteness) For every $x\in \mathcal{S}_{2}(\mathcal{H})$ and $\delta \in \mathbb{R}$, the quantity $\text{tr}[\mathbf{X}^{*}(\mathbf{M} + \rho \mathbf{I})^{-1}\mathbf{X}]$ is finite by the operator ideal property $\mathcal{S}_{2}\cdot\mathcal{B}(\mathcal{H})\cdot\mathcal{S}_{2}\subseteq \mathcal{S}_{1}$. 

        \item (Non-negativity and definiteness) $||(\bf{X}+ \delta \bf{I})||_{\bf{M}_{\infty}^{-1}} = 0$ if and only if $\bf{X} = 0$ and $\delta=0$.

        \item (Homogeneity) $||\lambda(\bf{X}+ \delta\bf{I})||_{\bf{M}_{\infty}^{-1}} = |\lambda|\cdot||(\bf{X}+ \delta\bf{I})||_{\bf{M}_{\infty}^{-1}}$ for all $\lambda\in \mathbb{R}$.

        \item (Triangle inequality) This follows from the Cauchy-Schwarz inequality for the inner product $\langle \bf{X} + \delta \bf{I}, \bf{Y} + \delta \bf{I}\rangle_{\mathcal{S}_{2}^{(\text{EXT})}} = \text{tr}[\bf{X}^{*}(\bf{M}+ \rho\bf{I})^{-1}\bf{Y}] + \delta^{2}$, which is well-defined and positive definite on $\mathcal{S}_{2}^{(\text{EXT})}$.
    \end{enumerate}
\end{lemma}

\section{Some convergence analyses with respect to Gaussian operators in the weak topology}\label{firstConvergenceStudies}
\begin{rem}
Given a sequence of probability measures $\mu_{n} \sim \mathcal{N}(0, {\bf{C}}_{n})$ with mean $0$ and covariance ${\bf{C}}_{n}$. We note that, for the purpose of later works, that it will be necessary to distinguish between the case when $\mu_{n}$ is close to a sequence $\mu_{m}$ with high enough probability or if they are independent Gaussians. In the latter case, they will not converge in probability. They will simply converge in distributions.
\end{rem}
In this section, we explore the topological properties of the Procrustes and Alpha-Procrustes distances. The topological properties associated to the $p$-Wasserstein metric are now well established. The study of the continuity of the $p$-Wasserstein metric with respect to converging sequences of probability measures $\{\mu_{n}\}_{n \in \mathbb{N}}$ provides an important characterisation of the metrizability of the weak topology \citep{villani2009}. The weak convergence of Gaussian processes is also very well understood \citep{bogachev1998}. It makes sense to leverage those techniques in order to better characterise the topology induced by the (Alpha)-Procrustes metric on the manifold of symmetric positive definite matrices. We first work out the topological properties of the Procrustes distance the following preparatory lemma before understanding how to generalize to the Alpha-Procrustes family of metrics.
\begin{lemma}\label{procrustesTopologyLemma}
Let $\{{\bf{C}}_{n}\}_{n \in \mathbb{N}}$ define the sequence of covariance operators ${\bf{C}}$ on $\mathcal{H}$. Let $\mathcal{N}(0, {\bf{C}})$ represent the zero-mean non-degenerate Gaussian distribution. The following are equivalent
\begin{itemize}
    \item [1.] $\lim_{n \rightarrow \infty}\mathcal{N}(0, {\bf{C}}_{n}) \rightarrow \mathcal{N}(0, {\bf{C}})$ weakly;
    
    \item [2.] $\lim_{n \rightarrow \infty} d_{\mathrm{BW}}({\bf{C}}_{n}, {\bf{C}}) \rightarrow 0$,
\end{itemize}
where $d_{\mathrm{BW}}$ corresponds to the Bures-Wasserstein distance (\ref{bwDistance}) defined on $\mathbb{S}_{++}^{n}$.
\end{lemma}
\begin{corollary}\label{procrustesTopologyCorr}
Let $\{{\bf{C}}_{n}\}_{n \in \mathbb{N}}$ define the sequence of covariance operators ${\bf{C}}$ on $\mathcal{H}$. One has that
\begin{equation}\label{limGBW}
    \lim_{n\rightarrow\infty}d_{\mathrm{GBW}}({\bf{C}}_{n}, {\bf{C}}) \rightarrow 0,
\end{equation}
where $d_{\mathrm{GBW}}$ corresponds to the generalised Bures-Wasserstein (\ref{gbwDistance}). \\

Further, $\lim_{n\rightarrow\infty}d_{\mathrm{GBW_{\infty}}}({\bf{C}}_{n}, {\bf{C}}) \rightarrow 0$ where $d_{\mathrm{GBW_{\infty}}}$ is the infinite dimensional generalized Bures-Wasserstein distance (\ref{corrInfAlphaGBW}). 
\end{corollary}

The proof of Lemma \ref{procrustesTopologyLemma} relies on ~\cite[Definition 6.8 and Theorem 6.9]{villani2009}. The proof of Theorem 6.9 is instructive for the discussion that follows. The latter shows that weak convergence in the $p$-Wasserstein space is equivalent to the tight convergence of sequence of probability measures $\{\mu_{n}\}_{n\in \mathbb{N}}$ and the convergence of the moments of order $p$. A simpler demonstration of Theorem 6.9 follows by working in a complete metric space which is compact and by leveraging the dual formulation of the $1$-Wasserstein distance $\mathcal{W}_{1}$ as well as comparison inequalities between $\mathcal{W}_{1}$ and $\mathcal{W}_{p}$. For the sake of readability, we recall ~\cite[Definition 6.8 (i)]{villani2009}.
\begin{definition}\label{weakConvInWp}(\textbf{Weak convergence in Wasserstein Space} \citep{villani2009}) Let $(\mathcal{X}, d)$ be a Polish metric space, and let $p \in [1, \infty)$. Let $\{\mu_{n}\}_{n\in \mathbb{N}}$ be a sequence of probability measures in the Wasserstein space $\mathcal{P}_{p}(\mathcal{X})$ (see Definition 6.4 \citep{villani2009}) and let $\mu \in \mathcal{P}_{p}(\mathcal{X})$ also. Then for some $x_{0} \in \mathcal{X}$
\begin{equation}
    \mu_{n} \xrightarrow[n \rightarrow \infty]{} \mu \,\,\,\,\, \mathrm{weakly} \,\,\,\,\, \mathrm{and}\,\,\,\,\, \int d(x_{0}, x)^{p}d\mu_{k}(x) \xrightarrow[n\rightarrow\infty]{}\int d(x_{0}, x)^{p}d\mu(x),
\end{equation}
therefore, ensuring convergence of the moment of order $p$. 
\end{definition}
\begin{proof}(\textbf{Lemma} \ref{procrustesTopologyLemma})
Define the zero-mean non-degenerate Gaussian measures $\mu_{n} \sim \mathcal{N}(0, {\bf{C}}_{n})$ and $\mu \sim \mathcal{N}(0, {\bf{C}})$. By virtue of (\ref{W_2Distance}), it follows that $d_{\mathrm{BW}}({\bf{C}}_{n}, {\bf{C}}) = \mathcal{W}_{2}(\mu_{n}, \mu)$ for $m = n = 0$. The limit is then clear by thanks to Definition \ref{weakConvInWp} and Theorem 6.9 \citep{villani2009} that statements 1. and 2. are equivalent.  
\end{proof}
\begin{proof}(\textbf{Corollary} \ref{procrustesTopologyCorr}) We recall from \citep{HMJG2021b} that the generalized Bures-Wasserstein $d_{\mathrm{GBW}}$ coincide with the generalised $2$-Wasserstein distance $\Tilde{\mathcal{W}}_{2}$ which is defined as 
\begin{equation}\label{generalisedWp}
    \Tilde{\mathcal{W}}_{2}(\mu, \nu) := \bigg\{ \inf_{\pi \in \Pi(\mu, \nu)} \int _{\mathbb{R}^{n} \times \mathbb{R}^{n}} d(x, y)^{2}  d\gamma(x, y) \bigg\}^{1/2}\,\,\,\,\, \mathrm{with} \,\,\,\,\, d(x, y) = \vert\vert x-y \vert\vert_{{\bf{M}}^{-1}}
\end{equation}
where $\vert\vert \cdot \vert\vert_{{\bf{M}}^{-1}}$ corresponds to the Mahalanobis defined earlier. $\pi$ represent the transport plan and $\Pi(\mu, \nu)$ is the set of joints distributions with marginals $\mu$ and $\nu$. By virtue of Lemma \ref{procrustesTopologyLemma}, we have that $d_{\mathrm{GBW}}({\bf{C}}_{n}, {\bf{C}}) = \Tilde{\mathcal{W}}_{2}(\mu_{n}, \mu)$. It is clear that in this setting the limit in (\ref{limGBW}) is licit by virtue of Definition \ref{weakConvInWp}, Lemma \ref{procrustesTopologyLemma} and ~\cite[Theorem 6.9]{villani2009}. \\

We now move to the proof of the second part of the corollary. The $2$-Wasserstein distance along with its generalised version (\ref{generalisedWp}) can be formulated on $\mathrm{dim}(\mathcal{H}) = \infty$ also \citep{Gelbrich1990}. Define the generalised $2$-Wasserstein as $ \Tilde{\mathcal{W}}_{2, \infty}(\mu, \nu)$ which is (\ref{generalisedWp}) but with $d(x, y) = \vert\vert x-y\vert\vert_{{\bf{M}}_{\infty}^{-1}}$ and with $\mu, \nu \in \mathcal{P}_{2}(\mathcal{H})$, the $2$-Wasserstein space. By virtue of Lemma \ref{procrustesTopologyLemma} and the proof of the first part of \ref{procrustesTopologyCorr} we can infer that the infinite dimensional generalized Bures-Wasserstein coincide with the infinite dimensional generalized $2$-Wasserstein distance. Consequently, the weak convergence of the probability measures in $\mathcal{P}_{2}(\mathcal{H})$ follows from Definition \ref{weakConvInWp}, Lemma \ref{procrustesTopologyLemma} and ~\cite[Theorem 6.9]{villani2009}. To establish the convergence of the $p$-th order moments, we first consider the definition of the infinite dimensional generalised Bures Wasserstein distance (\ref{corrInfAlphaGBW}) with ${\bf{C}}_{n}$ and ${\bf{C}}$. Let $x\sim \mu_{n}$. The trace $\mathrm{tr}[({\bf{M}} + \rho{\bf{I}})^{-1}{\bf{C}}_{n}]$ is expressed as 
\begin{equation}
    \mathrm{tr}[({\bf{M}} + \rho{\bf{I}})^{-1}{\bf{C}}_{n}] = \int_{\Sigma_{\mathcal{S}_{2}^{(\mathrm{EXT})}}(\mathcal{H})}\vert\vert x\vert\vert^{2}_{{\bf{M}}_{\infty}^{-1}} d\mu_{n}(x)
\end{equation}
By virtue of theorems on convergence of Gaussian measures \citep{bogachev1998}, one has that
\begin{equation}
    \lim_{n\rightarrow\infty} \int_{\Sigma_{\mathcal{S}_{2}^{(\mathrm{EXT})}}(\mathcal{H})}\vert\vert x\vert\vert^{2}_{{\bf{M}}_{\infty}^{-1}} d\mu_{n}(x) = \int_{\Sigma_{\mathcal{S}_{2}^{(\mathrm{EXT})}}(\mathcal{H})}\vert\vert x\vert\vert^{2}_{{\bf{M}}_{\infty}^{-1}} d\mu(x) = \mathrm{tr}[({\bf{M}} + \rho{\bf{I}})^{-1}{\bf{C}}]. 
\end{equation}
Applying the above the result to the rest of the traces ensures that $ \Tilde{\mathcal{W}}_{2, \infty}(\mu_{n}, \mu) \rightarrow 0$ as $n \rightarrow \infty$. Clearly, this is equivalent to $d_{\mathrm{GBW_{\infty}}}({\bf{C}}_{n}, {\bf{C}})\rightarrow 0$ as $n \rightarrow \infty$. This entails the claim. 
\end{proof}
%
%
%
%
%{\color{red}The details of the second part of the corollary still needs reviewing. Extending to Alpha-Procrustes metrics should not be too complicated and follow the above structure, although it remain to understand carefully how the values of $\alpha$ might affect the convergence.}
% \subsection{Convergence of Alpha-Procrustes distances}

%
%
%
%
\section{Proofs for the finite dimensional setting}\label{finiteDimProofs}
\subsection{Proof of Proposition \ref{propAlphaGBW}}
Let $\alpha \in \mathbb{R}_{>0}$ be fixed. Given two matrices ${\bf{X}}, {\bf{Y}} \in \mathbb{S}^{n}_{++}$, one has that
\begin{equation*}
        \min_{{\bf{O}}\in \mathbb{O}(n)}\bigg\vert\bigg\vert
        \frac{{\bf{X}}^{\alpha} - {\bf{Y}}^{\alpha}{\bf{O}} }{\alpha}\bigg\vert\bigg\vert_{{\bf{M}}^{-1}}^{2} = \frac{1}{\alpha^{2}}\mathrm{tr}\big[({\bf{X}}^{\alpha} - {\bf{Y}}^{\alpha}{\bf{O}})^\top{\bf{M}}^{-1}({\bf{X}}^{\alpha} - {\bf{Y}}^{\alpha}{\bf{O}}) \big], 
\end{equation*}
and by expanding the r.h.s one has that
\begin{equation}\label{eqAlphaGBW}
    \begin{split}
        \frac{1}{\alpha^{2}}\mathrm{tr}\big[({\bf{X}}^{\alpha} - {\bf{Y}}^{\alpha}{\bf{O}})^\top{\bf{M}}^{-1}({\bf{X}}^{\alpha} - {\bf{Y}}^{\alpha}{\bf{O}}) \big] =  & \frac{1}{\alpha^{2}}\big[\mathrm{tr}({\bf{M}}^{-1}{\bf{X}}^{2\alpha}) + \mathrm{tr}({\bf{M}}^{-1} {\bf{Y}}^{2\alpha}) \\ & - \mathrm{tr}\big({\bf{X}}^{\alpha}{\bf{M}}^{-1}{\bf{Y}}^{\alpha}{\bf{O}} + {\bf{O}}^{T}{\bf{Y}}^{\alpha}{\bf{M}}^{-1}{\bf{X}}^{\alpha}\big)  \big]. 
     \end{split}
\end{equation}
The term $\mathrm{tr}\big({\bf{X}}^{\alpha}{\bf{M}}^{-1}{\bf{Y}}^{\alpha}{\bf{O}} + {\bf{O}}^{T}{\bf{Y}}^{\alpha}{\bf{M}}^{-1}{\bf{X}}^{\alpha}\big)$ is dealt with in a similar manner as in \citep{HMJG2021b}. The minimum of (\ref{eqAlphaGBW}) is attained when $\mathbf{O}$ is the orthogonal polar factor of ${\bf{Y}}^{\alpha}{\bf{M}}^{-1}{\bf{X}}^{\alpha}$. This is shown as follows. Define a matrix ${\bf{Q}} = {\bf{Y}}^{\alpha}{\bf{M}}^{-1}{\bf{X}}^{\alpha}$ such that $|{\bf{Q}}| = ({\bf{Q}}^\top{\bf{Q}})^{1/2} = ({\bf{X}}^{\alpha}{\bf{M}}^{-1}{\bf{Y}}^{2\alpha}{\bf{M}}^{-1}{\bf{X}}^{\alpha})^{1/2}$. Since ${\bf{Q}}$ can be re-expressed as ${\Omega}{\bf{P}}$ where ${\bf{P}} = |\Omega|$ and $\Omega \in \mathcal{O}(n)$. This implies that
\begin{align*}
    \Omega = \mathbf{Q}\mathbf{P}^{-1} = {\bf{Y}}^{\alpha}{\bf{M}}^{-1}{\bf{X}}^{\alpha}({\bf{X}}^{\alpha}{\bf{M}}^{-1}{\bf{Y}}^{2\alpha}{\bf{M}}^{-1}{\bf{X}}^{\alpha})^{-1/2}.
\end{align*}
Then for $\Omega = \mathbf{O}$ we have $\mathrm{tr}\big(\mathbf{O}^\top\mathbf{Q} + \mathbf{Q}^\top\mathbf{O}\big)$. One then has that $\mathrm{tr}(2\mathbf{P}) = 2\mathrm{tr}({\bf{X}}^{\alpha}{\bf{M}}^{-1}{\bf{Y}}^{2\alpha}{\bf{M}}^{-1}{\bf{X}}^{\alpha})^{1/2}$. Setting $\alpha = 1/2$ one recovers the GBW distance. This entails the proof. 
\subsection{The finite dimensional generalised Log-Euclidean distance}\label{finiteDimGeneralizedLogEU}
The generalised Log-Euclidean distance can also be realised from the generalised Alpha-Procrustes distance (\ref{alphaProcrustesM}). 
\begin{proposition}\label{propAlphaLogEU}
    Let ${\bf{X}}, {\bf{Y}} \in \mathbb{S}_{++}^{n}$ and let ${\bf{M}} \in \mathbb{S}_{++}^{n}$ be jointly diagonalizable with ${\bf{X}}$ and ${\bf{Y}}$ (equivalently, $[\mathbf{M}, \mathbf{X}] = [\mathbf{M}, \mathbf{Y}] = [\mathbf{X}, \mathbf{Y}] = 0$). The Log-Euclidean limit of the distance (\ref{alphaProcrustesM}), defined by continuous extension as the one-sided limit $\alpha \to 0^{+}$ (taken outside the domain $\alpha\in\mathbb{R}_{>0}$ of $d^{\alpha}_{\mathrm{proM}}$), admits the following representation
   
    \begin{equation}\label{LogEUDistance}
    \lim_{\alpha \rightarrow 0^{+}}[d^{\alpha}_{\mathrm{proM}}({\bf{X}}, {\bf{Y}})] = \vert\vert \log({\bf{X}}) - \log({\bf{Y}}) \vert\vert_{{\bf{M}}^{-1}}
    \end{equation}
which we called the \textbf{generalized Log-EU distance}.
\end{proposition}
\begin{proof}
    The proof is slightly more involved but follows along the lines of the proof of Theorem 2 in \citep{MINH202225}. Under the joint-diagonalizability hypothesis, ${\bf{X}}$, ${\bf{Y}}$, and ${\bf{M}}$ share a common orthonormal eigenbasis, so all operator products commute and reduce to scalar computations on the eigenvalues. Given that ${\bf{X}}$ and ${\bf{Y}}$ commute, it is clear that one can express 
    % (\ref{alphaGBW}) as follows
    %
    %
    %
    %
    \begin{equation*}
        d^{\alpha}_{\mathrm{proM}}({\bf{X}}, {\bf{Y}}) = \bigg\vert\bigg\vert\frac{{\bf{X}}^{\alpha} - {\bf{Y}}^{\alpha}}{\alpha} \bigg\vert\bigg\vert_{{\bf{M}}^{-1}}.  
    \end{equation*}
    as follows. We have
    \begin{equation*}
        \begin{split}
            \bigg\vert\bigg\vert\frac{{\bf{X}}^{\alpha} - {\bf{Y}}^{\alpha}}{\alpha} \bigg\vert\bigg\vert^{2}_{{\bf{M}}^{-1}} =
            \frac{1}{\alpha^{2}}
            \bigg\{
            \vert\vert {\bf{X}}^{\alpha} - {\bf{I}}\vert\vert^{2}_{{\bf{M}}^{-1}} & + \vert\vert {\bf{Y}}^{\alpha} - {\bf{I}} \vert\vert^{2}_{{\bf{M}}^{-1}} \\ &
            -2 \mathrm{tr}[({\bf{X}}^{\alpha}{\bf{M}}^{-1}{\bf{Y}}^{2\alpha}{\bf{M}}^{-1}{\bf{X}}^{\alpha})^{1/2} \\ &- {\bf{M}}^{-1}{\bf{X}}^{\alpha} - {\bf{M}}^{-1}{\bf{Y}}^{\alpha} + 
            {\bf{M}}^{-1} ] 
            \bigg\}. 
         \end{split}
    \end{equation*}
    At this point the expressions are in a suitable form such that one can readily take limit $\alpha \rightarrow 0$ for the first two terms above. We first express ${\bf{X}}^{\alpha}$ and ${\bf{Y}}^{\alpha}$ as 
    \begin{equation*}
        \begin{split}
            & {\bf{X}}^{\alpha} = \exp(\log({\bf{X}}^{\alpha})) \\ &
            {\bf{Y}}^{\alpha} = \exp(\log({\bf{Y}}^{\alpha})), 
        \end{split}
    \end{equation*}
    such that
    \begin{equation}\label{lim1LogEU}
        \lim_{\alpha \rightarrow 0} \bigg\vert\bigg\vert \frac{\exp(\log({\bf{X}}^{\alpha})) - {\bf{I}}}{\alpha}\bigg\vert\bigg\vert^{2}_{{\bf{M}}^{-1}} = \vert\vert \log({\bf{X}}) \vert\vert^{2}_{{\bf{M}}^{-1}},
    \end{equation}
    \begin{equation}\label{lim2LogEU}
        \lim_{\alpha \rightarrow 0} \bigg\vert\bigg\vert \frac{\exp(\log({\bf{Y}}^{\alpha})) - {\bf{I}}}{\alpha}\bigg\vert\bigg\vert^{2}_{{\bf{M}}^{-1}} = \vert\vert \log({\bf{Y}}) \vert\vert^{2}_{{\bf{M}}^{-1}}.
    \end{equation}
    We now look at the final term. We substitute the expansions for ${\bf{X}^{\alpha}}$ and ${\bf{Y}^{\alpha}}$ in ${\bf{X}}^{\alpha}{\bf{M}}^{-1}{\bf{Y}}^{2\alpha}{\bf{M}}^{-1}{\bf{X}}^{\alpha}$ and upon simplifying we have that 
    \begin{equation*}
        \begin{split}
            {\bf{X}}^{\alpha}{\bf{M}}^{-1}{\bf{Y}}^{2\alpha}{\bf{M}}^{-1}{\bf{X}}^{\alpha}  =
            ({\bf{M}}^{-1})^{2} & +
            2\alpha\big[({\bf{M}}^{-1})^{2}\log({\bf{X}}) + ({\bf{M}}^{-1})^{2}\log({\bf{Y}})\big] \\ & +
            2\alpha^{2}\big[({\bf{M}}^{-1})^{2}\log({\bf{X}}) +  ({\bf{M}}^{-1})^{2}\log({\bf{Y}})\big]^{2}.
        \end{split}
    \end{equation*}
    Consider $\sqrt{1 + x}$. The power series for $\sqrt{1 + x} = 1 + \sum_{k=1}^{\infty}c_{k}x^{k}$ for $\vert x \vert \leq 1$ converges absolutely\footnote{This immediately follows from the fact that: 
    \begin{equation*}
        1 + x = \bigg(1 + \sum_{k=1}^{\infty}c_{k}x^{k}\bigg)\bigg(1 + \sum_{k=1}^{\infty}c_{k}x^{k}\bigg) = \sum_{j, k\geq 0}c_{j}c_{k}x^{j+k} = \sum_{l=0}^{\infty}\bigg(\sum_{k = 0}^{l} c_{k}c_{l-k} x^{l}\bigg).
    \end{equation*}
    Clearly, $c_{0} = 1$ and $c_{0}c_{1} + c_{1}c_{0} = -1$ and the coefficients for $l\geq2$ are all zero. This means that $$\sum_{l\geq 0}\vert c_{l} \vert = 2 - \sum_{l \geq 0 } c_{l} \leq 2 - \lim_{x \rightarrow 1-}\sqrt{1+x} = 2,$$ where $\lim_{x \rightarrow 1-}$ means the limit tends to one from below. Since this is licit for all $N$, $\sum_{l\geq 0}\vert c_{l} \vert \leq 2$ implies that the series $1 + \sum_{k=1}^{\infty}c_{k}x^{k}$ converges absolutely for $\vert x \vert \leq 1$.}. Consequently for $\vert\vert {\bf{X}} \vert\vert < 1$, it is licit to express the series below as
    \begin{equation*}
        (1 + {\bf{X}})^{1/2} = 1 + \frac{{\bf{X}}}{2} - \frac{{\bf{X}}^{2}}{8} + \cdots .
    \end{equation*}
    Therefore, 
    \begin{equation*}
        \begin{split}
            ({\bf{X}}^{\alpha}{\bf{M}}^{-1}{\bf{Y}}^{2\alpha}{\bf{M}}^{-1}{\bf{X}}^{\alpha})^{1/2}  =
            ({\bf{M}}^{-1}) \bigg\{\mathbb{I} +
            \alpha\big[\log({\bf{X}}) + \log({\bf{Y}})\big] +
            \frac{\alpha^{2}}{2}\big[\log({\bf{X}}) +  \log({\bf{Y}})\big]^{2}\bigg\}.
        \end{split}
    \end{equation*}
    The above expression has now been greatly simplified and upon expanding the terms on the right hand side and collecting terms entering in the expansion of ${\bf{X}}^{\alpha}$ and ${\bf{Y}}^{\alpha}$ respectively we are able to take the limit $\alpha \rightarrow 0$
    \begin{equation*}\label{lim3LogEU}
        \begin{split}
            \lim_{\alpha \rightarrow 0} \frac{1}{\alpha^{2}} 2 \mathrm{tr}[({\bf{X}}^{\alpha}{\bf{M}}^{-1}{\bf{Y}}^{2\alpha}{\bf{M}}^{-1}{\bf{X}}^{\alpha})^{1/2} - & {\bf{M}}^{-1}{\bf{X}}^{\alpha} - {\bf{M}}^{-1}{\bf{Y}}^{\alpha} + 
            {\bf{M}}^{-1} ] \\ &
            = \mathrm{tr}\big[\log({\bf{X}}){\bf{M}}^{-1}\log({\bf{Y}}) + \log({\bf{Y}}){\bf{M}}^{-1}\log({\bf{X}})\big]. 
        \end{split}
    \end{equation*}
    Hence the limit in (\ref{LogEUDistance}) is 
    \begin{equation*}
        \begin{split}
            \lim_{\alpha \rightarrow 0}[d^{\alpha}_{\mathrm{proM}}({\bf{X}}, {\bf{Y}})]^{2} = \vert\vert \log({\bf{X}}) \vert\vert^{2}_{{\bf{M}}^{-1}} + & \vert\vert \log({\bf{Y}}) \vert\vert^{2}_{{\bf{M}}^{-1}} \\ & - \mathrm{tr}\big[\log({\bf{X}}){\bf{M}}^{-1}\log({\bf{Y}}) + \log({\bf{Y}}){\bf{M}}^{-1}\log({\bf{X}})\big]
        \end{split}
    \end{equation*}
    which is equal to $\vert\vert \log({\bf{X}}) - \log({\bf{Y}}) \vert\vert^{2}_{{\bf{M}}^{-1}}$. This establishes the claim. 
\end{proof}
\section{Proofs for the infinite dimensional setting}\label{infiniteDimProofs}
\subsection{Proof of Proposition \ref{generalisedAlphaProcrusteInfDim}}\label{propgeneralisedAlphaProcrusteInfDim}
Let $\alpha \in \mathbb{R}_{>0}$ be fixed. Let $\delta, \gamma \in \mathbb{R}$ and $\rho \in \mathbb{R}_{>0}$ be fixed. 
Let $({\bf{X}} + \delta{\bf{I}}), ({\bf{Y}} + \gamma{\bf{I}}) \in \Sigma_{\mathcal{S}_{2}^{(\mathrm{EXT})}}(\mathcal{H})$.  ${\bf{M}} + \rho{\bf{I}} \in \Sigma_{\mathcal{S}_{2}^{(\mathrm{EXT})}}(\mathcal{H})$. One has that \\
\begin{equation}
    \begin{split}
        \bigg\vert\bigg\vert \frac{({\bf{X}} + \delta{\bf{I}})^{\alpha^{2}} - ({\bf{Y}} + \gamma{\bf{I}})^{\alpha}({\bf{I}} + {\bf{U}})}{\alpha} \bigg\vert\bigg\vert_{{\bf{M}}_{\infty}^{-1}}^{2} & = \frac{1}{\alpha}\bigg\{ \vert\vert ({\bf{X}} + \delta{\bf{I}})^{\alpha} \vert\vert_{{\bf{M}}_{\infty}^{-1}}^{2} + \vert\vert ({\bf{Y}} + \gamma{\bf{I}})^{\alpha} \vert\vert_{{\bf{M}}_{\infty}^{-1}}^{2}
        \\&
        - \mathrm{tr}\big[({\bf{X}} + \delta{\bf{I}})^{\alpha}({\bf{M}} + \rho{\bf{I}})^{-1}({\bf{Y}} + \gamma{\bf{I}})^{\alpha}({\bf{I}} + {\bf{U}})
        \\&+
        ({\bf{I}} + {\bf{U}}^{*})({\bf{Y}} + \gamma{\bf{I}})^{\alpha}({\bf{M}} + \rho{\bf{I}})^{-1}({\bf{X}} + \delta{\bf{I}})^{\alpha}\big]\bigg\}.
    \end{split}
\end{equation}
Expanding the trace above leads to:
    \begin{equation}
        \begin{split}
            &\mathrm{tr}\big[({\bf{X}} + \delta{\bf{I}})^{\alpha}({\bf{M}} + \rho{\bf{I}})^{-1}({\bf{Y}} + \gamma{\bf{I}})^{\alpha}({\bf{I}} + {\bf{U}})
            +
            ({\bf{I}} + {\bf{U}}^{*})({\bf{Y}} + \gamma{\bf{I}})^{\alpha}({\bf{M}} + \rho{\bf{I}})^{-1}({\bf{X}} + \delta{\bf{I}})^{\alpha}\big].
            \\&
            = 2\mathrm{tr}\big[({\bf{Y}} + \gamma{\bf{I}})^{\alpha}({\bf{M}} + \rho{\bf{I}})^{-1}({\bf{X}} + \delta{\bf{I}})^{\alpha}\big] - \mathrm{tr}\big[ {\bf{U}}{\bf{U}}^{*}({\bf{Y}} + \gamma{\bf{I}})^{\alpha}({\bf{M}} + \rho{\bf{I}})^{-1}({\bf{X}} + \delta{\bf{I}})^{\alpha}\big],
        \end{split}
    \end{equation}
    where in the second term above we have used the fact that ${\bf{U}} + {\bf{U}}^{*} = -{\bf{U}}{\bf{U}}^{*}$. This follows directly Lemma 7 of \cite{MINH202225}. This polar decomposition $({\bf{Y}} + \gamma{\bf{I}})^{\alpha}({\bf{M}} + \rho{\bf{I}})^{-1}({\bf{X}} + \delta{\bf{I}})^{\alpha}$ in the first term above then leads to 
    \begin{equation}
        2\mathrm{tr}\big[({\bf{X}} + \delta{\bf{I}})^{\alpha}({\bf{M}} + \rho{\bf{I}})^{-1}({\bf{Y}} + \gamma{\bf{I}})^{2\alpha}({\bf{M}} + \rho{\bf{I}})^{-1}({\bf{X}} + \delta{\bf{I}})^{\alpha}\big]
    \end{equation}
    and the minimum is attained when ${\bf{U}} = 0$ in the second term above. Putting everything together leads to $(d^{\alpha}_{\mathrm{proM_{\infty}}}(({\bf{X}} + \delta{\bf{I}}), ({\bf{Y}} + \gamma{\bf{I}})))^{2}$ as in Definition \ref{generalisedAlphaProcrusteInfDim}.
\subsection{Proof of Corollary \ref{infDimW2}}\label{proofinfDimW2}
We first establish with a few definitions required for the generalization of Proposition 2 in \citep{HMJG2021b}. For $\delta, \rho \in \mathbb{R}_{>0}$, define the quantity $F_{\infty}(\mathbf{X} + \delta\mathbf{I}, \mathbf{Y} + \delta\mathbf{I})$ as 
\begin{equation}
    \mathrm{tr}\Big[\big((\mathbf{X} + \delta\mathbf{I})^{1/2}(\mathbf{M} + \rho\mathbf{I})^{-1}(\mathbf{Y} + \delta\mathbf{I})(\mathbf{M} + \rho\mathbf{I})^{-1}(\mathbf{X} + \delta\mathbf{I})^{1/2}\big)^{1/2}\Big].
\end{equation}
The functional $F_{\infty}$ is well-posed. Let $\mu, \nu \in \mathcal{P}(\mathcal{H})$. The infimum below is achieved because the set of couplings $\Gamma(\mu, \nu)$ is tight and weakly compact by virtue of Prokhorov's theorem. The functional $\bf{A} + \delta\bf{I} \mapsto \mathrm{tr}[(\bf{X}+ \delta\bf{I})(\bf{A} + \delta\bf{I})] + \mathrm{tr}[(\bf{M} + \rho\bf{I})^{-1}(\bf{Y}+ \delta\bf{I})(\bf{M} + \rho\bf{I})^{-1}(\bf{A} + \delta\bf{I})]$ is weakly lower semi-continuous in $(\bf{A} + \delta\bf{I})$. Finiteness follows from  $({\bf{X}} + \delta{\bf{I}}), ({\bf{Y}} + \gamma{\bf{I}}) \in \Sigma_{\mathcal{S}_{2}^{(\mathrm{EXT})}}(\mathcal{H})$ and $(\bf{M} + \rho\bf{I})^{-1}\in \mathcal{B}(\mathcal{H}$) with $||(\bf{M} + \rho\bf{I})^{-1}|| \leq \frac{1}{\rho}$, where $|| \cdot || $ represent the operator norm, so all the trace expressions are bounded by the HS norms of $\bf{X}$ and $\bf{Y}$. Then for any $(\mathbf{X} + \delta\mathbf{I}), (\mathbf{Y} + \gamma\mathbf{I}) \in \Sigma_{\mathcal{S}_{2}^{(\mathrm{EXT})}}(\mathcal{H})$ one writes:
\begin{equation}
    \begin{split}
        F_{\infty}(\mathbf{X} + \delta\mathbf{I}, \mathbf{Y} + \delta\mathbf{I}) := \min_{(\mathbf{A} + \delta\mathbf{I}) \in \Sigma_{\mathcal{S}_{2}^{(\mathrm{EXT})}}(\mathcal{H})} \frac{1}{2}\mathrm{tr}\big[&(\mathbf{X} + \delta\mathbf{I})(\mathbf{A} + \delta\mathbf{I}) \\ & + (\mathbf{M} + \rho\mathbf{I})^{-1}(\mathbf{Y} + \delta\mathbf{I})(\mathbf{M} + \rho\mathbf{I})^{-1}(\mathbf{A} + \delta\mathbf{I})\big]
    \end{split}
\end{equation}
and 
\begin{equation}
    \begin{split}
        & F_{\infty}(\mathbf{X} + \delta\mathbf{I}, \mathbf{Y} + \gamma\mathbf{I}) \\& := \min_{(\mathbf{A} + \delta\mathbf{I}) \in \Sigma_{\mathcal{S}_{2}^{(\mathrm{EXT})}}(\mathcal{H})}\big\{\mathrm{tr}\big[(\mathbf{X} + \delta\mathbf{I})(\mathbf{A} + \delta\mathbf{I})\big] \mathrm{tr}\big[(\mathbf{M} + \rho\mathbf{I})^{-1}(\mathbf{Y} + \delta\mathbf{I})(\mathbf{M} + \rho\mathbf{I})^{-1}(\mathbf{A} + \delta\mathbf{I}) \big] \big\}^{\frac{1}{2}}.
    \end{split}
\end{equation}
By virtue of the above it follows that when $\delta=0$, $F_{\infty}(\mathbf{X} + \delta\mathbf{I}, \mathbf{Y} + \delta\mathbf{I}) = F(\mathbf{X}, \mathbf{Y})$. The proof follows from \citep{BJL2019} with $(\mathbf{M} + \rho\mathbf{I})^{-1}$. Note that we do not consider the case when $\delta \neq \gamma$ as this represent a non-trivial situation that will be explored in a future work. For simplicity we denote
\begin{equation}\label{shortHandNotations}
    \bf{X}_{\infty} = (\bf{X}+ \delta\bf{I}); \,\,\,\,\,\,\,\,\,\,\bf{M}_{\infty} = (\bf{M}+ \rho\bf{I}); \,\,\,\,\,\,\,\,\,\, \bf{Y}_{\infty} = (\bf{Y}+ \delta\bf{I}). 
\end{equation}
Throughout this proof, in accordance with the extended Hilbert--Schmidt convention of Remark \ref{eHSGaussianConvention}, the symbol $\mathrm{tr}[\bf{M}_{\infty}^{-1}\bf{X}_{\infty}]$ is understood as the eHS-extended trace
$$
    \mathrm{tr}[\bf{M}_{\infty}^{-1}\bf{X}_{\infty}] := \mathrm{tr}[(\bf{M}+\rho\bf{I})^{-1}\bf{X}] + \delta^{2},
$$
and analogously for $\mathrm{tr}[\bf{M}_{\infty}^{-1}\bf{Y}_{\infty}]$ and $\mathrm{tr}[\bf{M}_{\infty}^{-1}\bf{K}_{\infty}]$. The first summand is a genuine trace on $\mathcal{S}_{1}(\mathcal{H})$ (well-defined since $\bf{X}\in\mathcal{S}_{2}(\mathcal{H})$, $(\bf{M}+\rho\bf{I})^{-1}\in\mathcal{B}(\mathcal{H})$, and the operator-ideal property $\mathcal{S}_{2}\cdot\mathcal{B}(\mathcal{H})\cdot\mathcal{S}_{2}\subseteq\mathcal{S}_{1}$), while the $\delta^{2}$ summand encodes the scalar slot of $\Sigma_{\mathcal{S}_{2}^{(\mathrm{EXT})}}(\mathcal{H})$. This avoids the divergence one would obtain from the naive trace $\mathrm{tr}[(\bf{M}+\rho\bf{I})^{-1}(\bf{X}+\delta\bf{I})] = \mathrm{tr}[(\bf{M}+\rho\bf{I})^{-1}\bf{X}] + \delta\,\mathrm{tr}[(\bf{M}+\rho\bf{I})^{-1}]$, whose second term is infinite on $\mathcal{H}$ infinite-dimensional. Reducing to $\delta=0$ and $\bf{X},\bf{Y}\in\mathcal{S}_{1}(\mathcal{H})$ recovers the classical Borel-Gaussian trace formulae of \cite{Minh2021,HMJG2021b}. \\

For a coupling $\gamma \in \Gamma(\mu, \nu)$ we expand the quadratic cost defining $\widetilde{\mathcal{W}}_{2, \infty}$
\begin{equation}\label{infDimWasserteinIntegral}
    \begin{split}
        \int_{\mathcal{H}\times \mathcal{H}}|| \bf{x} -\bf{y}||^{2}_{\bf{M}_{\infty}^{-1}}d\gamma
        =\int_{\mathcal{H}}||\bf{x}||^{2}_{\bf{M}_{\infty}^{-1}}d\gamma  + \int_{\mathcal{H}}||\bf{y}||^{2}_{\bf{M}_{\infty}^{-1}}d\gamma - 2\int_{\mathcal{H}}\langle \bf{M}_{\infty}^{-1}\bf{x}, \bf{y}\rangle_{\mathcal{H}}d\gamma, 
    \end{split}
\end{equation}
where $\langle\cdot, \cdot\rangle$ is the inner product on $\mathcal{H}$. Since the marginals of $\gamma$ are $\mu$ and $\nu$, we write:
\begin{equation}
    \int_{\mathcal{H}}||\bf{x}||^{2}_{\bf{M}_{\infty}^{-1}}d\gamma = \text{tr}(\bf{M}_{\infty}^{-1}\bf{X}_{\infty}) \,\,\,\,\,\,\,\,\,\,\text{and} \,\,\,\,\,\,\,\,\,\,\int_{\mathcal{H}}||\bf{y}||^{2}_{\bf{M}_{\infty}^{-1}}d\gamma = \text{tr}(\bf{M}_{\infty}^{-1}\bf{Y}_{\infty}). 
\end{equation}
We proceed to justify the above trace formulae. We note that for any $\bf{M}_{\infty} \in \mathcal{B}(\mathcal{H})$ and $\mathcal{X}\in \mathcal{S}_{1}\cap\mathcal{S}_{2}^{(\text{EXT})}(\mathcal{H})$ ensures that $\bf{M}_{\infty}^{-1}\mathcal{X} \in \mathcal{S}_{1}\cap\mathcal{S}_{2}^{(\text{EXT})}(\mathcal{H})$ such that
\begin{equation}
    \int_{\mathcal{H}}||\bf{x}||^{2}_{\bf{M}_{\infty}^{-1}}\,d\mathcal{N}(0, \mathcal{X}) = \text{tr}(\bf{M}_{\infty}^{-1}\mathcal{X})
\end{equation}
by the Gaussian second moment formula, licit for bounded $\bf{M}_{\infty}^{-1}$ and trace-class $\mathcal{X}$. Here, "trace-class $\mathcal{X}$" refers to the trace-class part $\mathbf{X}\in\mathcal{S}_{1}\cap\mathcal{S}_{2}^{(\mathrm{EXT})}$ of $\mathbf{X}_{\infty}=\mathbf{X}+\delta\mathbf{I}$ (the full operator $\mathbf{X}_{\infty}$ is not itself trace-class on infinite-dimensional $\mathcal{H}$ when $\delta>0$), and the corresponding integral is read in the eHS sense of Remark \ref{eHSGaussianConvention}: $\int_{\mathcal{H}}\|\mathbf{x}\|^{2}_{\mathbf{M}_{\infty}^{-1}}\,d\gamma := \mathrm{tr}[(\mathbf{M}+\rho\mathbf{I})^{-1}\mathbf{X}]+\delta^{2}$, consistent with the shorthand convention introduced after (\ref{shortHandNotations}).\\

Thanks to the infinite dimensional generalization of $\mathcal{W}_{2}$ in ~\cite[Theorem 3.5]{Gelbrich1990}, and the trace definitions above we can rewrite Eq. (\ref{infDimWasserteinIntegral}) as 

\begin{equation}\label{infdimWexpression0}
    \begin{split}
        \widetilde{\mathcal{W}}^{2}_{2, \infty}(\mu, \nu) = \mathrm{tr}\big[\bf{M}_{\infty}^{-1}\bf{X}_{\infty}] + \mathrm{tr}\big[\bf{M}_{\infty}^{-1}\bf{Y}_{\infty}\big]   - 2 \sup_{\mathbf{K}_{\infty}} \mathrm{tr}\big[(\mathbf{M}_{\infty})^{-1} \mathbf{K}_{\infty} \big]\}, 
    \end{split}
\end{equation}
where $\mathbf{K}_{\infty} = (\bf{K} + \delta\bf{I})$ is the covariance matrix between $(\mathbf{X} + \delta\mathbf{I})$ and $(\mathbf{Y} + \delta\mathbf{I})$ such that the joint covariance matrix is 
\begin{equation}
    J = \begin{bmatrix}
            (\mathbf{X} + \delta\mathbf{I}) & \mathbf{K}_{\infty} \\
            \mathbf{K}_{\infty}^{*} & (\mathbf{Y} + \delta\mathbf{I})
        \end{bmatrix} \succeq 0. 
\end{equation}
It follows that:

\begin{equation}\label{infdimWexpression1}
    \begin{split}
        \widetilde{\mathcal{W}}^{2}_{2, \infty}(\mu, \nu) = \mathrm{tr}\big[(\mathbf{M} + \rho\mathbf{I})^{-1}(\mathbf{X} + \delta\mathbf{I}) \big] + \mathrm{tr}\big[(\mathbf{M} + \rho\mathbf{I})^{-1}(\mathbf{Y} + \delta\mathbf{I}) \big]  - \sup_{\mathbf{K}_{\infty}: J \succeq 0} 2 \mathrm{tr}\big[(\mathbf{M} + \rho\mathbf{I})^{-1} \mathbf{K}_{\infty} \big]\}, 
    \end{split}
\end{equation}
We identify two necessary and sufficient conditions for $J \succeq 0$: 

\begin{enumerate}
    \item $(\mathbf{X} + \delta\mathbf{I}) \succeq \mathbf{K}_{\infty} (\mathbf{Y} + \delta\mathbf{I})^{-1}\mathbf{K}_{\infty}^{*}$;

    \item $\mathbf{K}_{\infty} =  (\mathbf{X} + \delta\mathbf{I})^{1/2}C_{\infty}(\mathbf{Y} + \delta\mathbf{I})^{1/2}$,
\end{enumerate}
where $C_{\infty}$ is some contraction on ${\mathcal{S}_{2}^{(\mathrm{EXT})}}(\mathcal{H})$. We proceed to compute the trace $\mathrm{tr}\big[(\mathbf{M} + \rho\mathbf{I})^{-1} \mathbf{K}_{\infty} \big]$. Using the short hand notations (\ref{shortHandNotations}), we will show that for any $\mathbf{K}_{\infty} =  \mathbf{X}_{\infty}^{1/2}C_{\infty}\mathbf{Y}_{\infty}^{1/2}$ with $||C_{\infty}|| \leq 1$ where $||\cdot||$ is the operator norm and any $\mathbf{A}_{\infty} = (\mathbf{A}+ \delta\mathbf{I}) \in \Sigma_{\mathcal{S}_{2}^{(\text{EXT})}}(\mathcal{H})$ the trace $\text{tr}(\bf{M}_{\infty}^{-1}\mathbf{K}_{\infty})$ is bounded above by:
\begin{equation}
    \text{tr}(\mathbf{M}_{\infty}^{-1}\mathbf{K}_{\infty}) \leq \big\{\text{tr}(\mathbf{X}_{\infty}\mathbf{A}_{\infty})\text{tr}(\mathbf{M}_{\infty}^{-1}\mathbf{Y}_{\infty}\mathbf{M}_{\infty}^{-1}\mathbf{A}_{\infty}^{-1}) \big\}^{1/2}. 
\end{equation}
The trace of $\mathbf{M}_{\infty}^{-1}\mathbf{K}_{\infty}$ is well-defined. This follows from the fact that $\mathbf{M}_{\infty}^{-1}\mathbf{K}_{\infty}\in \mathcal{S}_{1}\cap \mathcal{S}_{2}^{\text{(EXT)}}$. After some algebra, we write $\text{tr}(\mathbf{M}_{\infty}^{-1}\mathbf{K}_{\infty}) \leq [\text{tr}(\mathbf{M}_{\infty}^{-1} \mathbf{X}_{\infty})\cdot\text{tr}(\mathbf{M}_{\infty}^{-1} \mathbf{Y}_{\infty})]^{1/2}$. Then define the block matrix $\widetilde{P} = [\mathbf{A}_{\infty}^{1/2}\, 0; 0 \, \mathbf{A}_{\infty}^{1/2}]$. Since  $J \succeq 0$ and $\widetilde{P}$ is invertible then $\widetilde{P}J\widetilde{P} \succeq 0$. Then the block has trace $\text{tr}(\mathbf{M}_{\infty}^{-1}\mathbf{K}_{\infty})$. It follows that
\begin{equation}
    \sup_{\mathbf{K}_{\infty}}\text{tr}(\mathbf{M}_{\infty}^{-1}\mathbf{K}_{\infty}) = \inf_{\mathbf{A}_{\infty}} \big\{\text{tr}(\mathbf{X}_{\infty}\mathbf{A}_{\infty})\text{tr}(\mathbf{M}_{\infty}^{-1}\mathbf{Y}_{\infty}\mathbf{M}_{\infty}^{-1}\mathbf{A}_{\infty}^{-1}) \big\}^{1/2} =: F_{\infty}(\mathbf{X}_{\infty}, \mathbf{Y}_{\infty}).
\end{equation}
Substituting by the optimal $\widetilde{A}_{\infty}$ recovers the relation $\widetilde{\mathcal{W}}^{2}_{2, \infty}(\mu, \nu) = d^{2}_{\text{GBW}_\infty}(\mathbf{X}_{\infty}, \mathbf{Y}_{\infty})$. This entails the proof. 
\begin{corollary}\label{genInfDimWassLimitCorr}
    Let $\mathcal{H}$ be an infinite dimensional separable Hilbert space and let $\mu, \nu \in \mathcal{P}(\mathcal{H})$, where $\mu = \mathcal{N}(0, \bf{X} + \delta\bf{I})$ and $\nu = \mathcal{N}(0, \bf{Y} + \delta\bf{I})$ with covariance operators as $({\bf{X}} + \delta{\bf{I}}), ({\bf{Y}} + \gamma{\bf{I}}) \in \Sigma_{\mathcal{S}_{2}^{(\mathrm{EXT})}}(\mathcal{H})$. Taking the limits in the traces above is licit thanks to theorems established in Appendix \ref{firstConvergenceStudies} 
    \begin{equation}
        \widetilde{\mathcal{W}}_{2, \infty}(\mu, \nu) = \lim_{n\rightarrow \infty} \widetilde{\mathcal{W}}_{2, \infty}(\mu_{n}, \nu_{n}). 
    \end{equation}
\end{corollary}
\subsection{Proof of Corollary \ref{corrInfGenLogHS}}\label{proofCorrInfGenLogHS}
We first assume that $\mathbf{M}_{\infty}$ is spectrally aligned with $(\mathbf{X}+ \delta\mathbf{I})$ and $(\mathbf{Y}+ \delta\mathbf{I})$. This assumption is motivated by experiments carried out in Section \ref{generalisedLESDetails}. Recall that $\log(\mathbf{X}+ \delta\mathbf{I})$ is well-defined and bounded thanks to \cite{Minh2014, Minh2021}. This entails that $[\log(\mathbf{X}+ \delta\mathbf{I}) - \log(\mathbf{Y}+ \delta\mathbf{I})] \in \mathcal{S}_{2}(\mathcal{H})$. Therefore, the norm $||\log(\mathbf{X}+ \delta\mathbf{I}) - \log(\mathbf{Y}+ \delta\mathbf{I})||_{\mathbf{M}_{\infty}^{-1}}<\infty$ is finite. The limit $\alpha \rightarrow 0$ is carried in a similar manner as for the finite dimensional case by expanding $(\mathbf{X} + \delta\mathbf{I})^{\alpha} = \exp[\log(\mathbf{X} + \delta\mathbf{I})]$. For small $\alpha$, one has that $((\mathbf{X}+ \delta\mathbf{I})^{\alpha} - \mathbf{I})/ \alpha$ converges to $\log(\mathbf{X}+ \delta\mathbf{I})$ in operator norm since the eigenvalues satisfies $(\lambda^{(\mathbf{X})}_{k} + \delta)^{\alpha} = \exp(\alpha\log(\lambda^{(\mathbf{X})}_{k} + \delta))$ and 
\begin{equation}
    \sup_{k}\big|((\lambda^{(\mathbf{X})}_{k} + \delta) - 1)/\alpha - \log(\lambda^{(\mathbf{X})}_{k} + \delta))\big| \rightarrow 0 \,\,\,\,\,\,\, \text{as} \,\,\,\,\,\,\, \alpha \rightarrow 0
\end{equation}
uniformly in $k$, as $\log(\lambda^{(\mathbf{X})}_{k} + \delta) \in [\log \delta, \log(||\mathbf{X}|| + \delta)]$ is bounded. This settles the convergence in operator norm and one then recovers the limit in (\ref{infDimGenLHSEq}).

\section{Proofs for Section \ref{applications}}\label{applicationsProof}
\subsection{Lower semi-continuity of the robust GBW}\label{proofLowerSemCont}
We formalize the proof of the robust GBW \citep{HMJG2021b} by rigorously establishing the lower semi-continuity (lsc) property of the mapping
\begin{equation}
    \gamma \mapsto \inf_{\gamma \in \Gamma(\mu, \nu)} \int \| \mathbf{W}^\top(x - y) \|^2 \, d\gamma(x, y) 
\end{equation}
as the generalization of this property for the infinite dimensional case (\textit{i.e}: Proposition \ref{infDimRGBW}) will follow from latter. Let $\mu, \nu \in \mathcal{P}(\mathbb{R}^n)$: the space of $n$ dimensional Borel probability measures with finite second moments, equipped with the Wasserstein-2 topology, \textit{i.e.}, in (\ref{W_2Distance}). Let 
$$\mathrm{St}(d, n) := \{\mathbf{W}\in\mathbb{R}^{n\times d}: \mathbf{W}^{\top}\mathbf{W} = \mathbf{I}_{d}\}$$
denote the Stiefel manifold of orthonormal $d$-frames in $\mathbb{R}^{n}$, where $d<n$. Define the functional
\begin{equation}
    F(\mu, \nu, \mathbf{W}) := \inf_{\gamma \in \Gamma(\mu, \nu)} \int \| \mathbf{W}^\top(x - y) \|^2 \, d\gamma(x, y).
\end{equation}
We aim to show that if $(\mu_k, \nu_k, \mathbf{W}_k) \to (\mu, \nu, \mathbf{W})$ in $\mathcal{P} \times \mathcal{P} \times \mathbb{R}^{n\times d} \, (d < n)$ then
\begin{equation}
    \liminf_{k \to \infty} F(\mu_k, \nu_k, \mathbf{W}_k) \geq F(\mu, \nu, \mathbf{W}). 
\end{equation}

\textbf{Continuity of the cost function:} let
$$c_{\mathbf{W}}(x, y) := \| \mathbf{W}^\top (x - y) \|^2 = (x - y)^\top \mathbf{W} \mathbf{W}^\top (x - y) $$
This cost is:
\begin{itemize}
    \item Continuous in $x, y$ for fixed $\mathbf{W}$;
    \item Continuous in $\mathbf{W}$ for fixed $x, y$.
\end{itemize}
This ensures that $c_{\mathbf{W}_k} \to c_{\mathbf{W}}$ uniformly on compact sets as $\mathbf{W}_k \to \mathbf{W}$.\\ 

\textbf{Lower semicontinuity of the optimal transport cost:} From classical results in optimal transport ~\cite[Theorem 5.20]{villani2009}, we know that for continuous cost $c$, the map $$ (\mu, \nu) \mapsto \inf_{\gamma \in \Gamma(\mu, \nu)} \int c(x, y) d\gamma(x, y) $$ is lower semi-continuous with respect to weak convergence, if $c$ is bounded from below and lower semi-continuous. Here
\begin{itemize}
    \item $c_{\mathbf{W}}(x, y) \geq 0$;
    \item $c_{\mathbf{W}}$ is continuous $\implies$ l.s.c;
    \item $\gamma_k \in \Gamma(\mu_k, \nu_k)$ where $(\mu_k, \nu_k) \to (\mu, \nu)$.
\end{itemize}
So, 
$$
\liminf_{k \to \infty} \inf_{\gamma \in \Gamma(\mu_k, \nu_k)} \int c_{\mathbf{W}_k}(x, y) \, d\gamma(x, y) \geq \inf_{\gamma \in \Gamma(\mu, \nu)} \int c_{\mathbf{W}}(x, y) \, d\gamma(x, y). 
$$
Hence $F$ is lower semi-continuous. Finally, since $\mathfrak{P}_d(\mu, \nu) = \sup_{\mathbf{W} \in \mathrm{St}(d, n)} F(\mu, \nu, \mathbf{W})$, and each $F(\mu, \nu, \mathbf{W})$ is lower semi-continuous in $(\mu, \nu)$, we have:
$$
(\mu_k, \nu_k) \to (\mu, \nu) \quad \Rightarrow \quad \liminf_{k} \mathcal{P}_d(\mu_k, \nu_k) \geq \mathcal{P}_d(\mu, \nu)
$$
because the supremum of lower semi-continuous functions indexed over a compact set $\mathrm{St}(d, n)$ remain lower semi-continuous. \\

\textbf{Equivalence between RGBW and GBW}: Thanks to the last argument above and to the compactness of $\Gamma(\mu, \nu)$, one obtains the following structured max-min optimization problem if $\mathbf{W}^{*}$ is an optimal solution \citep[Eq.~(22)]{HMJG2021b}: 
\begin{equation}
    \mathfrak{P}_d(\mu, \nu) = \max_{\mathbf{M}^{-1} \in \,\mathcal{C}}d^{2}_{\mathrm{GBW}}(\mathbf{X}, \mathbf{Y}), 
\end{equation}
for $\mathbf{M}^{-1} = \mathbf{W}^{*}(\mathbf{W}^{*})^{\top}$ and where the maximum above is taken with respect to $\mathbf{M}^{-1}$ over the closed convex set $\mathcal{C} \subseteq \mathbb{S}^{n}_{++}$ leading to \citep[Eq. ~(23)]{HMJG2021b}. We proceed to generalise those arguments in the section below. 
\subsection{Proof of Proposition \ref{infDimRGBW}}\label{proofinfDimRGBW}
We begin this proof by establishing the following technical Lemma which follows from \cite{brezis2011functional} (Theorem 3.7). We reproduce the proof for clarity. 
\begin{lemma}\label{infDimClosedCnxSet}
    Let $\mathcal{H}$ be an infinite dimensional separable Hilbert space. \, Let $\widetilde{\mathcal{C}} \subseteq \Sigma_{\mathcal{S}_{2}^{(\mathrm{EXT})}}(\mathcal{H})$, where $\Sigma_{\mathcal{S}_{2}^{(\mathrm{EXT})}}(\mathcal{H})$ represents the space of \textbf{generalized} extended unitized Hilbert-Schmidt operators. $\widetilde{\mathcal{C}}$ is a closed and convex set. Then, $\widetilde{\mathcal{C}}$ is weakly sequentially closed. 
\end{lemma}
\begin{proof}
    Let $\{x_{k}\}$ be a sequence in $\widetilde{\mathcal{C}}$ and suppose $x_{k} \rightharpoonup x^{\prime}$. There exists an operator $\Pi$ of $x^{\prime}$ such that $x^{\prime} = \Pi(x^{\prime})$ into the closed convex set $\widetilde{\mathcal{C}}$. Thus, by virtue of the variational inequality
    \begin{equation}
        \langle x^{\prime} - \Pi(x^{\prime}), x_{k} - \Pi(x^{\prime})  \rangle_{\mathcal{S}_{2}^{(\text{EXT})}} \leq 0 \,\,\, \forall \,\,\, k, 
    \end{equation}
    it follows that since $x_{k} \rightharpoonup x^{\prime}$,
    \begin{equation}
        ||  x^{\prime} - \Pi(x^{\prime})||_{\mathcal{S}_{2}^{(\text{EXT})}}^{2} = \langle x^{\prime} - \Pi(x^{\prime}), x^{\prime} - \Pi(x^{\prime})  \rangle_{\mathcal{S}_{2}^{(\text{EXT})}} = \lim_{k\rightarrow \infty} \langle x^{\prime} - \Pi(x^{\prime}), x_{k} - \Pi(x^{\prime})  \rangle_{\mathcal{S}_{2}^{(\text{EXT})}}.  
    \end{equation}
    Consequently, we have that $||  x^{\prime} - \Pi(x^{\prime})|| = 0$ which entails the claim. 
\end{proof}
\begin{lemma}\label{lscFunctional}
    Let $\mathcal{H}$ be an infinite dimensional separable Hilbert space and $\mu, \nu \in \mathcal{P}(\mathcal{H})$ be zero-centered Gaussian measures with trace-class covariance operators. The functional
    \begin{equation}
        F_{\infty}(\mu, \nu, \Pi) := \inf_{\gamma \in \Gamma(\mu, \nu)} \int \|\Pi(x - y)\|^2\, d\gamma(x, y)
    \end{equation}
    is lower semi-continuous and convex. Then, $F_{\infty}$ is also weakly lower semi-continuous. 
\end{lemma}
\begin{proof}
    Since $F_{\infty}$ is convex it follows that $\mathrm{epi}(F_{\infty})$ is convex as well. 
    To establish lower semicontinuity of \(F_{\infty}(\Pi)\) with $\gamma$ fixed, take any weakly converging sequence \(\Pi_k \rightharpoonup \Pi\) in $\widetilde{\mathcal{C}}$ which is licit by virtue of Lemma \ref{infDimClosedCnxSet}. Therefore, the map 
    \[(x, y, \Pi) \mapsto \| \Pi(x - y) \|^2\]
    is lower semi-continuous. We apply a standard epi-convergence argument to get
    \begin{equation}
        \liminf_{k \to \infty} F_{\infty}(\Pi_k) \geq F_{\infty}(\Pi).
    \end{equation}
    Moreover, because $F_{\infty}$ is lower semi-continuous, $\mathrm{epi}(F_{\infty})$ is closed. By virtue of the Lemma \ref{infDimClosedCnxSet} we have that $\mathrm{epi}(F_{\infty})$ is weakly sequentially closed, which implies that $F_{\infty}$ is weakly sequentially lower semi-continuous. 
\end{proof}
\begin{proof} (\textbf{of Proposition \ref{infDimRGBW}}) Let             $\widetilde{\mathcal{C}} := \{(\mathbf{A} +                \rho\mathbf{I})^{-1}: \mathbf{A}\in \mathcal{S}_{2}(\mathcal{H}),          \rho > 0, ||\mathbf{A}||_{\mathcal{S}_{2}(\mathcal{H}}\leq R\}$ for some radius $R$. From (\ref{infDimPRW}) we have that
    \begin{equation}\label{eqRgbwInfty}
        \begin{split}                \mathfrak{P}_{\widetilde{\mathcal{C}}}(\mu, \nu) = \sup_{\Pi\in\mathcal{B}(\mathcal{H}),\,\Pi\Pi^{*}\in\widetilde{\mathcal{C}}} \mathrm{tr}(\Pi\Pi^{*}(\mathbf{X} & + \delta\mathbf{I})) + \mathrm{tr}(\Pi\Pi^{*}(\mathbf{Y} + \delta\mathbf{I})) \\ &- 2\mathrm{tr}\big[({\bf{X}} + \delta{\bf{I}})^{1/2}\Pi\Pi^{*}({\bf{Y}} + \delta{\bf{I}})\Pi\Pi^{*}({\bf{X}} + \delta{\bf{I}})^{1/2}\big]^{1/2}
         \end{split}
    \end{equation}
    All traces appearing in (\ref{eqRgbwInfty}) are read in the eHS-extended sense of Remark \ref{eHSGaussianConvention}: for instance, $\mathrm{tr}[\Pi\Pi^{*}(\mathbf{X}+\delta\mathbf{I})] := \mathrm{tr}[\Pi\Pi^{*}\mathbf{X}] + \delta^{2}$, so that the scalar slots of $\Sigma_{\mathcal{S}_{2}^{(\mathrm{EXT})}}(\mathcal{H})$ are absorbed into the cross-term identity that links (\ref{eqRgbwInfty}) to (\ref{rgbwInftyGbwInfty}); the raw operator trace would diverge unless $\Pi\Pi^{*}$ were itself trace-class.
    Consequently, the term $\mathrm{tr}\big[({\bf{X}} + \delta{\bf{I}})^{1/2}\Pi\Pi^{*}({\bf{Y}} - \delta{\bf{I}})\Pi\Pi^{*}({\bf{X}} + \delta{\bf{I}})^{1/2}\big]^{1/2}$ can be re-expressed as $\mathrm{tr}\big[\Pi^{*}({\bf{X}} + \delta{\bf{I}})\Pi\Pi^{*}({\bf{Y}} - \delta{\bf{I}})\Pi\big]^{1/2}$.$\mathcal{P}_{\widetilde{\mathcal{C}}}(\mu, \nu)$ then coincides with $ d^{2}_{\mathrm{GBW}_{\infty}}[({\bf{X}} + \delta{\bf{I}}), ({\bf{Y}} + \delta{\bf{I}})]$ for $({\bf{M}} + \rho{\bf{I}})^{-1} = \Pi^{*}(\Pi^{*})^{*} \in \mathcal{B}(\mathcal{H})$, where $\Pi^{*}$ is the maximizer of (\ref{eqRgbwInfty}) on the closed convex set $\widetilde{\mathcal{C}}$. The existence of $\Pi^{*}$ follows directly from Lemma \ref{infDimClosedCnxSet} and \ref{lscFunctional}. This together with boundedness of $\widetilde{\mathcal{C}}$ and compactness of the transport plan $\Gamma(\mu, \nu)$ (with respect to the weak topology in infinite dimensions) implies that the infimum is a minimum. Consequently, (\ref{rgbwInftyGbwInfty}) immediately follows from the following structured max-min problem:
    \begin{equation*}\label{infDimPRW1}
        \mathfrak{P}_{\widetilde{\mathcal{C}}}(\mu, \nu) = \max_{({\bf{M}} + \rho{\bf{I}})^{-1} \in \widetilde{\mathcal{C}}} \min_{\gamma \sim \Gamma(\mu, \nu)}\int ||\Pi(x -y) ||^{2}\,d\gamma(x, y).
    \end{equation*}
   for $\widetilde{\mathcal{C}} \subseteq \Sigma_{\mathcal{S}_{2}^{(\mathrm{EXT})}}(\mathcal{H})$ which entails the claim. 
\end{proof}

\section{Generalized Log-Euclidean Signature details}\label{generalisedLESDetails}
We provide some technical pertaining the estimation of the truncated spectrum of our infinite dimensional operators. This is a computationally expensive task and finding the optimal truncation is key to accelerating the learning and simulations presented in Section \ref{applications}. The approach in \citep{shnitzer2022} relies on the spectral approximation technique underlying the highly stable truncation of the Nyström method for low-rank ($K < N$) positive semidefinite (PSD) matrix approximation devised in \citep{tropp2017fixed}. This allows for a rigorous approximation of the top $K$ eigenvalues of a PSD matrix $\mathbf{X} \in \mathbb{R}^{n \times n}$. The bound pertaining to the eigenvalue approximation error of $\mathbb{E}[\sum_{i=1}^{K}|\lambda_{i}^{(\mathbf{X})} - \lambda_{i}^{(\mathbf{X}_{K})}|]$ is established in \citep{shnitzer2022} by virtue of theorems from matrix analysis \citep{Bhatia2007}. We proceed to establish bounds that will allow us to estimate our approximation errors with respect to the Mahalanobis norm by generalizing those results. The rest follows immediately from part 2 of ~\cite[Proposition 2]{shnitzer2022}. 
\subsection{Bounds on generalized Log-Euclidean signature distance and spectrum estimations}\label{generalisedLESDetailsResults}
The Mahalanobis norm naturally induces a spectrally weighted Schatten-2 norm, that we can leverage to derive a generalized error bound in this setting. We employ the results in  ~\cite[Corollary 8.2]{tropp2023randomized} and follows the strategy behind the proof of ~\cite[Proposition 2]{shnitzer2022}. 
\begin{proposition}\label{rankXFiniteDim}
    The real symmetric PSD matrix $\mathbf{X}$ admits a rank-$K$ approximation $\mathbf{X}_{K}$  which satisfies the following eigenvalue error bounds
    \begin{equation}
        \begin{split}
            \mathbb{E}[|| \mathbf{X} - \mathbf{X}_{K}||_{\mathbf{M}^{-1}}] & = \mathbb{E}\bigg[\sum_{i=1}^{K}\bigg|\frac{1}{\omega_{i}^{(\mathbf{M})}}\bigg(\lambda_{i}^{(\mathbf{X})} - \lambda_{i}^{(\mathbf{X}_{K})}\bigg)\bigg|\bigg]  \\ & \leq \bigg (\sum_{i=1}^{N_{\mathbf{M}}}\frac{1}{\omega_{i}^{(\mathbf{M})}} \bigg) \cdot  \bigg\{\bigg(\sum_{i = K + 1}^{N_{\mathbf{X}}}(\lambda_{i}^{(\mathbf{X})})^{2}\bigg)^{1/2} + \frac{K}{M-K-1}\cdot \sum_{i = K + 1}^{N_{\mathbf{X}}}\lambda_{i}^{(\mathbf{X})} \bigg\}
        \end{split}
    \end{equation}
    where $\omega_{i}^{(\mathbf{M})}$ represent the eigenvalues of $\mathbf{M}$. $M$ is denotes the number of random vectors \citep{tropp2017fixed, shnitzer2022}. \\

    Furthermore, the generalized log-eigenvalue error is bounded by:
    \begin{equation}
        \begin{split}
             \mathbb{E} \bigg[\bigg( \sum_{i=1}^{K}\bigg| \frac{1}{(\omega_{i}^{(\mathbf{M})} + \rho)}\big[\log\big(\lambda_{i}^{({\bf{X}})} + \delta\big) & - \log\big(\lambda_{i}^{(\mathbf{X}_{K})} + \delta\big)\big]^{2} \bigg|\bigg)^{1/2}\bigg] \\ &
             \leq \frac{1.5}{\lambda_{K}^{(\mathbf{X})}+ \delta}\cdot \frac{1}{\omega_{K}^{(\mathbf{M})}+ \rho}\cdot \alpha(\omega^{(\mathbf{M})}, \lambda^{(\mathbf{X})}),
        \end{split}
    \end{equation}
    where $\alpha(\omega^{(\mathbf{M})}, \lambda^{(\mathbf{X})})$ is:
    \begin{equation}
        \bigg(\sum_{i=K+1}^{N_{\mathbf{M}}}\frac{1}{(\omega_{i}^{(\mathbf{M})} + \rho)}\cdot \bigg[ \bigg(\sum_{i = K + 1}^{N_{\mathbf{X}}}(\lambda_{i}^{(\mathbf{X})})^{2}\bigg)^{1/2} + \frac{K}{M-K-1}\cdot \sum_{i = K + 1}^{N_{\mathbf{X}}}\lambda_{i}^{(\mathbf{X})} \bigg]\bigg)^{1/2}
    \end{equation}
\end{proposition}
\begin{proof}
    We assume that $\omega_{i} \propto \lambda_{i}$, \textit{i.e}: $\mathbf{M}$ and $\mathbf{X}_{K}$ share the same eigenbasis. The proof follows from \citep{shnitzer2022} and \citep{tropp2023randomized} after rewriting the Mahalanobis norm as a spectrally weighted Schatten-2 norm. 
\end{proof}
The above result rigorously quantify errors when when estimating a PSD matrix $\mathbf{X}$ via a low-rank Nyström approximation $\mathbf{X}$ under a spectrally weighted Schatten-2 norm induced by the Mahalanobis norm. Proposition \ref{rankXFiniteDim} shows that when $\omega_{i} \propto \lambda_{i}$, the weighted approximation error admits an interpretable bound that reflects both the tail spectrum ${\lambda_{i>K}^{(\mathbf{X})}}$ and decay of the weighting spectrum $\omega_{i}^{(\mathbf{M})}$. This structure carries over to a bound on the generalized Log-Euclidean signature error. Furthermore, the above bound is particularly necessary as it allows us to fix $\rho$ and $N_{\mathbf{M}}$ on top of fixing $M$ in Proposition \ref{rankXFiniteDim}. \\

However, the case where $\omega_{i} \, \cancel{\propto}\, \lambda_{i}$ is significantly more challenging. In this general case, the Mahalanobis norm cannot be jointly diagonalized with $\mathbf{X}$, and thus the spectral weights no longer commute with the approximation. This prevents the decoupling of the spectral decay and the weighting in above bound. Intuitively, in experiments, this can lead to artifacts: even if $\mathbf{X}_{K}$ is well approximated, a poorly aligned $\mathbf{M}$ may emphasize directions that are not well-captured, inflating the error. Understanding and bounding this misalignment case would require a more intricate analysis involving subspace perturbation bounds or non-commutative trace inequalities —a direction that remains open and nontrivial. We provide experiments where we optimize $\mathbf{M}$ for fixed $\rho$ to compensate for this over a number of eigenvalues. A detailed analysis is left for future work. 
\subsection{LES results for various sample sizes}
In the figure below we reproduce some plots from \citep{shnitzer2022}  for various sample sizes for the sake of clarity as we perform experiments for the generalized cases and compare with LES for these particular sample sizes. 

% Side-by-side figures
\begin{figure}[H]
\centering 
\begin{minipage}{.4\textwidth}
  \centering
  \includegraphics[width=1.1\linewidth]{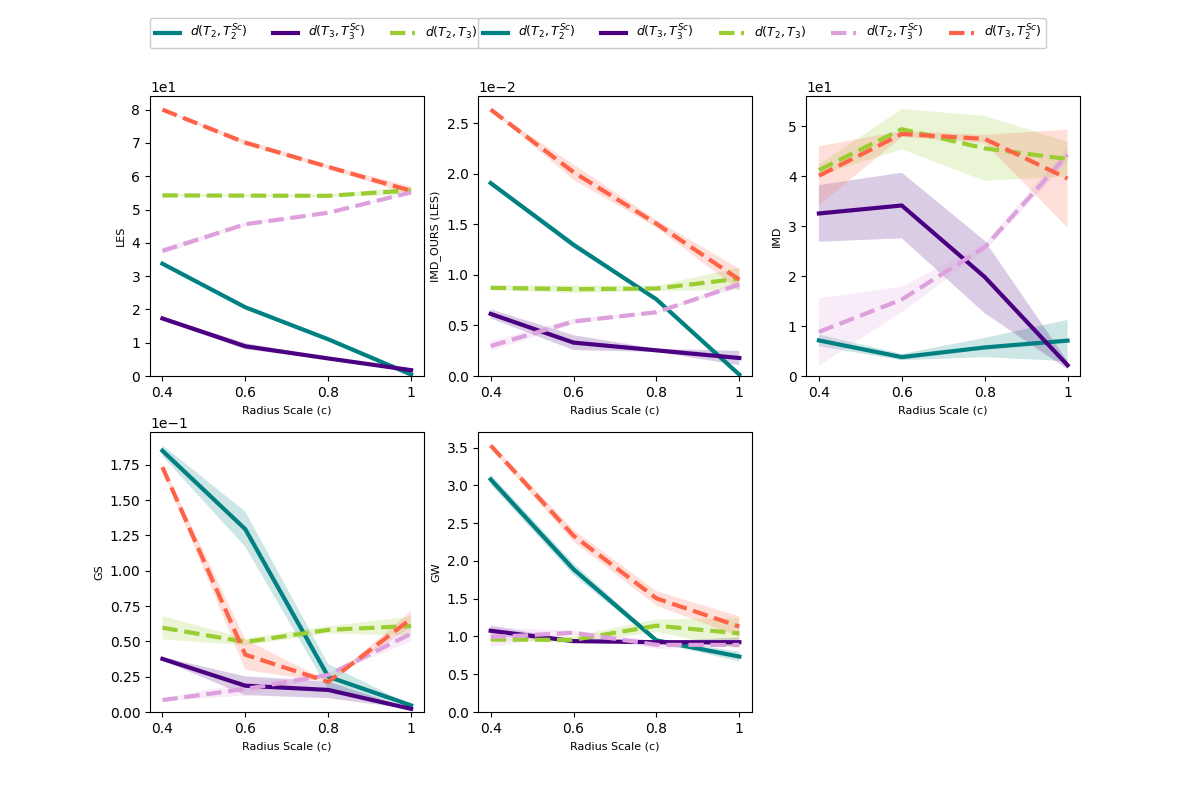}
  %\captionof{figure}{A figure}
  \label{fig:test1}
\end{minipage}%
\begin{minipage}{.4\textwidth} 
  \centering
  \includegraphics[width=1.1\linewidth]{img/section_4/tori_comparisons_les_NEV_200_N_2000.png}
  %\captionof{figure}{Another figure}
  \label{fig:test2}
\end{minipage}
\caption{Comparing distance measures on 2D and 3D tori data with radii related by scaling c for $N = 1000$ and $2000$ by reproducing the algorithm from \citep{shnitzer2022}. Experiments become computationally expensive for sample sizes exceeding $N = 2000$.}
\end{figure}
\subsection{Generalized LES experiments with $\omega_{i} \propto \lambda_{i}$}
We produce some more plots for various values of $\rho$ and gradually increase the number samples $N$. In doing so we aim to show that there might be an optimal range for $\rho$ in high dimensions. 
\subsubsection{$N = 1000$}
\begin{figure}[H]
    \centering
    \includegraphics[width=0.9\linewidth]{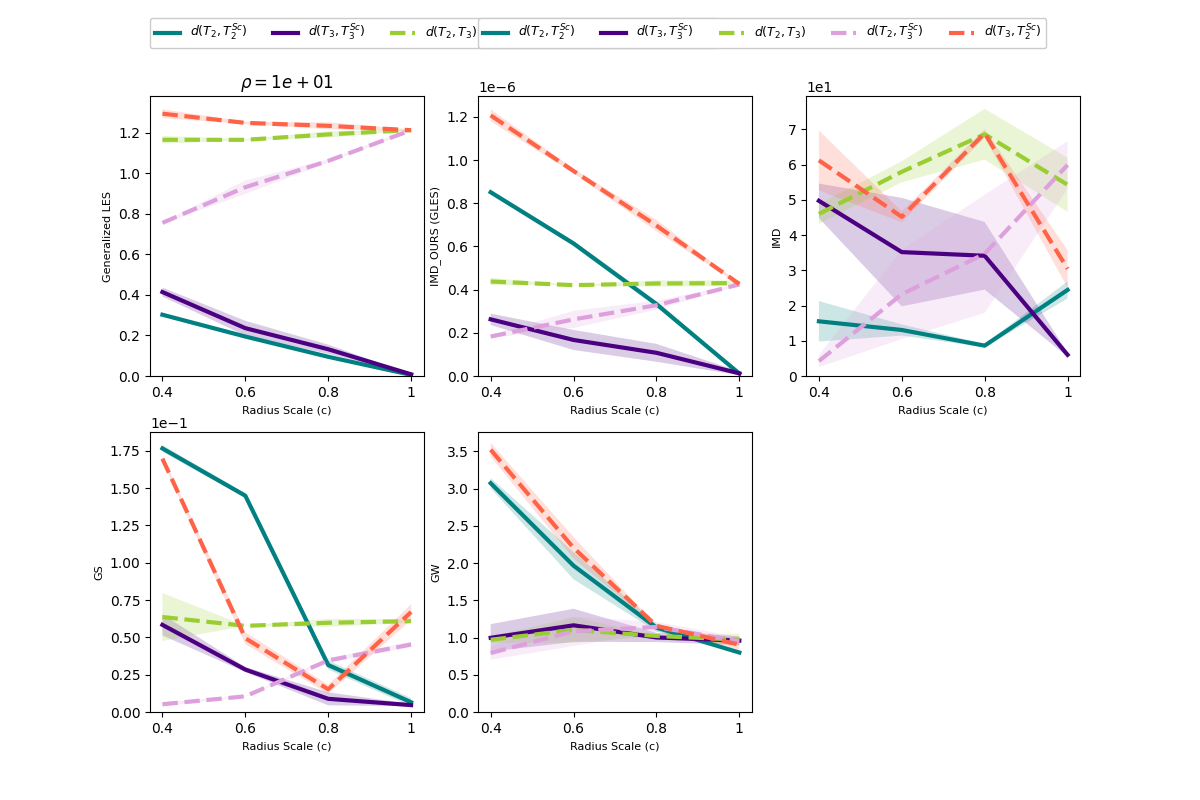}
    \caption{$N = 1000$ and $\rho = 1.0 \times 10$.}
    % \label{fig:placeholder}
\end{figure}
\begin{figure}[H]
    \centering
    \includegraphics[width=0.9\linewidth]{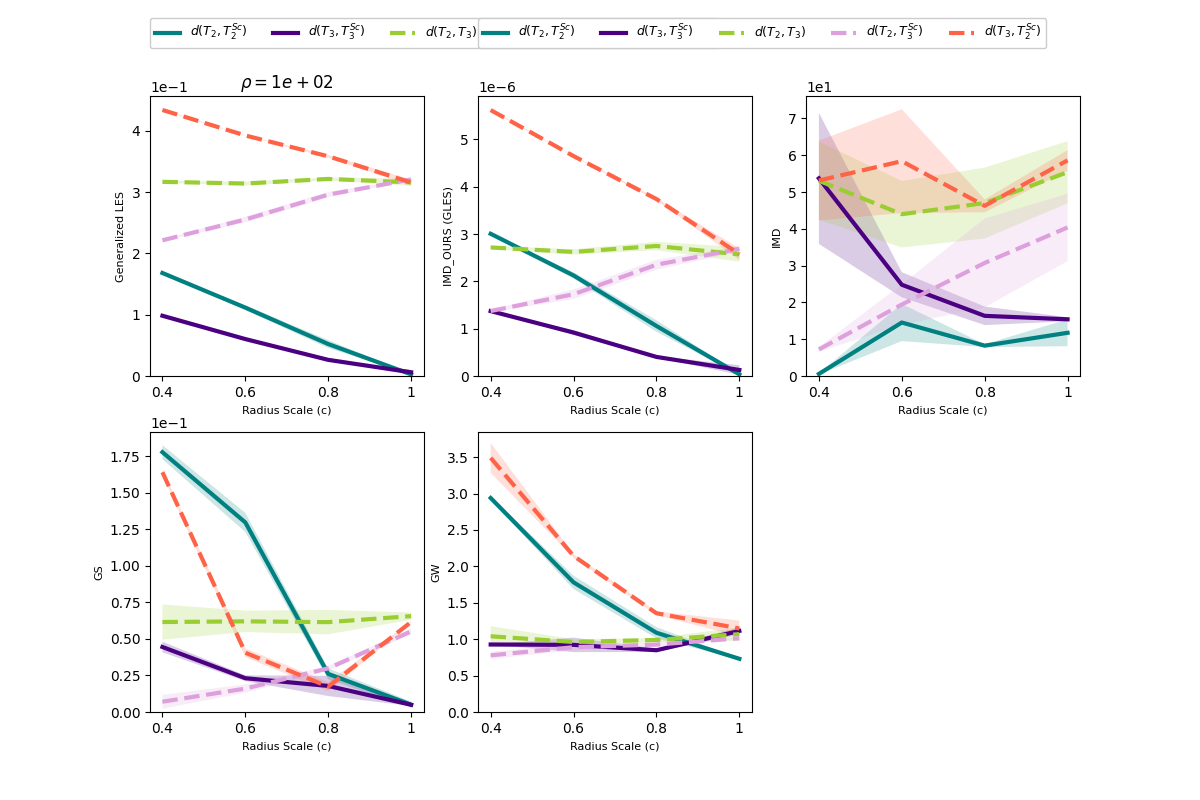}
    \caption{$N = 1000$ and $\rho = 1.0 \times 10^{2}$.}
    % \label{fig:placeholder}
\end{figure}
\begin{figure}[H]
    \centering
    \includegraphics[width=0.9\linewidth]{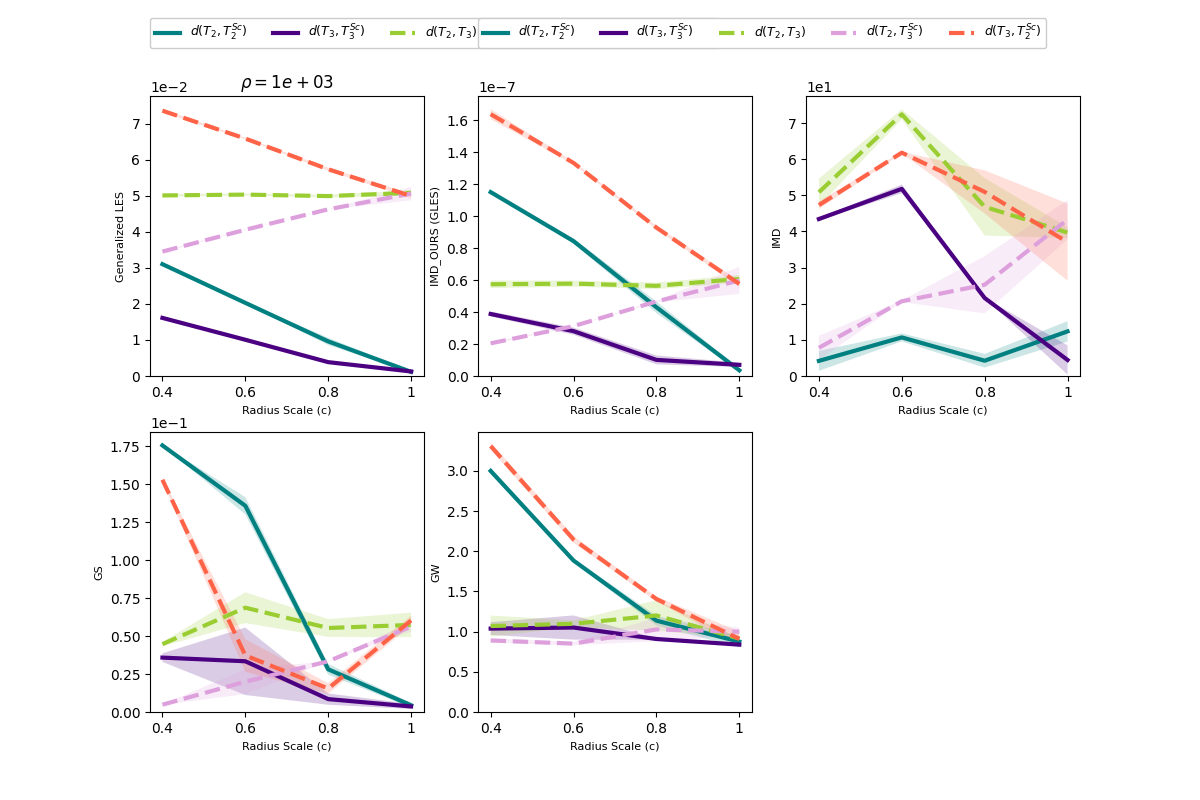}
    \caption{$N = 1000$ and $\rho = 1.0 \times 10^{3}$.}
    % \label{fig:placeholder}
\end{figure}
\begin{figure}[H]
    \centering
    \includegraphics[width=0.9\linewidth]{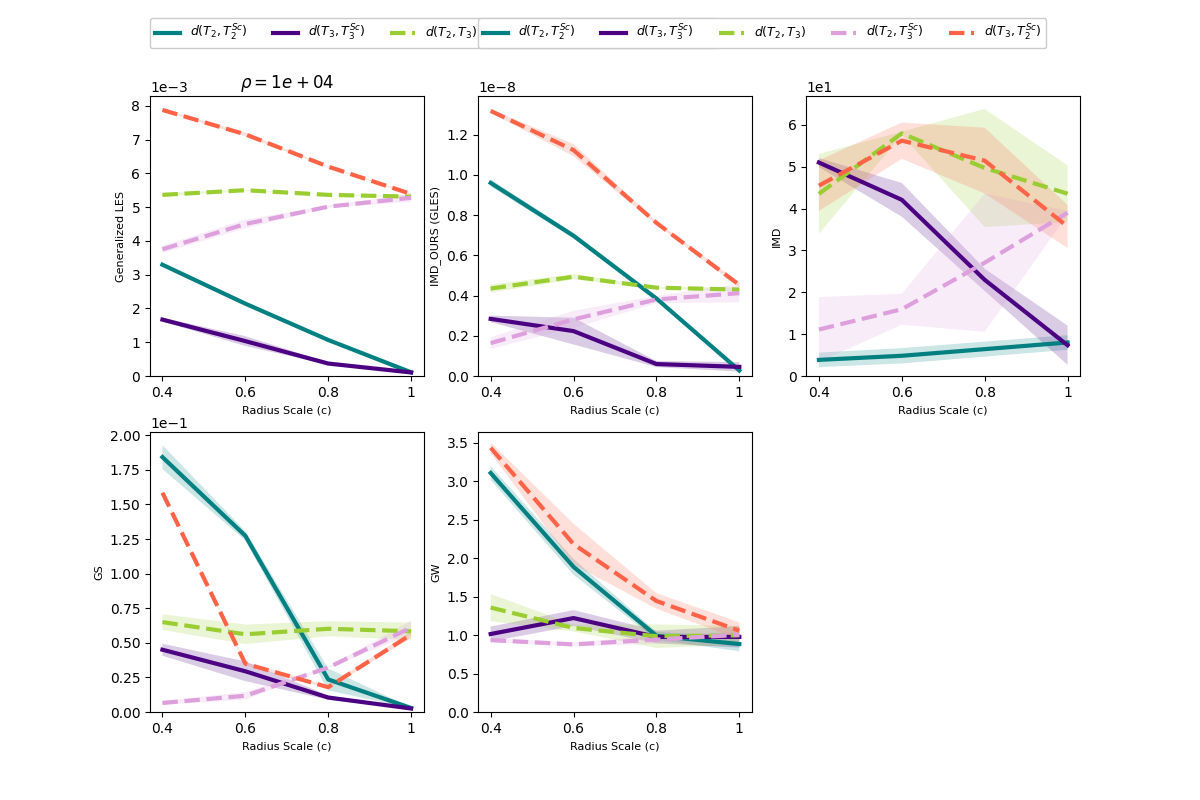}
    \caption{$N = 1000$ and $\rho = 1.0 \times 10^{4}$.}
    % \label{fig:placeholder}
\end{figure}
\subsubsection{$N = 2000$}
\begin{figure}[H]
    \centering
    \includegraphics[width=0.85\linewidth]{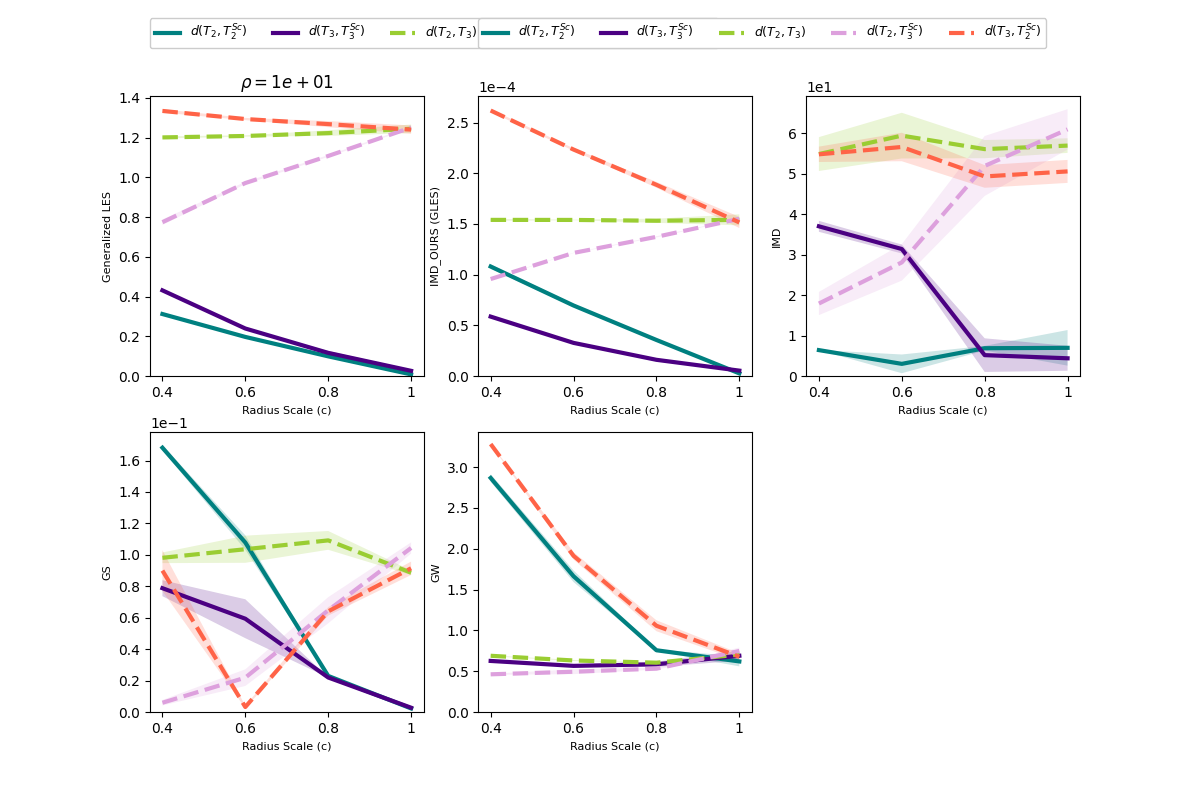}
    \caption{$N = 2000$ and $\rho = 1.0 \times 10$.}
    % \label{fig:placeholder}
\end{figure}
\begin{figure}[H]
    \centering
    \includegraphics[width=0.9\linewidth]{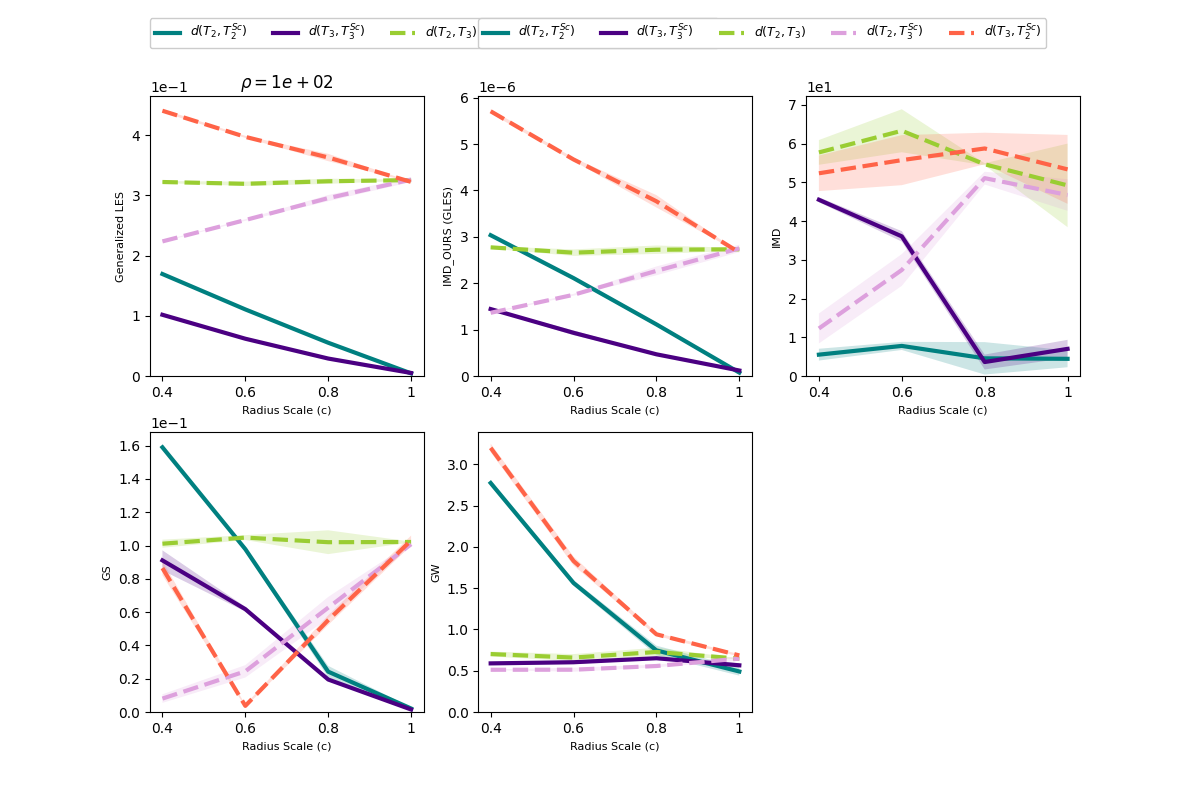}
    \caption{$N = 2000$ and $\rho = 1.0 \times 10^{2}$.}
    % \label{fig:placeholder}
\end{figure}
\begin{figure}[H]
    \centering
    \includegraphics[width=0.9\linewidth]{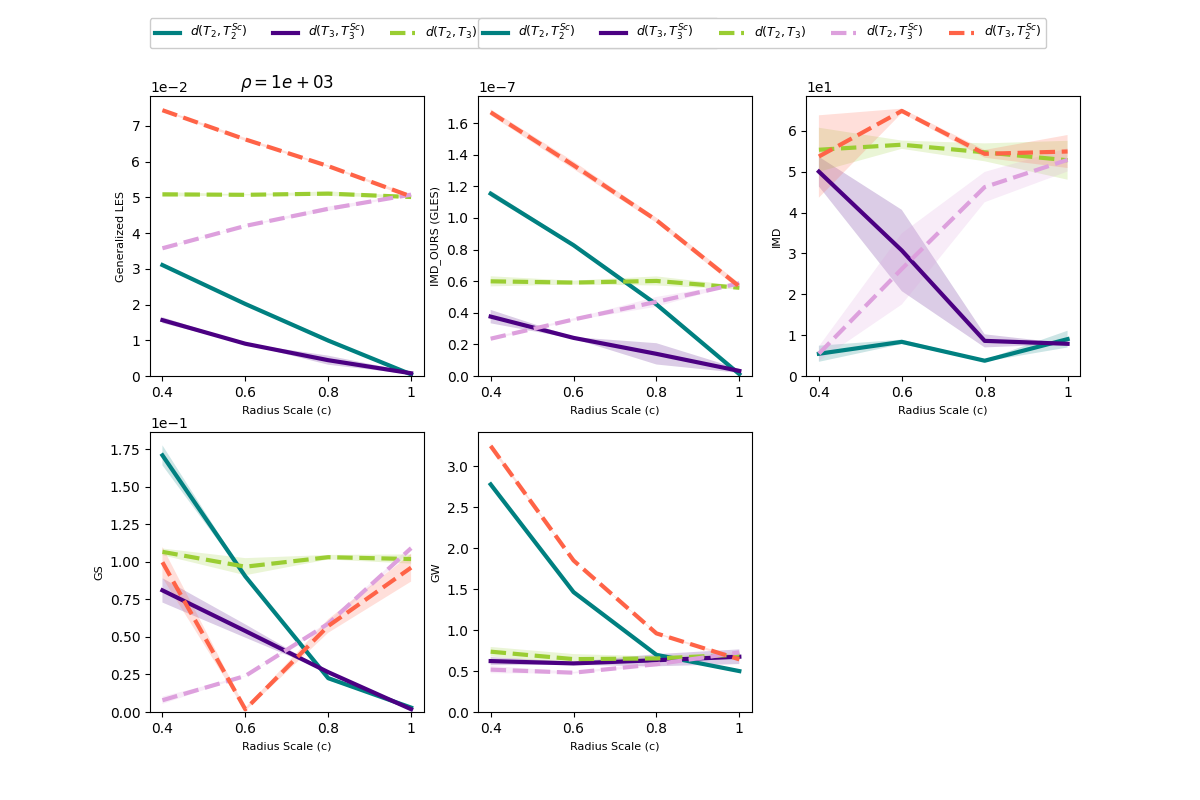}
    \caption{$N = 2000$ and $\rho = 1.0 \times 10^{3}$.}
    % \label{fig:placeholder}
\end{figure}
\subsection{Generalized LES experiments with $\omega_{i} \, \cancel{\propto}\, \lambda_{i}$: an optimization framework for $\mathbf{M}$}
\begin{figure}[h]
    \centering
    \includegraphics[width=0.9\linewidth]{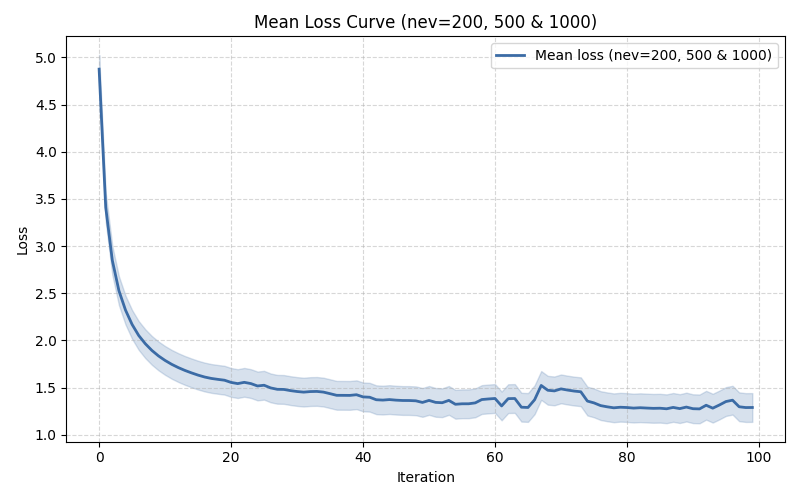}
    \caption{Mean training loss curve for learning $\mathbf{\Tilde{M}}$ for $K = 200, 500, 1000$.}
    % \label{fig:placeholder}
\end{figure}
\subsubsection{$N = 1000$}
\begin{figure}[H]
    \centering
    \includegraphics[width=0.85\linewidth]{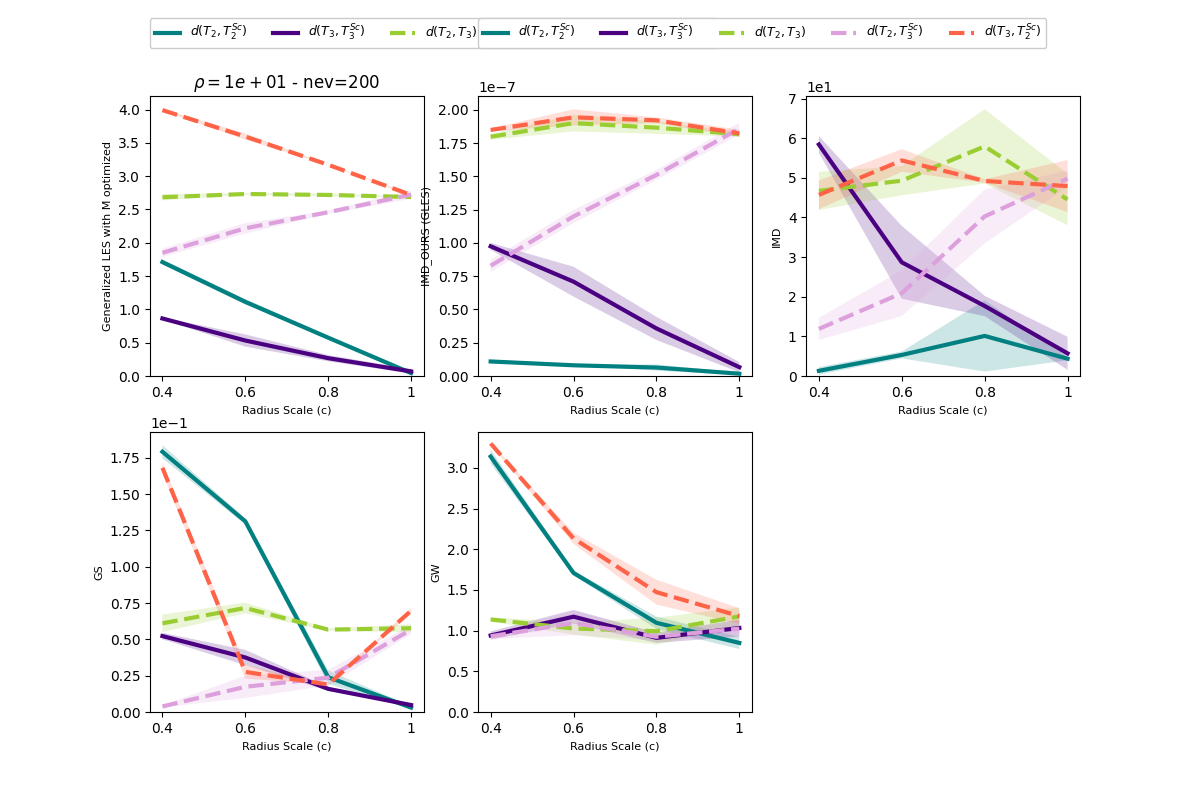}
    \caption{$N = 1000$ and $\rho = 1.0 \times 10$.}
    % \label{fig:placeholder}
\end{figure}
\begin{figure}[H]
    \centering
    \includegraphics[width=0.9\linewidth]{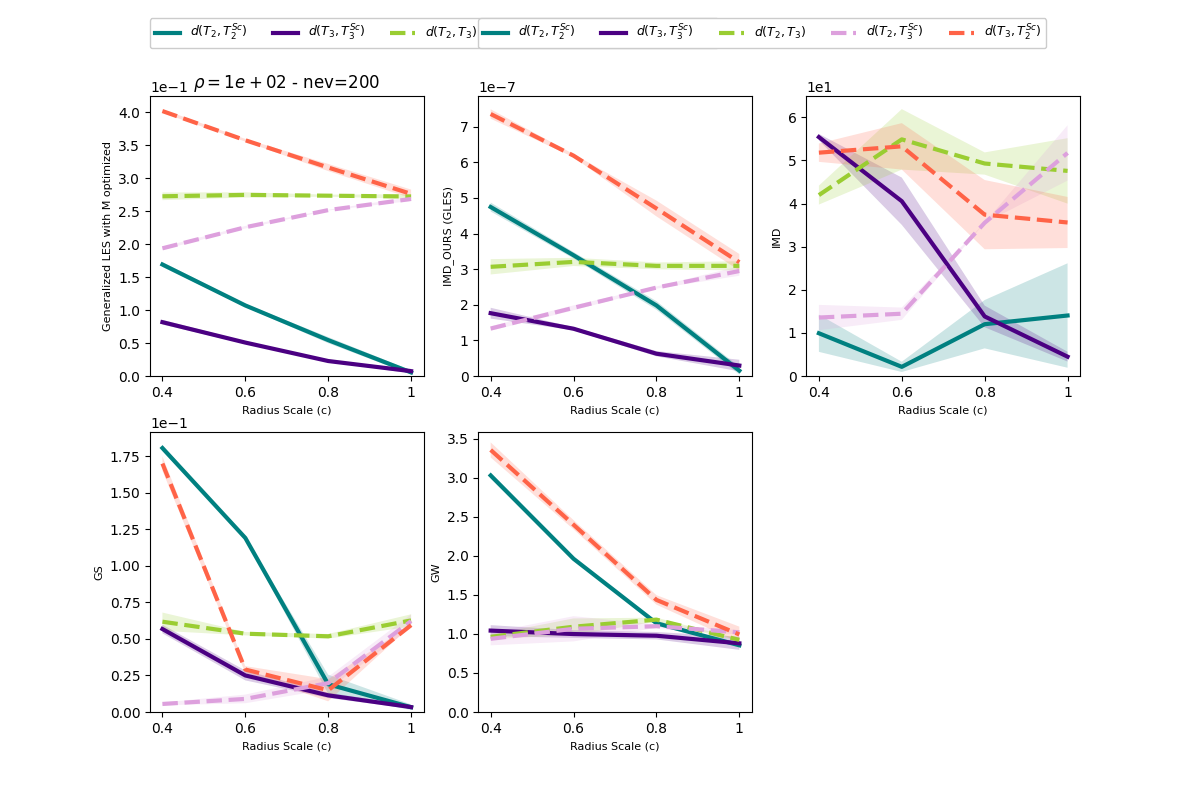}
    \caption{$N = 1000$ and $\rho = 1.0 \times 10^{2}$.}
    % \label{fig:placeholder}
\end{figure}
\begin{figure}[H]
    \centering
    \includegraphics[width=0.9\linewidth]{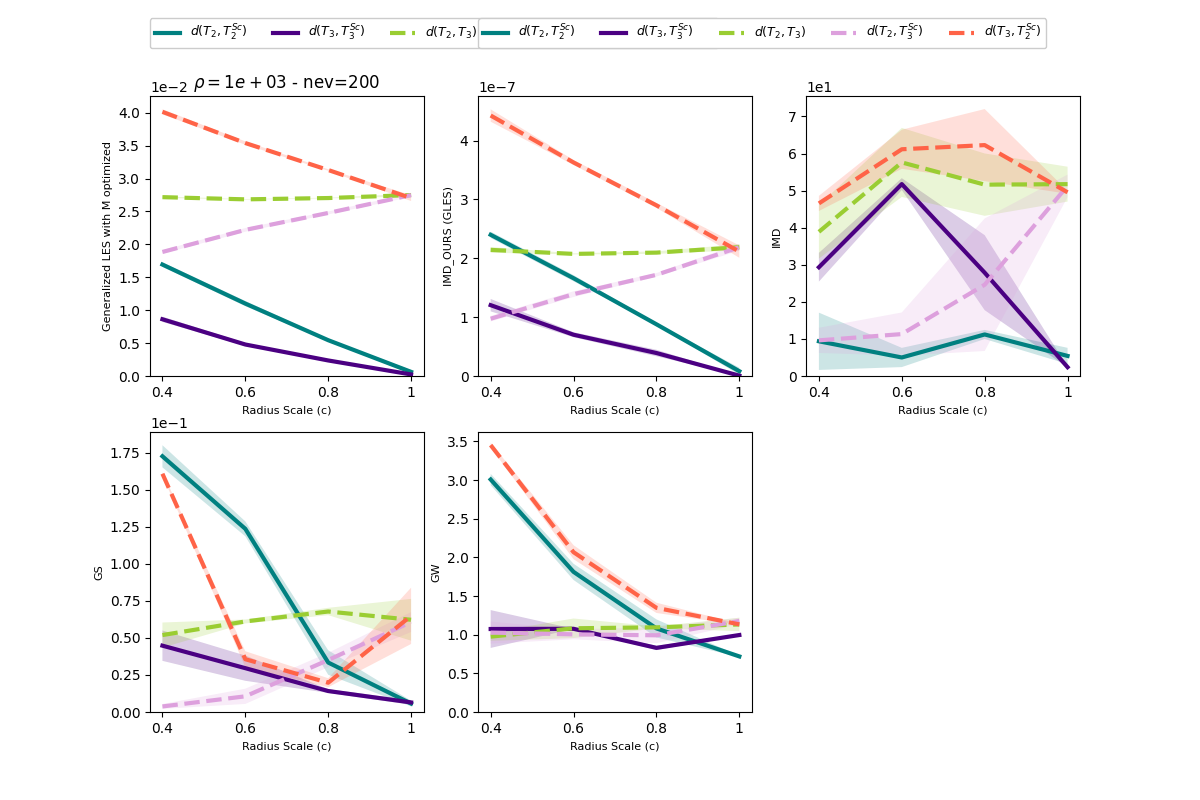}
    \caption{$N = 1000$ and $\rho = 1.0 \times 10^{3}$.}
    % \label{fig:placeholder}
\end{figure}
\begin{figure}[H]
    \centering
    \includegraphics[width=0.9\linewidth]{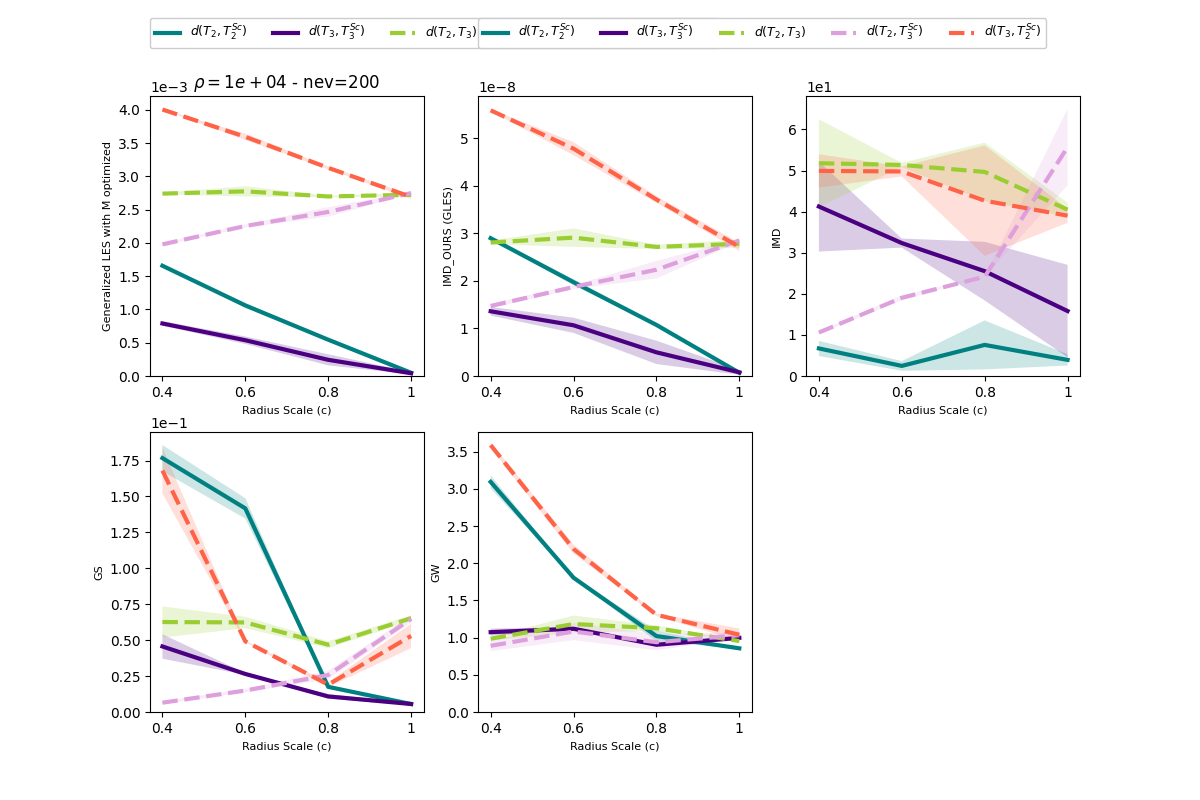}
    \caption{$N = 1000$ and $\rho = 1.0 \times 10^{4}$.}
    % \label{fig:placeholder}
\end{figure}
\subsubsection{$N = 2000$}
%
%
%
%

% \begin{figure}[h]
%     \centering
%     \includegraphics[width=0.9\linewidth]{img/section_4/tori_comparisons_generalised_les_M_opt_NEV_200_N_2000_rho_10.0_mu_0.000_gamma_1e-08.png}
%     \caption{$N = 2000$ and $\rho = 1.0 \times 10$.}
%     % \label{fig:placeholder}
% \end{figure}
%
%
%
%
\begin{figure}[H]
    \centering
    \includegraphics[width=0.85\linewidth]{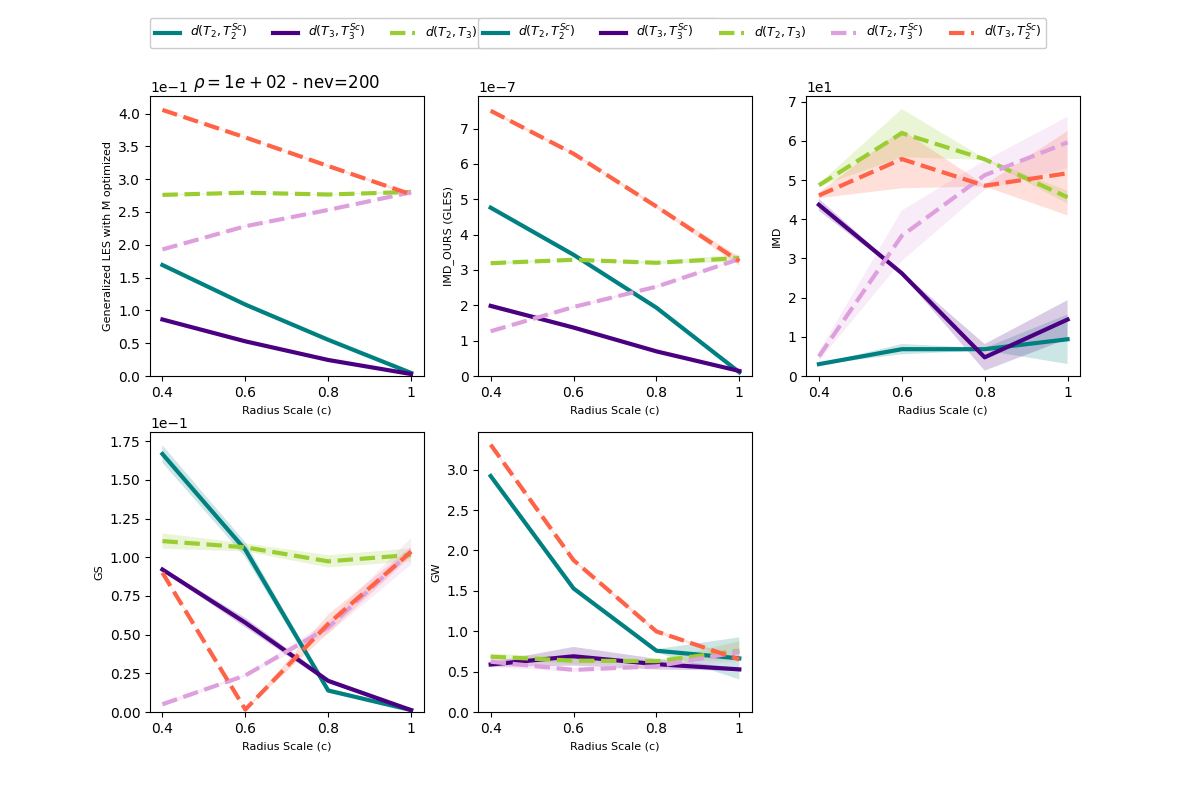}
    \caption{$N = 2000$ and $\rho = 1.0 \times 10^{2}$.}
    % \label{fig:placeholder}
\end{figure}
\begin{figure}[H]
    \centering
    \includegraphics[width=0.9\linewidth]{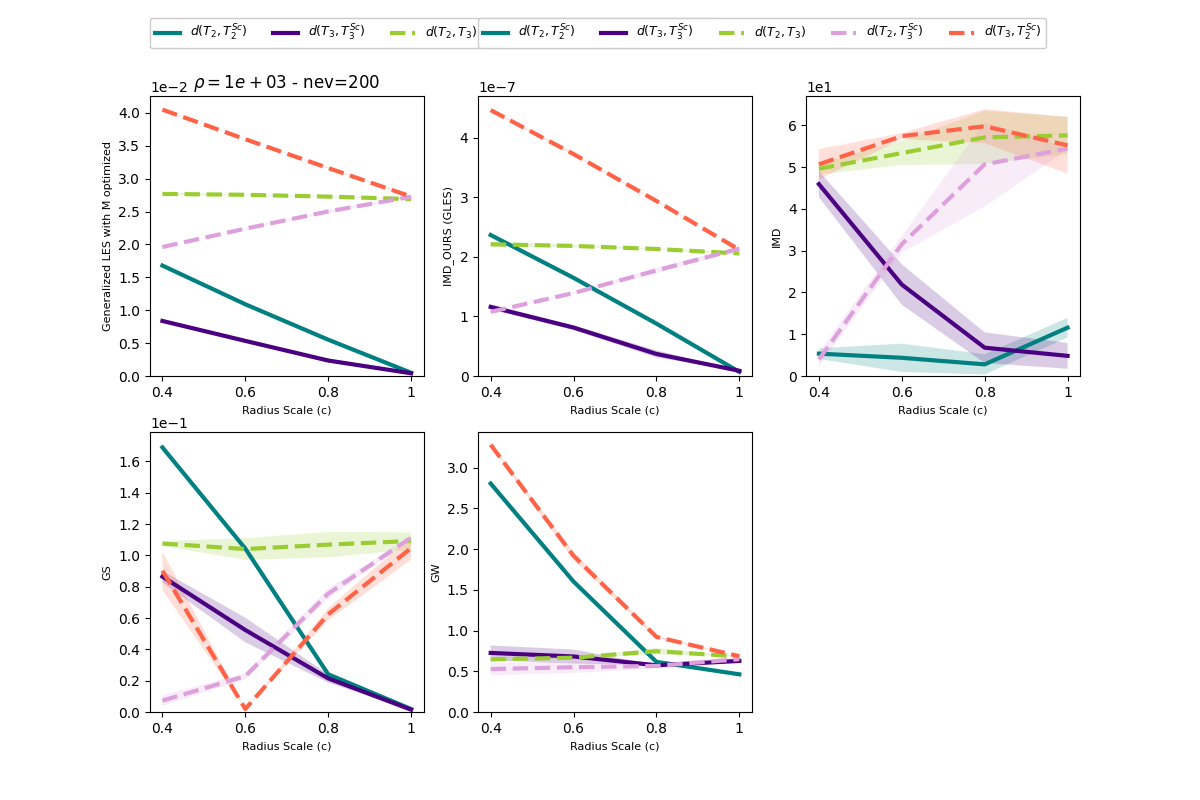}
    \caption{$N = 2000$ and $\rho = 1.0 \times 10^{3}$.}
    % \label{fig:placeholder}
\end{figure}
\begin{figure}[H]
    \centering
    \includegraphics[width=0.9  \linewidth]{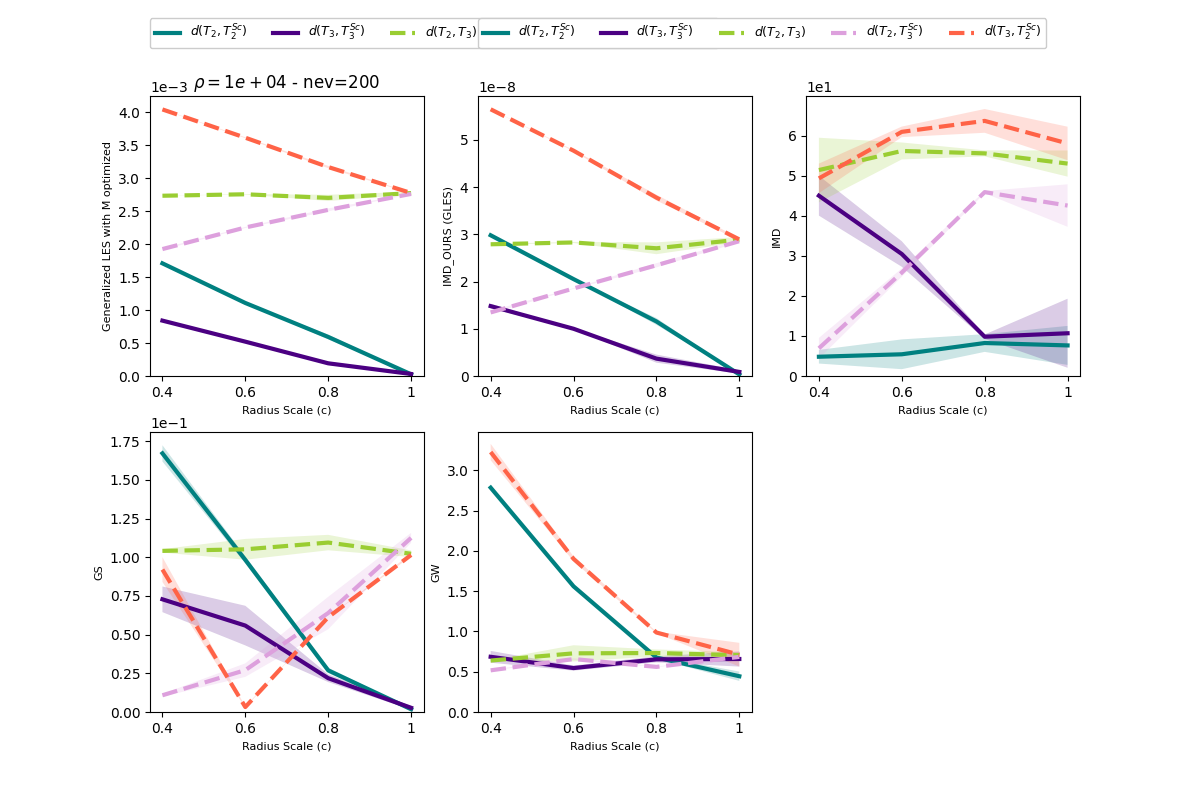}
    \caption{$N = 2000$ and $\rho = 1.0 \times 10^{4 }$.}
    % \label{fig:placeholder}
\end{figure}
\section{Further applications: towards functional GNNs }\label{functionalGNNS}
Constructing sophisticated embeddings for graph neural networks represent a way to learn non-trivial representation that underlies the complex data. This also provides a way to rigorously establish a message passing or attention based algorithms with respect to the embedding manifold. In light of this, we apply part of the generalized alpha-Procrustes framework in order to establish some novel embeddings for GNNs operating with respect to probability measures and covariance operators. 
\subsection{Generalised embeddings for learning probability measures}
Finite dimensional approaches exposed in \citep{kolouri2021wasserstein} establishes a way to learn graph embeddings based on the definition of the \textbf{linear Wasserstein embedding}. In this work, the authors compute the isometric embedding of the probability measures such that the $2$-Wasserstein distance is approximated by the Euclidean distance between embedded images. We, therefore, seek to provide some arguments that generalises this construction using our framework. \\

For the sake of clarity we review the Monge problem that underlies the method in \citep{kolouri2021wasserstein}. Let $\mu, \nu$ be Borel probability measures defined on a subset of $\mathbb{R}^{d}$ with corresponding probability density function $f_{\mu}$ and $f_{\nu}$, \textit{i.e.}, $d\mu(x) = f_{\mu}(x)dx$. Let $\mu$ be an absolutely continuous probability measure with respect to the Lebesgue measure then by virtue of Brenier's theorem, given some convex function one has the unique transport plan $T = \nabla \Phi$. Conversely, given some function $\Phi:\mathbb{R}^{d} \rightarrow \mathbb{R}$ which is convex and is such that $T_{\#}\mu = \nu$ then $T := \nabla\Phi$ is \textbf{necessarily} the optimal transport plan which minimizes the quantity 
\begin{equation}
    \int_{\mathbb{R}^{d}}||x - T(x)||^{2}_{2} \, d\mu(x). 
\end{equation}
Now if both $\mu$ and $\nu$ are absolutely continuous with respect to the Lebesgue measure then $T_{\#}\mu = \nu$ is equivalent the non-linear partial differential equation
\begin{equation}\label{mongeAmpereIntro}
    f_{\nu}(x) = f_{\mu}(T(x)) \cdot \det (\nabla T (x)) \,\,\, \text{a.s}.
\end{equation}
This is the so-called the \textbf{Monge-Ampère equation} and enters in the definition of the isometric embedding $\phi$ (see \citep{kolouri2021wasserstein}). It is therefore licit to ask if one can extend the Definition \ref{mongeAmpereIntro} by virtue of the generalised alpha-Procrustes framework and hence create generalized embeddings for high dimensional multivariate data comparison/alignment. \textit{For instance, can the generalised $2$-Wasserstein \ref{generalisedWp} approximate $|| \tilde{\phi}(\mu) - \tilde{\phi}(\nu)||_{\mathbf{M}^{-1}}$ where $\tilde{\phi}$ is the generalised isometric embedding (see \citep{kolouri2021wasserstein} for the definition of $\phi$)?} We immediately note that this will require the extension of Definition \ref{mongeAmpereIntro}.
\subsubsection{Generalized graph embeddings in finite dimensions}\label{functionalGNNFiniteDim}
We apply optimal transport techniques established above not to move mass over a fixed graph structure, but to embed entire graphs into a common reference space in order to compare them directly. Concretely, define a set of $N$ individual graphs $G_{i}$ for $i=1,\cdots N$ such for each $i$ we have $V(G_{i})$, the set of vertices and $E(G_{i})$, the set of edges for each $G_{i}$.  The set of node features is defined as $\{\mathbf{x}_{\nu}\}_{\nu \in V(G_{i})}$ and the set of edge features $\{\mathbf{e}_{u\nu}\}_{u, \nu \in E(G_{i})}$. The graph embedding at the level of the nodes are defined as $\mathbf{h}_{u}$ for $u \in V(G_{i})$. The final graph embedding is expressed as $Z_{i} = h(G_{i})$, where $h$ is the embedding function for the \textbf{whole graph} $G_{i}$. We interpret this set of embeddings as the support of a discrete probability measure
\begin{equation}
    \mu_{i} = \frac{1}{|V(G_{i})|}\sum_{\nu \in V(G_{i})}\delta_{z_{\nu}},
\end{equation}
where $\delta_{z_{u}}$ denotes the Dirac mass at the node embedding $z_{u}$. This allows us to treat each graph as a probability measure in the feature space. \\

The generalized Monge map that pushes the reference measure $\alpha$ to $\mu$ is defined as
\begin{equation}
    \widetilde{T}_{\mu_{i}} = \arg\min_{T \in MP(\mu_{i}, \nu_{i})} \int_{\mathbb{R^{d}}}|| z - T(z) ||_{\mathbf{M}^{-1}}d\alpha(z). 
\end{equation}
Define $\tilde{\phi}$ the isometric embedding for probability measures on the  space endowed with $||\cdot ||_{\mathbf{M}^{-1}}$. The \textbf{generalized WEGL} of $G_{i}$ is then $\tilde{\phi}(h(G_{i}))$. This is ensured by the Jacobian
\begin{equation}
    f_{\nu_{i}}(x) = f_{\mu_{i}}(\widetilde{T}(z)) \cdot \det (\nabla \widetilde{T} (z)) \,\,\, \text{a.s}.
\end{equation}
A trivial expansion allows us to write down the generalized LOT approximation for the node embeddings as follows 
\begin{equation}
    || \tilde{\phi}(\mathbf{h}_{u}) - \tilde{\phi}(\mathbf{h}_{\nu})||_{\mathbf{M}^{-1}} \approx \tilde{W}_{2}(\mathbf{h}_{u}, \mathbf{h}_{\nu}),
\end{equation}
where $\tilde{W}_{2}$ corresponds to the generalized 2-Wasserstein distance (\ref{generalisedWp}). \\

While elegant and effective for distributions in $\mathbb{R}^{d}$, this approach is intrinsically limited to finite-dimensional settings and relies heavily on the choice of reference measure as well as assumptions about smoothness and absolute continuity. Generalising the latter turns out to be a highly non-trivial task as the infinite dimensional considerations require careful treatment of the Monge-Ampère equation (\ref{mongeAmpereIntro}) as exposed in \citep{nolot2013}. We leave this part for future work. 

\end{document}